\documentclass[11pt]{article}

\usepackage{acl}
\usepackage{times}
\usepackage{latexsym}

\makeatletter
\patchcmd{\@float}{\@fp@prep}{\@fp@prep\thispagestyle{empty}}{}{}
\patchcmd{\@dblfloat}{\@fp@prep}{\@fp@prep\thispagestyle{empty}}{}{}
\patchcmd{\@fp@output}{\@outputpage}{\thispagestyle{empty}\@outputpage}{}{}
\makeatother

\makeatletter
\def\ps@plain{%
  \let\@oddhead\@empty
  \let\@evenhead\@empty
  \let\@oddfoot\@empty
  \let\@evenfoot\@empty
}
\makeatother

\usepackage[T1]{fontenc}

\usepackage[utf8]{inputenc}

\usepackage{microtype}

\usepackage{inconsolata}

\usepackage{graphicx}
\usepackage{multirow}
\usepackage{colortbl}
\usepackage{amsmath}
\usepackage{booktabs}
\usepackage{amssymb}
\usepackage{pifont}
\usepackage{enumitem}
\usepackage{todonotes}
\usepackage[most]{tcolorbox}
\usepackage{listings}
\usepackage{floatpag} 
\usepackage{multicol}
\usepackage{lipsum}

\usepackage{longtable}
\usepackage{xcolor}
\usepackage{floatpag}

\colorlet{punct}{red!60!black}    
\colorlet{delim}{green!60!black}  
\colorlet{numb}{magenta!60!black}

\lstdefinelanguage{json}{
    basicstyle=\ttfamily\small,
    numbers=none,
    numberstyle=\tiny\color{gray},
    stepnumber=1,
    numbersep=8pt,
    showstringspaces=false,
    breaklines=true,
    frame=single,
    backgroundcolor=\color{gray!10},
    literate=
     *{0}{{{\color{blue}0}}}{1}
      {1}{{{\color{blue}1}}}{1}
      {2}{{{\color{blue}2}}}{1}
      {3}{{{\color{blue}3}}}{1}
      {4}{{{\color{blue}4}}}{1}
      {5}{{{\color{blue}5}}}{1}
      {6}{{{\color{blue}6}}}{1}
      {7}{{{\color{blue}7}}}{1}
      {8}{{{\color{blue}8}}}{1}
      {9}{{{\color{blue}9}}}{1}
      {:}{{{\color{punct}{:}}}}{1}
      {,}{{{\color{punct}{,}}}}{1}
      {\{}{{{\color{delim}{\{}}}}{1}
      {\}}{{{\color{delim}{\}}}}}{1}
      {[}{{{\color{delim}{[}}}}{1}
      {]}{{{\color{delim}{]}}}}{1},
}

\tcbset{
    colback=white, 
    colframe=black, 
    boxrule=0.2mm,  
    width=\textwidth,
    arc=0mm          
}

\definecolor{color3}{rgb}{0.7,0.3,0.7}

\floatpagestyle{empty}  

\title{\textsc{PRMBench}: A Fine-grained and Challenging Benchmark for \\ Process-Level Reward Models}

\definecolor{mygreen}{RGB}{64, 114, 78}
\definecolor{myred}{RGB}{192, 70, 74}

\renewcommand{\thefootnote}{\dag}
\makeatletter
\renewcommand\@fnsymbol[1]{%
  \ifcase#1\or \dag\else \@arabic{#1}\fi}
\makeatother

\author{
Mingyang Song$^{1,2}$, Zhaochen Su$^3$, Xiaoye Qu$^2$, Jiawei Zhou$^4$\thanks{Equal senior contribution}, Yu Cheng$^{5}$\footnotemark[1] \\
$^{1}$Fudan University, 
$^{2}$Shanghai AI Laboratory,
$^{3}$Soochow University \\
$^{4}$Stony Brook University, $^{5}$The Chinese University of Hong Kong\\
\texttt{mysong23@m.fudan.edu.cn}; \texttt{suzhaochen0110@gmail.com}; \\
\texttt{quxiaoye@pjlab.org.cn};
\texttt{jzhou@ttic.edu};
\texttt{chengyu@cse.cuhk.edu.hk};\\
 Project Page: \href{https://prmbench.github.io/}{\texttt{%
    \textcolor[rgb]{0.7,0.1,0.1}{h}%
    \textcolor[rgb]{0.8,0.4,0.1}{t}%
    \textcolor[rgb]{0.7,0.7,0.0}{t}%
    \textcolor[rgb]{0.1,0.5,0.1}{p}%
    \textcolor[rgb]{0.1,0.5,0.5}{s}%
    \textcolor[rgb]{0.1,0.1,0.7}{:}%
    \textcolor[rgb]{0.5,0.1,0.5}{/}%
    \textcolor[rgb]{0.5,0.1,0.5}{/}%
    \textcolor[rgb]{0.7,0.1,0.1}{p}%
    \textcolor[rgb]{0.8,0.4,0.1}{r}%
    \textcolor[rgb]{0.7,0.7,0.0}{m}%
    \textcolor[rgb]{0.1,0.5,0.1}{b}%
    \textcolor[rgb]{0.1,0.5,0.5}{e}%
    \textcolor[rgb]{0.1,0.1,0.7}{n}%
    \textcolor[rgb]{0.5,0.1,0.5}{c}%
    \textcolor[rgb]{0.7,0.1,0.1}{h}%
    \textcolor[rgb]{0.8,0.4,0.1}{.}%
    \textcolor[rgb]{0.1,0.5,0.5}{g}%
    \textcolor[rgb]{0.1,0.1,0.7}{i}%
    \textcolor[rgb]{0.5,0.1,0.5}{t}%
    \textcolor[rgb]{0.7,0.1,0.1}{h}%
    \textcolor[rgb]{0.8,0.4,0.1}{u}%
    \textcolor[rgb]{0.7,0.7,0.0}{b}%
    \textcolor[rgb]{0.1,0.5,0.1}{.}%
    \textcolor[rgb]{0.1,0.5,0.5}{i}%
    \textcolor[rgb]{0.1,0.1,0.7}{o}%
}}
}

\begin{document}
\maketitle

\begin{abstract}
Process-level Reward Models (PRMs) are crucial for complex reasoning and decision-making tasks, where each intermediate step plays an important role in the reasoning process. 
Since large language models (LLMs) suffer from various types of errors during the reasoning process, PRMs are required to possess nuanced capabilities for detecting various implicit error types in real-world scenarios. However, current benchmarks primarily focus on step correctness, failing to evaluate PRMs' performance systematically. To address this gap, we introduce \textsc{PRMBench}, a process-level benchmark specifically designed to assess the fine-grained error detection capabilities of PRMs. \textsc{PRMBench} comprises 6,216 carefully designed problems and 83,456 step-level labels, evaluating models across multiple dimensions, including \textit{simplicity}, \textit{soundness}, and \textit{sensitivity}. In our experiments on 25 models, spanning across both open-source PRMs and LLMs prompted as critic models, we uncover significant weaknesses in current PRMs. These findings reveal the challenges inherent in process-level evaluation and highlight key directions for future research, establishing \textsc{PRMBench} as a robust testbed for advancing research on PRM evaluation and development. 
\end{abstract}

\begin{figure}
    \centering
    \includegraphics[width=\linewidth]{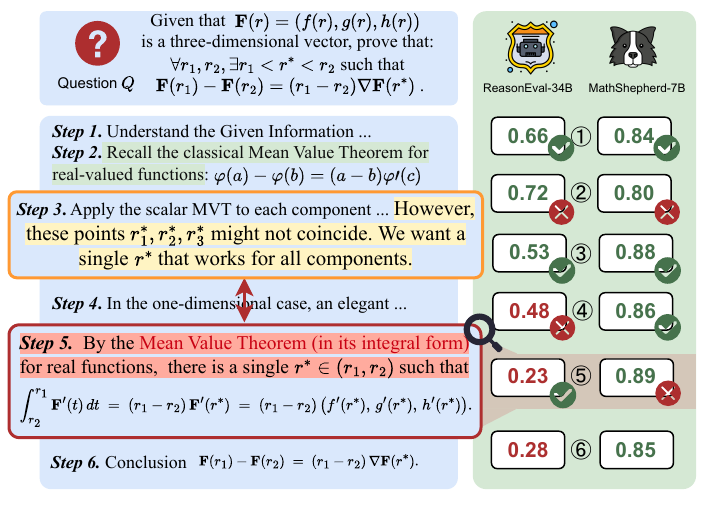}
    \caption{(Left): Given a question $Q$, the reasoning step 2 and 5 of OpenAI-o1 model contains errors. (Right): The step-level reward scores generated by ReasonEval-34B \cite{reasoneval} and MathShepherd-7B \cite{wang2023math_shepherd}. Green scores indicate the PRM prefer labeling this step as correct while red scores indicate the PRM prefer identifying this step as incorrect.
    }
    \label{fig:head_fig}
    \vspace{-10pt}
\end{figure}
\renewcommand{\arraystretch}{0.9} 
\newcommand{\tabincell}[2]{\begin{tabular}{@{}#1@{}}#2\end{tabular}}
\begin{table*}[ht]
  \centering
  \vspace{1mm}

  \resizebox{\textwidth}{!}{
    \begin{tabular}{lccccccc}
    \toprule
     & \tabincell{c}{\textbf{PRM}\\ \textbf{Benchmarks?} } & \tabincell{c}{\textbf{Error Type}\\ \textbf{Detection?}}&  \tabincell{c}{\textbf{Fine-grained} \\ \textbf{classes}$^\dagger$} & \tabincell{c}{\textbf{Step} \\ \textbf{Evaluation}}   & \textbf{Annotator} & \tabincell{c}{\textbf{Test Case}\\ \textbf{Size}} & \tabincell{c}{\textbf{Average}\\ \textbf{Steps}} \\
    \midrule
    MR-GSM8K \cite{mr_gsm8k}& \ding{55}& \ding{55}& 1 &\ding{51} &  Human & 2,999 & 8.3 \\
    RMBench \cite{liu2024rmbench}& \ding{55} & \ding{55} & 1 &\ding{55}& Synthetic + Human & 1,327 & -\\
    CriticBench \cite{lin2024criticbench}& \ding{55}& \ding{55} & 1 & \ding{55}  & - & - & -\\
    MathCheck-GSM \cite{MathCheck_GSM}& \ding{55}  & \ding{55} & 1 & \ding{51}  & Synthetic & 516 & - \\
    MR-Ben \cite{zeng2024mrben} & \ding{55}& \ding{55}& 1 &\ding{51} & Human & 5,975 & 9.5 \\
    ProcessBench \cite{qwenprocessbench}& \ding{51} & \ding{55} & 1 & \ding{51} & Human & 3,400 & 7.1 \\

    \midrule
    \textsc{PRMBench}& \ding{51}& \ding{51}  & 9 &\ding{51} & Synthetic + Human & 6,216 & 13.4 \\ 
    \bottomrule
    \end{tabular}
  }
    \caption{Comparison between our proposed \textsc{PRMBench} and other benchmarks or datasets related to reasoning process assessment.
  $^\dagger$: Fine-grained classes mean the number of evaluation categories according to fine-grained error types of model generation. 
  }
  \vspace{-10pt }
  \label{tab:comparison}
\end{table*}

\section{Introduction}

Recent large language models (LLMs) \cite{openai2024gpt4o, openai_o1_2024, qwen-qwq-32b-preview}, trained on large-scale reinforcement learning, have achieved significant performance in complex reasoning tasks such as mathematics and code generation \cite{ yu2023metamath,guo2024deepseekcoder,deepmind_gemini_2_flash,luo2023wizardmath,qu2025survey}.
A key factor behind their successes is the use of process reward models (PRMs) \cite{wang2023math_shepherd,lets_verify_stepbystep,uesato2022solving}, which can help evaluate the correctness of reasoning steps and train LLMs with appropriate rewards. \cite{qin2024o1replication_pengfeiliu,zhang2024llamaberry}. 

However, during the reasoning process of the LLM, it suffers from various types of errors, while recent PRMs are not able to identify all these error types precisely.
For instance, as illustrated in Figure \ref{fig:head_fig}, given a question $Q$, OpenAI o1 model \cite{openai_o1_2024} generates a reasoning procedure containing errors, where step 2 is redundant, step 5 is inconsistent with step 3, and the theory used in step 5 is incorrect due to deception falling. 
Under these circumstances, ReasonEval-34B \cite{reasoneval} and Math-Shepherd-7B \cite{wang2023math_shepherd} fail to identify these errors accurately. Math-Shepherd-7B fails to recognize step 5 as an error, while ReasonEval-34B correctly identifies step 5 but incorrectly classifies step 4 as an error, indicating the unreliability of current PRMs.
\label{sec:many_error_type}





To evaluate the diverse error-detection capabilities of PRMs, we present \textsc{PRMBench}, a comprehensive and fine-grained benchmark specifically designed for assessing PRMs. In contrast to existing process-level benchmarks, which can only evaluate the detection of a single error type~\cite{qwenprocessbench,zeng2024mrben}, \textsc{PRMBench} offers a more nuanced evaluation.
Specifically, \textsc{PRMBench} systematically assesses the performance of PRMs across diverse error categories, including \textit{simplicity, soundness}, and \textit{sensitivity}. 
Our benchmark includes 6,216 fine-grained data instances spreading across three major evaluation categories and nine sub-categories, whose quality is ensured by professional annotators. Additionally, we utilize style-controlled data curation methods to ensure evaluation samples under consistent difficulty levels, mitigating confounding variables.

In our study, we conduct extensive experiments using \textsc{PRMBench} to evaluate 25 models, including dedicated PRMs and SOTA general-purpose or mathematical LLMs, prompted as critic models. We observe that all PRMs partially grasp multi-step process evaluation. Specifically, Gemini-2-Thinking achieves the best performance of 68.8, but still significantly falls behind the human 
performance of 83.8. Through extensive analysis, we discover a significant inconsistency between step-level and outcome-level evaluations. By evaluating models with \textsc{PRMBench}, we can assess PRMs' ability to detect step-level errors and false positives, reducing the risk of outcome hacking~\cite{gao2025on}.
To sum up, our contributions are as follows:

\begin{figure*}
    \centering
    \includegraphics[width=0.9\linewidth]{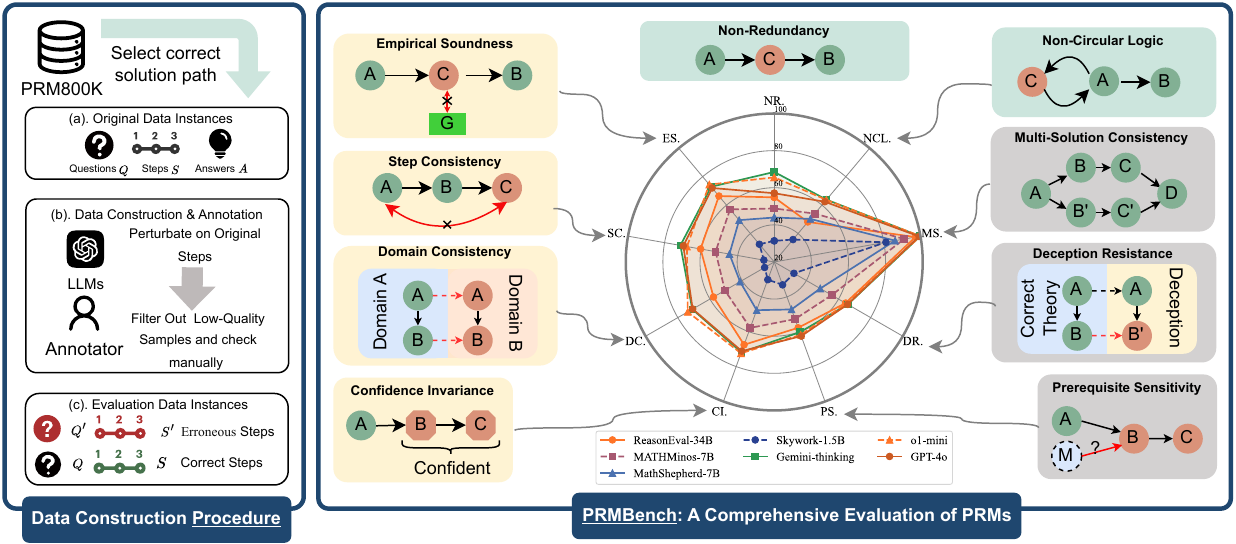}
    \caption{An overview of our \textsc{PRMBench}. The left part illustrates our data curation procedure. In the right part of the figure, we showcase demonstrations of our evaluation categories and the relative performance of tested models, with \colorbox[rgb]{0.788235294117647, 0.8980392156862745, 0.8666666666666667}{green}, \colorbox[rgb]{1.0, 0.9529411764705882, 0.8352941176470589}{yellow}, and \colorbox[rgb]{0.8274509803921568, 0.8196078431372549, 0.8196078431372549}{gray} boxes indicating \textit{simplicity}, \textit{soundness}, and \textit{sensitivity} respectively, where \colorbox[rgb]{0.88,0.57,0.49}{red} circles represent erroneous steps and \colorbox[rgb]{0.49,0.69,0.58}{green} circles indicate correct regular steps. 
    } 
    \vspace{-10pt}
    \label{fig:main_fig}
\end{figure*}

\begin{itemize}[leftmargin=*]
\setlength{\itemsep}{0pt}
    \item We present \textsc{PRMBench}, the first comprehensive process-level reward model benchmark, comprising 6,216 carefully curated samples and 83,456 step-level labels for a series of evaluations on process-level reward models.

    \item \textsc{PRMBench} covers three carefully-crafted evaluation categories and nine sub-categories including \textit{simplicity}, \textit{soundness}, and \textit{sensitivity}. With these fine-grained evaluation axes, we can conduct tailored assessments of models on their specified capabilities and reveal their potential weaknesses during the rewarding procedure.

    \item Based on our proposed \textsc{PRMBench}, we conduct in-depth pilot experiments on twenty-five models including PRMs along with SOTA LLMs. Our findings uncover critical weaknesses and provide valuable insights to guide future research to improve the capabilities of PRMs.

    \item To facilitate future research, we release the PRM-EVAL toolkit, offering an automated evaluation framework and customizable data generation system. We hope \textsc{PRMBench} will drive progress in step-level reasoning for RLHF and foster further development of more reliable PRMs.
    
 
\end{itemize}

\section{Related Work}
\subsection{Process-level Reward Models}
Process-level reward models (PRMs) have shown improvements over traditional outcome-level reward models (ORMs) in enhancing process-level reasoning accuracy and long-process reasoning abilities \citep{lets_verify_stepbystep, uesato2022solving}. Recently, several PRMs have been proposed for process-level RLHF \cite{wang2023math_shepherd, reasoneval, skyworkprms}, with \citet{lets_verify_stepbystep} releasing a large dataset for multi-step reasoning, and \citet{wang2023math_shepherd} introducing an automatic self-supervised pipeline for process-level labeling. \citet{reasoneval} uses PRMs as auto-evaluators for multi-step reasoning accuracy. As PRM training and data curation have grown, numerous PRMs \cite{skyworkprms, xiong2024rlhflowmath, llemmaprm, mathminos} have emerged, along with critic models using LLM-generated feedback \cite{criticgpt, zhang2024criticv, mathminos}. However, both PRMs and critic models remain fallible, highlighting the need for comprehensive benchmarks.
In this paper, we propose \textsc{PRMBench}, a comprehensive benchmark for evaluating PRMs on fine-grained subjects, establishing a strong foundation for PRM evaluation.

\subsection{Reasoning Benchmarks}
Evaluating the reasoning capabilities of LLMs is crucial for understanding their potential and limitations. ROSCOE \cite{golovneva2022roscoe} introduces a semantic comparison-based multi-step reasoning accuracy evaluation benchmark. However, recent research suggests that labeled data cannot be assumed to cover all possible solution paths exhaustively \cite{wang2023math_shepherd, reasoneval}. To address this, \citet{reasoneval} uses PRMs or Critic models to evaluate step-level reasoning accuracy. However, PRMs are not always accurate in assessing process-level data, underscoring the need for a comprehensive evaluation benchmark. While other benchmarks \cite{liu2024rmbench, li2024vlrewardbench, lin2024criticbench, su2024timo} exist, they are not tailored for PRMs and can’t assess step-level reasoning. Some works \cite{mr_gsm8k, zeng2024mrben, yan2024errorradar} use LLMs to evaluate reasoning steps, but they often overlook implicit error types. Existing error classification works are not specific to PRMs and lack fine-grained step-level labels \cite{li2024error_classification}. To address these gaps, we propose \textsc{PRMBench}, a solution that offers fine-grained evaluation and detects various error types.
\renewcommand{\arraystretch}{0.9} 
\begin{table*}[ht]
  \centering
  \vspace{1mm}

  \resizebox{0.9\textwidth}{!}{
    \begin{tabular}{l ccc ccc cccc}
    \toprule
    	&\textbf{Overall}& \textbf{NR.}&\textbf{NCL.}&\textbf{ES.}&\textbf{SC.}&\textbf{DC.}&\textbf{CI.}&\textbf{PS.}&\textbf{DR.}&\textbf{MS.}\\
        \midrule
            
            Avg. Steps&13.4&15.3&10.3&13.8&14.2&13.3&14.2&12.7&13.4&14.1\\
            Avg. Error Steps&2.1&2.0&2.8&2.8&1.6&1.8&1.7&2.5&2.3&0.0\\
            Avg. First Error Step&7.8&7.8&4.9&8.0&9.1&6.8&11.4&6.2&8.3&N/A\\
            Avg. Question Length&152.7&153.6&152.5&153.5&149.7&152.5&152.7&158.0&153.5&132.2\\
            \# of Instances&6216&758&758&757&758&757&757&756&750&165\\

        \bottomrule
    \end{tabular}
  }
  \vspace{1mm}
    \caption{Statistics of \textsc{PRMBench}. NR., NCL., ES., SC., DC., CI., PS., DR., and MS. represent for Non-Redundancy, Non-Circular Logic, Empirical Soundness, Step Consistency, Domain Consistency, Confidence Invariance, Prerequisite Sensitivity, Deception Resistance, and Multi-Solution Consistency respectively.
  }
  \label{tab:data_statistics}
\end{table*}

\section{\textsc{PRMBench}}

\subsection{Evaluation Categories}
In this section, we provide a detailed introduction to the evaluation categories of \textsc{PRMBench}, which is organized into three main domains: 
\begin{itemize}[leftmargin=*]
\setlength{\itemsep}{0pt}
    \item \textbf{Simplicity} evaluates the ability of PRMs to detect redundancy in reasoning steps. Although redundant steps do not affect correctness, they increase computational costs and reduce efficiency. Additionally, simplifying the reasoning process enhances the clarity of the problem’s core and improves overall understandability.
    \item \textbf{Soundness} assesses the accuracy of the rewards produced by PRMs. As discussed in Section \ref{sec:many_error_type}, errors in reasoning can vary in both causes and manifestations \cite{li2024error_classification}. Therefore, we evaluate not only the correctness of rewards but also the fine-grained performance across different error types and their nuances.
    \item \textbf{Sensitivity} measures PRMs’ robustness to details, such as critical conditions or implicit requirements. Sensitivity is vital for ensuring logical completeness and resilience to misleading information~\cite{wen2025rethinking}, contributing to the overall robustness of PRMs.
    
\end{itemize}

Each domain is further divided into detailed sub-categories for a more granular evaluation, which is discussed in detail below. The overall structure of \textsc{PRMBench} along with representative examples of each sub-category are illustrated in Figure \ref{fig:main_fig}, and the details of every evaluation category and sub-category are shown in Appendix \ref{appendix:evaluation_subjects}.

\subsubsection{Simplicity}

Specifically, the simplicity evaluation category is divided into two sub-categories: Non-Redundancy and Non-Circular Logic, with detailed descriptions provided below:

\paragraph{Non-Redundancy} evaluates the PRMs’ ability to identify redundancy within the reasoning process. Redundancy occurs when the reasoning includes unnecessary steps that do not contribute to the solution, making the process less concise and efficient. These steps can be removed without affecting the correctness of the final solution path.

\paragraph{Non-Circular Logic} assesses the PRMs' ability to detect circular reasoning within the process. Circular logic is a form of redundancy where the reasoning eventually loops back to a previous step, creating an infinite cycle. This sub-category is treated separately due to the frequent occurrence of circular logic in reasoning processes.


\subsubsection{Soundness}
We divide the Soundness category into four sub-categories due to its complexity: Empirical Soundness, Step Consistency, Domain Consistency, and Confidence Invariance. The definition of each sub-category is discussed below.


\paragraph{Empirical Soundness} demands PRMs to detect the counterfactual mistakes within the reasoning process. A counterfactual step refers to a statement within a reasoning chain that contradicts established ground truth $G$. 

\paragraph{Step Consistency} expects PRMs to detect the step-wise contradiction, which means a conflict between a specific step and other steps within a reasoning path. Given a reasoning path $ P = \{S_1, S_2, \dots, S_n\} $, a step contradiction exists if $ S_i \perp S_j $, where $ i, j \in [1, n] $ and $ i \neq j $. 

\paragraph{Domain Consistency} requires PRMs to detect domain inconsistency mistakes, which is a special type of counterfactual. It refers to a step within the reasoning chain that uses a statement or theory valid in other domains or cases but is not valid within the current reasoning chain. 

\paragraph{Confidence Invariance} demands PRMs to detect over-confident errors, a type of counterfactual where an incorrect statement is made with high confidence, contradicting established ground truth.

\subsubsection{Sensitivity}

This category includes three sub-categories: Prerequisite Sensitivity, Deception Resistance, and Multi-Solution Consistency, with detailed descriptions provided below.

\paragraph{Prerequisite Sensitivity} requires PRMs to maintain sensitivity to missing conditions or prerequisite mistakes, which means a flaw in the reasoning chain where critical premises, assumptions, or necessary conditions are absent and this omission results in logical gaps, incomplete reasoning, or biased conclusions.

\paragraph{Deception Resistance} demands PRMs to detect the deception or trap within a reasoning process, that is, statements that appear to be correct but are subtly altered to introduce inaccuracies while maintaining the illusion of correctness. 

\paragraph{Multi-Solution Consistency} expects PRMs to maintain consistency when faced with different solution paths of the same problem. Concretely, we utilize multiple correct reasoning processes of the same question to test whether the PRM can perform correctly. 

\subsection{Data Curation}
\label{sec:data_curation}
We curate the dataset by extracting metadata and constructing test cases according to our category definitions. Detailed statistics of \textsc{PRMBench} are displayed in Table \ref{tab:data_statistics}, with the curation procedure outlined below.

\paragraph{Meta Data Extraction} 
Our metadata is built upon PRM800K \cite{lets_verify_stepbystep}, which provides the questions ($Q$), ground truth answers ($A$), and ground truth step-level solution processes ($S$). We select completely correct solutions from both the training and test sets, filtering out low-quality instances to establish our ground truth answers.

\renewcommand{\thefootnote}{\arabic{footnote}}
\paragraph{Test Case Construction} 
Each test case instance is represented as $(Q', A, S')$, where $Q'$ denotes the test question and $S'$ represents the test solution process, which may include errors. 
With class-specific prompts, as demonstrated in Appendix \ref{appendix_generation_prompts}, we query GPT-4o \cite{openai2024gpt4o} to modify the ground-truth reasoning process into versions containing erroneous steps. 
For the multi-solution, we leverage the newly proposed multi-step reasoning model QwQ\footnote[1]{Qwen/QwQ-32B-Preview: \url{https://huggingface.co/Qwen/QwQ-32B-Preview}} \cite{qwen-qwq-32b-preview} to generate candidate answers for the given questions. These answers are then filtered to exclude unreasonable or incorrect ones, resulting in multi-solution reasoning processes for a single question.

\renewcommand{\arraystretch}{0.85} 
\begin{table*}[h]
\belowrulesep=0pt
\aboverulesep=0pt
\fontsize{14}{21}\selectfont
\centering

\resizebox{\textwidth}{!}{
\begin{tabular}{l c|ccc| ccccc| cc cc}
\toprule[1.5pt]
\multirow{2}{*}{\textbf{Model}} & \multirow{2}{*}{\textbf{Overall}} & \multicolumn{3}{c|}{\textbf{Simplicity}}  & \multicolumn{5}{c|}{\textbf{Soundness}}& \multicolumn{4}{c}{\textbf{Sensitivity}}\\
\cmidrule(lr){3-5} \cmidrule(lr){6-10} \cmidrule(lr){11-14} 
&& \textbf{NR.} & \textbf{NCL.} & \textbf{Avg.} &\textbf{ES} &\textbf{SC.}&\textbf{DC.} &\textbf{CI} & \textbf{Avg.} &\textbf{PS} & \textbf{DR.} & \textbf{MS.} & \textbf{Avg.}   \\
 \midrule
Human Performance & 83.8 & 80.2 & 81.7 & 81.0 & 84.8 & 85.0& 81.3 & 86.1 & 84.3 & 81.6 & 82.1 & 96.2 & 86.0 \\

\hline \multicolumn{14}{c}{\textit{\textbf{Open-source Process Level Reward Models}}} \\   \hline 
\href{https://huggingface.co/Skywork/Skywork-o1-Open-PRM-Qwen-2.5-1.5B}{Skywork-PRM-1.5B} & 61.1 & 52.0 & 56.4 & 54.2 & 64.8 & 64.9 & 63.3 & 66.5 & 64.9 & 57.5 & 63.3 & 91.1 & 70.7\\
\href{https://huggingface.co/Skywork/Skywork-o1-Open-PRM-Qwen-2.5-7B}{Skywork-PRM-7B} & 65.1 & \underline{56.4} & \textbf{62.8} & \textbf{59.6} & 69.4 & 67.1 & \underline{67.7} & 69.9 & 68.5 & \textbf{60.9} & 65.8 & 93.2 & 73.3\\
\href{https://huggingface.co/ScalableMath/llemma-7b-prm-prm800k-level-1to3-hf}{Llemma-PRM800k-7B} & 52.0 & 49.3 & 53.4 & 51.4 & 56.4 & 47.1 & 46.7 & 53.3 & 50.9 & 51.0 & 53.5 & 93.6 & 66.0\\
\href{https://huggingface.co/ScalableMath/llemma-7b-prm-metamath-level-1to3-hf}{Llemma-MetaMath-7B} & 50.5 & 50.2 & 50.5 & 50.3 & 51.9 & 47.6 & 44.4 & 52.1 & 49.0 & 50.5 & 51.3 & 96.0 & 66.0\\
\href{https://huggingface.co/ScalableMath/llemma-7b-oprm-prm800k-level-1to3-hf}{Llemma-oprm-7B} & 50.3 & 48.7 & 49.3 & 49.0 & 54.2 & 46.8 & 44.5 & 53.5 & 49.8 & 49.2 & 51.3 & 91.8 & 64.1\\
\href{https://github.com/KbsdJames/MATH-Minos}{MATHMinos-Mistral-7B} & 54.2 & 48.8 & 54.0 & 51.4 & 57.0 & 52.1 & 50.7 & 57.8 & 54.4 & 52.8 & 55.8 & 91.1 & 66.5\\
\href{https://huggingface.co/peiyi9979/math-shepherd-mistral-7b-prm}{MathShepherd-Mistral-7B} & 47.0 & 44.0 & 50.3 & 47.1 & 49.4 & 44.5 & 41.3 & 47.7 & 45.7 & 47.2 & 48.6 & 86.1 & 60.7\\
\href{https://huggingface.co/GAIR/ReasonEval-7B}{ReasonEval-7B} & 60.1 & \textbf{61.0} & 50.1 & \underline{55.6} & 62.1 & 65.9 & 61.5 & 66.0 & 63.9 & 55.7 & 58.0 & 99.5 & 71.1\\
\href{https://huggingface.co/GAIR/ReasonEval-34B}{ReasonEval-34B} & 60.5 & 54.8 & 48.1 & 51.5 & 66.4 & 60.3 & 57.8 & 67.5 & 63.0 & 57.7 & 64.3 & 97.2 & 73.1\\
\href{https://huggingface.co/RLHFlow/Llama3.1-8B-PRM-Mistral-Data}{RLHFlow-PRM-Mistral-8B} & 54.4 & 46.1 & 47.3 & 46.7 & 56.6 & 55.1 & 54.4 & 63.8 & 57.5 & 51.5 & 56.2 & 97.9 & 68.5\\
\href{https://huggingface.co/RLHFlow/Llama3.1-8B-PRM-Deepseek-Data}{RLHFlow-PRM-Deepseek-8B} & 54.2 & 46.4 & 48.9 & 47.6 & 55.7 & 55.0 & 53.2 & 66.2 & 57.5 & 49.0 & 55.4 & \textbf{99.8} & 68.1\\
\href{https://huggingface.co/Qwen/Qwen2.5-Math-PRM-7B}{Qwen2.5-Math-PRM-7B} & \underline{65.5} & 49.0 & 55.1 & 52.1 & \underline{71.8} & 67.3 & 66.3 & \underline{78.5} & \underline{71.0} & 57.6 & 69.1 & \underline{99.7} & 75.5\\
\href{https://huggingface.co/Qwen/Qwen2.5-Math-PRM-72B}{Qwen2.5-Math-PRM-72B} & \textbf{68.2} & 50.4 & \underline{58.8} & 54.6 & \textbf{73.7} & \textbf{71.1} & \textbf{72.2} & \textbf{78.6} & \textbf{73.9} & \underline{60.3} & \textbf{71.2} & 99.4 & \textbf{77.0}\\
\href{https://huggingface.co/jinachris/Qwen2.5-Math-7B-PRM800K}{Pure-PRM-7B} & 65.3 & 49.2 & 55.2 & 52.2 & 71.1 & \underline{68.8} & 64.0 & 76.9 & 70.2 & 60.3 & \underline{69.2} & 98.0 & \underline{75.8}\\
\cellcolor{gray!10} \textbf{Avg.} & \cellcolor{gray!10} 57.7 & \cellcolor{gray!10} 50.5 & \cellcolor{gray!10} 52.9 & \cellcolor{gray!10} 51.7 & \cellcolor{gray!10} 61.5 & \cellcolor{gray!10} 58.1 & \cellcolor{gray!10} 56.3 & \cellcolor{gray!10} 64.2 & \cellcolor{gray!10} 60.0 & \cellcolor{gray!10} 54.4 & \cellcolor{gray!10} 59.5 & \cellcolor{gray!10} 95.3 & \cellcolor{gray!10} 69.7 \\
\hline \multicolumn{14}{c}{\textit{\textbf{Open LLMs, Prompted as Critic Models}}} \\   \hline 
\href{https://huggingface.co/meta-math/MetaMath-7B-V1.0}{MetaMath-7B} & 49.7 & 48.9 & 46.9 & 47.9 & 47.3 & 48.9 & 48.4 & 48.8 & 48.3 & 46.5 & 48.3 & 98.0 & 64.2\\
\href{https://huggingface.co/meta-math/MetaMath-13B-V1.0}{MetaMath-13B} & 49.4 & 50.3 & 44.4 & 47.3 & 47.8 & 47.4 & 49.4 & 48.1 & 48.2 & 49.0 & 48.1 & 99.5 & 65.5\\
\href{https://huggingface.co/Qwen/Qwen2.5-Math-PRM-72B}{Qwen2.5-Math-72B} & 57.4 & 55.3 & 54.9 & 55.1 & 55.5 & \underline{71.6} & 58.1 & 59.1 & 61.1 & 47.4 & 53.8 & \textbf{100.0} & 67.1\\
\href{https://huggingface.co/Qwen/QwQ-32B-Preview}{QwQ-Preview-32B} & \underline{63.6} & \underline{57.2} & \underline{55.6} & \underline{56.4} & \underline{67.4} & \textbf{72.3} & \underline{66.2} & \underline{66.9} & \underline{68.2} & \underline{57.8} & \underline{62.7} & \textbf{100.0} & \underline{73.5}\\
\href{https://huggingface.co/deepseek-ai/DeepSeek-R1-Distill-Llama-70B}{R1-Distill-Llama3.1-70B} & 57.5 & 49.5 & 48.1 & 48.8 & 61.4 & 65.5 & 65.8 & 61.1 & 63.4 & 48.8 & 54.1 & 100.0 & 67.6\\
\href{https://huggingface.co/deepseek-ai/DeepSeek-R1-Distill-Qwen-7B}{R1-Distill-Qwen-7B} & 52.6 & 32.9 & 37.9 & 35.4 & 47.3 & 54.1 & 48.4 & 48.0 & 49.4 & 45.6 & 46.8 & \textbf{100.0} & 64.1\\
\href{https://github.com/deepseek-ai/DeepSeek-R1}{DeepSeek-R1}$^\dagger$ & \textbf{67.8} & \textbf{63.0} & \textbf{62.7} & \textbf{62.9} & \textbf{68.2} & 68.5 & \textbf{73.5} & \textbf{75.4} & \textbf{71.4} & \textbf{63.3} & \textbf{68.0} & \textbf{100.0} & \textbf{77.1}\\
\cellcolor{gray!10} \textbf{Avg.} & \cellcolor{gray!10} 56.8 & \cellcolor{gray!10} 51.0 & \cellcolor{gray!10} 50.1 & \cellcolor{gray!10} 50.5 & \cellcolor{gray!10} 56.4 & \cellcolor{gray!10} 61.2 & \cellcolor{gray!10} 58.5 & \cellcolor{gray!10} 58.2 & \cellcolor{gray!10} 58.6 & \cellcolor{gray!10} 51.2 & \cellcolor{gray!10} 54.5 & \cellcolor{gray!10} 99.6 & \cellcolor{gray!10} 68.5 \\
\hline \multicolumn{14}{c}{\textit{\textbf{Proprietary LLMs, Prompted as Critic Models}}} \\   \hline 
\href{https://openai.com/index/hello-gpt-4o/}{GPT-4o} & 66.8 & 57.0 & 62.4 & 59.7 & 72.0 & \underline{69.7} & 70.7 & 71.1 & 70.9 & \textbf{62.5} & 65.7 & 99.2 & \textbf{75.8}\\
\href{https://openai.com/index/openai-o1-mini-advancing-cost-efficient-reasoning/}{o1-mini}$^\dagger$ & \underline{68.8} & 65.6 & \underline{63.7} & \underline{64.6} & \textbf{74.5} & 67.7 & \textbf{73.8} & \textbf{72.3} & \textbf{72.1} & \underline{61.8} & 64.8 & \textbf{100.0} & \underline{75.5}\\
\href{https://deepmind.google/technologies/gemini/flash/}{Gemini-2.0-flash-exp} & 66.0 & \underline{67.2} & 58.1 & 62.7 & 70.4 & 65.7 & 66.0 & 67.3 & 67.3 & 61.8 & \textbf{66.2} & 98.2 & 75.4\\
\href{https://ai.google.dev/gemini-api/docs/thinking-mode}{Gemini-2.0-thinking-exp-1219} & \textbf{68.8} & \textbf{68.5} & \textbf{63.8} & \textbf{66.2} & \underline{72.9} & \textbf{71.3} & \underline{71.0} & \underline{71.8} & \underline{71.8} & 60.3 & \underline{65.7} & \underline{99.8} & 75.3\\
\cellcolor{gray!10} \textbf{Avg.} & \cellcolor{gray!10} 67.6 & \cellcolor{gray!10} 64.6 & \cellcolor{gray!10} 62.0 & \cellcolor{gray!10} 63.3 & \cellcolor{gray!10} 72.4 & \cellcolor{gray!10} 68.6 & \cellcolor{gray!10} 70.4 & \cellcolor{gray!10} 70.7 & \cellcolor{gray!10} 70.5 & \cellcolor{gray!10} 61.6 & \cellcolor{gray!10} 65.6 & \cellcolor{gray!10} 99.3 & \cellcolor{gray!10} 75.5 \\

\bottomrule[1.5pt]

\end{tabular}}
\caption{
Performances comparison of popular models on \textsc{PRMBench}. The best performance for each category and task is in \textbf{bold}, while the second-best performance is \underline{underlined}. $^\dagger$: To reduce costs, we evaluated only a subset of 394 samples for o1-mini and DeepSeek-R1.}
\vspace{-10pt}
\label{tab:main_results}
\end{table*}
\renewcommand{\arraystretch}{1} 

\subsection{Quality Control}
\label{sec:quality_control}
To ensure a high-quality dataset, we implement a series of steps to filter out unqualified data and maintain data integrity. The specific procedures are outlined below:

\paragraph{Feature Filtering} 
Our data curation procedure imposes strict structural requirements on the generated responses, where any outputs that do not satisfy these specifications cannot be considered valid for accurately assessing the performance of PRMs. However, even with detailed instructions, LLMs cannot consistently generate outputs that fully adhere to the required structure~\cite{openscholar, zeng2024evaluating,su2024conflictbank}. To maintain high data quality, we define stringent filtering rules to exclude instances that fail to meet the necessary structural criteria. Detailed structural requirements are provided in Appendix \ref{appendix_generation_prompts}, and the full description of our data generation process can be found in the supplementary materials.

\paragraph{Human Verification} Furthermore, to further ensure the quality of the data, we manually evaluate 10\% of the total instances. We focus on two key qualities for each data instance: \textbf{\ding{182} Correctness of modification:} Whether the modifications made to the data instance are correct and reasonable. \textbf{\ding{183} Difference in the modification:} Whether the modified data instance differs from the original. We recruited five volunteers to evaluate our proposed \textsc{PRMBench} and observe over 92\% qualification rate on the correctness metric and over 98\% qualification rate on the difference metric. The details of human annotation are provided in Appendix \ref{appendix:quality_control_setting}, and instructions for annotators are provided in Appendix \ref{appendix:annotator_instruction}.
This validation ensures the overall quality of our dataset and its suitability for studying process-level language reward models.

\section{Experiments}

\subsection{Models}
\label{sec:evaluate_models}
To provide a comprehensive evaluation of various models on \textsc{PRMBench}, we select a wide range of models, including open-source PRMs like QwenPRM \cite{zhang2025qwen_lessons} and RLHFlowPRMs \cite{xiong2024rlhflowmath}, as well as LLMs prompted as critic models, such as o1-mini \cite{openai_o1_2024} and DeepSeek R1 \cite{guo2025deepseekr1}. A complete list of these models can be found in Appendix \ref{appendix:models}. Additionally, we present the human evaluation results, with details available in Appendix \ref{appendix:human_performance_setting}.


All PRMs and LLMs are evaluated on the complete \textsc{PRMBench} dataset, except for o1-mini and DeepSeek-R1, which are evaluated on a subset of \textsc{PRMBench} comprising 394 samples, proportionally selected to reflect the class distribution, in order to reduce evaluation costs.

Considering the complexity of the task, which involves question comprehension, evaluation of the provided processes, and adherence to format constraints, few-shot demonstration setups are employed to help the model adapt to the output format through In-Context Learning (ICL) examples. 
Specifically, we use two-shot examples when prompting general-purpose LLMs. The impact of few-shot settings is discussed in Section \ref{fewshot_discuss}

\subsection{Evaluation Metrics}
\label{sec:evaluation_metrics}

Given our emphasis on evaluating the error detection capabilities, we use the negative F1 score as a metric for error detection performance. However, this metric may be affected by the inherent biases of models. To mitigate this and provide a unified, normalized score that reflects the overall competency of the evaluated model, following \citet{qwenprocessbench},
we introduce a metric called PRMScore, defined formally in Equation \ref{eq1}.

\begin{equation} \label{eq1}
\resizebox{0.9\columnwidth}{!}{$
\begin{aligned}
PRM\mbox{-}Score = w_1 * F1_{neg} + w_2 * F1
\end{aligned}
$}
\end{equation}

Where F1 and F1$_{neg}$ refer to F1 scores and negative F1 scores respectively. $w_1$ and $w_2$ are weights that are designed to maximize the differentiation between different models. The detailed evaluation procedure is provided in Appendix \ref{appendix:evaluation_procedure}. Besides, we also provide results of all evaluation categories in fine-grained metrics in Appendix \ref{appendix:detailed_results}.

\subsection{Main Results}
The main results are shown in Table \ref{tab:main_results}. Some observations can be summarized as follows:

\renewcommand{\arraystretch}{0.9} 
\begin{table}
\centering
\resizebox{\linewidth}{!}{
    \begin{tabular}{lcccc}
    \toprule
    \multirow{2}{*}{\textbf{Model}}  & \multicolumn{2}{c}{\textbf{Accuracy}} & \multirow{2}{*}{\tabincell{c}{\textbf{PRM}\\ \textbf{Score} }} &  \multirow{2}{*}{\textbf{Sim.}} \\
    \cline{2-3} 
    &Pos. & Neg. & \\
    
    \midrule
    ReasonEval-7B & 95.5 & 21.2 & 60.0 & 91.6 \\
    ReasonEval-34B & 79.1 & 48.4 & 60.5 & 82.8 \\
    Skywork-7B & 30.1 & 79.7 & 36.2 & 74.3\\
    RLHFlow-DeepSeek-8B & 95.0 & 13.0 & 54.2 & 95.0 \\
    GPT-4o & 82.9 & 58.2 & 66.8 & 76.6 \\
    Gemini-2-thinking & 89.0 & 49.8 & 68.8 & 82.0 \\
    \cellcolor{gray!10}Random &\cellcolor{gray!10}50.0 &\cellcolor{gray!10}50.0 &\cellcolor{gray!10}50.0 &\cellcolor{gray!10}79.4\\
    \bottomrule
    \end{tabular}
}
\caption{Comparison of model performance on positive and negative test cases, along with their similarities.}
\label{tab:bias}
\vspace{-10pt}
\end{table}

\paragraph{The PRMs partially grasp multi-step process evaluation}
Our analysis indicates that, although Gemini-2-Thinking achieves the highest performance among all evaluated models, its score is still significantly lower than human performance (68.8 vs. 83.8), highlighting substantial room for improvement in multi-step process evaluation.
Some models even perform worse than random guessing, highlighting their limited reliability and potential training biases. Notably, the best open-source PRMs fail to match the performance of general-purpose proprietary LLMs, which suggests that even specifically trained PRMs still lag behind leading general-purpose models. We provide a detailed error analysis in Appendix \ref{appendix:error_analysis}.

\paragraph{Simplicity is more challenging for PRMs}
Our analysis highlights significant variations in model reasoning capabilities across evaluation categories. For instance, in the Sensitivity category, ReasonEval-34B performs relatively well, achieving an average score of 73.1. Especially in the Multi-Solutions sub-category, it excels with a PRMScore of 97.2, approaching near-perfect classification accuracy. This suggests models perform relatively better on correct instance judgment. However, its performance declines markedly in more complex scenarios. In the \textbf{Simplicity} category, ReasonEval-34B's PRMScore drops to 51.5, suggesting partially reliable performance. 

Furthermore, to broaden the domain coverage of PRMBench, we additionally collect STEM-related data and construct \textsc{PRMBench-STEM}, which is designed to evaluate PRM performance across other domains. The data construction methodology and experimental results for PRMBench-STEM are provided in the appendix \ref{appendix:prmbench_stem}.

\begin{figure}
    \centering
    \includegraphics[width=1\linewidth]{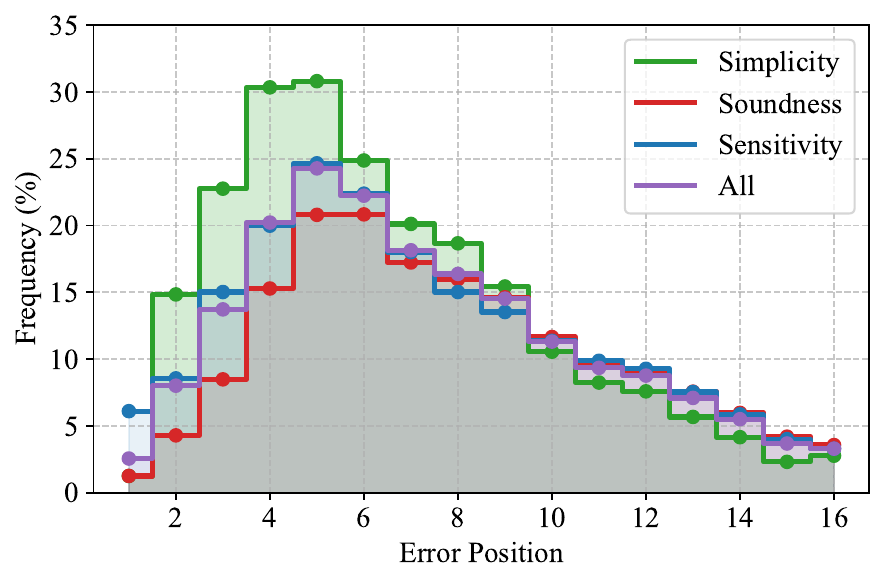}
    \caption{Distribution of error positions, truncated to 16 for better visualization, corresponding to the label field as shown in Figure \ref{fig:main_fig}.}
    \label{fig:error_position_distribution}
\end{figure}

\section{Detailed Analysis}

\subsection{Inference Bias within PRMs}
\label{sec:analyze_similarity}

\begin{tcolorbox}[ colback=white, colframe=black, width=\columnwidth, breakable,
boxsep=2pt,              
    left=4pt,               
    right=4pt,              
    top=4pt,              
    bottom=4pt,          
    boxrule=0.3mm  ]
\textbf{\textit{Takeaway 1.}} PRMs show a clear bias during evaluation, often favoring positive rewards.
\end{tcolorbox}

As shown in Table \ref{tab:main_results}, most open-source PRMs exhibit significant bias during evaluation, with some models performing worse than random guessing. suggesting the potential presence of bias within the inference procedure for our test cases. 
To validate this assumption, we compare the difference of models' performance on positive and negative instances. As shown in Table \ref{tab:bias},
 \textbf{some models exhibit a clear bias during evaluation, often favoring positive rewards.} For instance, ReasonEval-7B and RLHFlow-DeepSeek-8B achieve over 95\% accuracy on positive-labeled steps but only attain an average of 17\% accuracy on negative-labeled steps. Although proprietary LLMs outperform open-source PRMs, they also exhibit bias with a comparatively milder reward tendency.

Additionally, to further investigate inference bias, we evaluate the reward similarity of models' performance between completely correct reasoning processes and our test cases.
The solution-level similarity is defined as $S = 100 - |Acc_{pos} - Acc_{neg}|$, where $Acc$ denotes the average step accuracy within a solution. 
The results, shown in Table \ref{tab:bias}, reveal that certain models, such as ReasonEval-7B and RLHFlow-DeepSeek-8B, exhibit significantly higher similarity than the normal similarity score (79.4), showcasing potential limitations in differentiating positive and negative steps.

\begin{figure}
    \centering
    \includegraphics[width=1\linewidth]{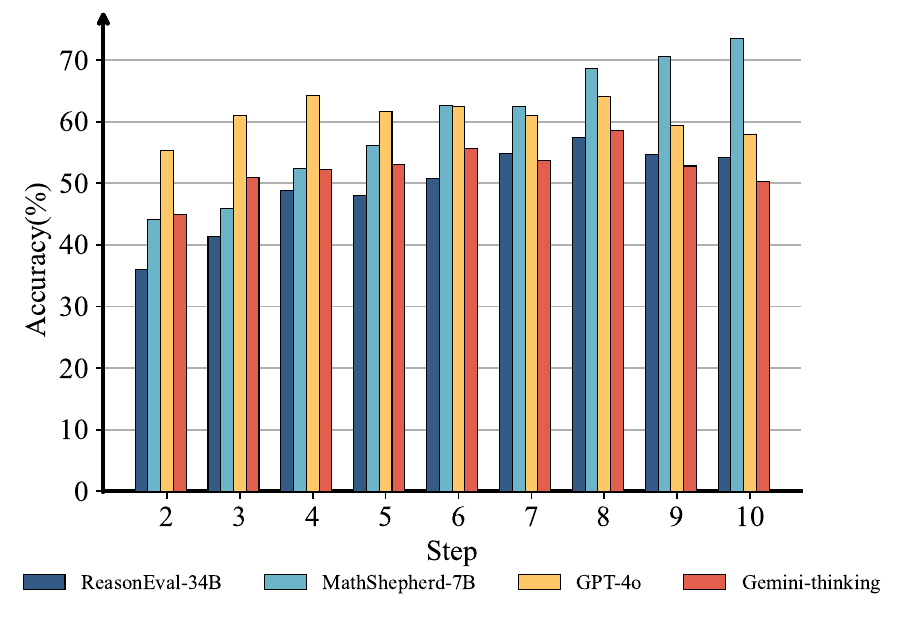}
    \caption{The models' error-detection accuracy across different error steps, where step 1 and steps beyond 11 are truncated for improved visualization.}
    \label{fig:step_acc}
\end{figure}
\renewcommand{\arraystretch}{0.8} 
\begin{table}
\centering
\resizebox{\linewidth}{!}{
    \begin{tabular}{lccc}
    \toprule
    \textbf{Model} & \textbf{0-shot} & \textbf{1-shot} & \textbf{2-shot} \\
    
    \midrule
    GPT-4o & 68.1 & 68.2 & 66.8 \\
    Gemini-2-flash & 65.3 & 64.9 & 66.0 \\
    Gemini-2-thinking & 67.8 & 67.8 & 68.8 \\
    \bottomrule
    \end{tabular}
}
\caption{The impact of ICL few-shot numbers on model performance. The number reported here is PRMScore. }
\label{tab:fewshot}
\end{table}

\subsection{Performance across Different Steps}

\begin{tcolorbox}[ colback=white, colframe=black, width=\columnwidth, breakable,
boxsep=2pt,             
    left=4pt,              
    right=4pt,             
    top=4pt,                
    bottom=4pt,             
    boxrule=0.3mm  ]
\textbf{\textit{Takeaway 2.}} PRMs show a gradual improvement in performance as the position of the steps increases.
\end{tcolorbox}

\textsc{PRMBench} includes a wide range of error step positions. The distribution of error positions is illustrated in Figure \ref{fig:error_position_distribution}. While differences exist across categories, the overall pattern remains consistent: all categories peak in frequency at step 5 and gradually decrease thereafter. This raises an interesting question: \textbf{Does the variation in step positions affect model performance?} To investigate, we focus on error steps to assess how erroneous step positions influence model accuracy. 
As depicted in Figure \ref{fig:step_acc}, proprietary LLMs maintain stable performance across different error step positions. In contrast, PRMs, including Math-Shepherd-7B and ReasonEval-7B, show a gradual improvement in performance as error step positions increase. 

\subsection{Impacts of ICL Settings}
\label{fewshot_discuss}
\begin{tcolorbox}[ colback=white, colframe=black, width=\columnwidth, breakable,
boxsep=2pt,             
    left=4pt,              
    right=4pt,              
    top=4pt,              
    bottom=4pt,             
    boxrule=0.3mm  ]
\textbf{\textit{Takeaway 3.}} In-context learning has subtle impact on models' performance on \textsc{PRMBench}.
\end{tcolorbox}

In this section, we investigate the impact of different ICL few-shot numbers on models' performance. We vary the number of ICL few-shots to 0, 1, and 2 to examine whether increasing the few-shot number enhances the performance of generative models prompted as critic models. 
As shown in  Table \ref{tab:fewshot}, for the Gemini-series models, a subtle improvement in performance is observed with a few-shot setup. However, for GPT-4o, no significant improvement is detected, and in some cases, a larger few-shot number even results in a decline in performance. These findings suggest that a few-shot approach exerts only a subtle impact on model performance on \textsc{PRMBench}.

\renewcommand{\arraystretch}{0.9} 
\begin{table}[tp]
\belowrulesep=0pt
\aboverulesep=0pt
\fontsize{14}{21}\selectfont
\centering

\resizebox{\columnwidth}{!}{
\begin{tabular}{l c c cc}
\toprule[1.5pt]
\textbf{Method} & \textbf{MATH} & \textbf{OlymBen} & \textbf{Avg.} & \textbf{PRMScore} \\
 \midrule 
\textbf{Pass@8} & 96.2 & 79.8 & 88.0 & - \\
\textbf{Maj@8} & 71.8 & 40.3 & 56.1 & - \\
 \midrule 
\href{https://huggingface.co/Skywork/Skywork-o1-Open-PRM-Qwen-2.5-7B}{SkyworkPRM-7B} & 90.0 & 60.1 & 75.1 & 65.1\\
\href{https://huggingface.co/ScalableMath/llemma-7b-prm-prm800k-level-1to3-hf}{LlemmaPRM-7B} & 87.4 & 58.3 & 72.8 & 52.0\\
\href{https://github.com/KbsdJames/MATH-Minos}{MATHMinos-7B} & 88.3 & 59.1 & 73.7 & 54.2\\
\href{https://huggingface.co/peiyi9979/math-shepherd-mistral-7b-prm}{MathShepherd-7B} & 88.6 & 60.0 & 74.3 & 47.0\\
\href{https://huggingface.co/GAIR/ReasonEval-7B}{ReasonEval-7B} & 87.0 & 58.4 & 72.7 & 60.1\\
\href{https://huggingface.co/RLHFlow/Llama3.1-8B-PRM-Deepseek-Data}{RLHFlowPRM-8B} & 87.6 & 58.5 & 73.0 & 54.2\\
\href{https://huggingface.co/Qwen/Qwen2.5-Math-PRM-7B}{Qwen2.5-PRM-7B} & 88.0 & 58.7 & 73.4 & 65.5\\
 \midrule 
\textbf{Standard Dev ($\sigma$)} & 0.91 & 0.71 & 0.81 & \textbf{6.40} \\
\textbf{Somers' D } & -0.05 & 0.05 & -0.05 & \textbf{1.00} \\

\bottomrule[1.5pt]

\end{tabular}}
\caption{
Performance comparison on Best-of-8 using different PRMs. $\sigma$ represents the standard deviation of model performances across all benchmarks. Somers’ D refers to the Somers’ D correlation between PRMScore and specific benchmarks.
}
\label{tab:correlation}
\vspace{-10pt}
\end{table}

\subsection{Comparison between BoN Evaluation and PRMBench}
\label{fewshot_discuss}
\begin{tcolorbox}[ colback=white, colframe=black, width=\columnwidth, breakable,
boxsep=2pt,             
    left=4pt,              
    right=4pt,              
    top=4pt,              
    bottom=4pt,             
    boxrule=0.3mm  ]
\textbf{\textit{Takeaway 4.}} PRMs struggle with detecting false positives, exposing the potential for reward hacking.
\end{tcolorbox}

We compare the results between our PRMBench and Best-of-N (BoN) evaluation to observe the correlation. Following \citet{zhang2025qwen_lessons,yang2024qwen25report}, we sampled eight responses (i.e., N=8) from Qwen-QwQ across multiple mathematical benchmarks, including GSM8K~\cite{cobbe2021gsm8k}, MATH~\cite{hendrycks2020math_benchmark}, Olympiad Bench~\cite{he2024olympiadbench} and MMLU \cite{hendrycks2020MMLU}. During evaluation, the PRMs are tasked with assigning a validity score to each step within every candidate response. The overall score for each candidate response is calculated by multiplying the individual step scores, as outlined in \citet{lets_verify_stepbystep}. We also provide majority voting as an baseline and pass@8 as the upper bound. The experiment setting and full BoN evaluation results are shown in Appendix \ref{appendix:bon_details}.

\textbf{Although PRMs excel at selecting correct outcomes, they struggle with step-level reward hacking.} As shown in Table \ref{tab:correlation}, the average Somers’ D correlation between PRMBench and BoN is only -0.05, highlighting the inconsistency between step-level and outcome-level evaluation. For instance, Math-Shepherd-7B achieves a PRMScore of 47.0 with 51.3\% accuracy in false-positive scenarios within PRMBench, but outperforms most PRMs, including the state-of-the-art QwenPRM-7B, in the BoN evaluation (74.3 vs. 73.4). This inconsistency reveals that PRMs are suboptimal at detecting step-level errors and false positives, exposing potential reward hacking~\citep{gao2025on}. Compared to BoN, \textsc{PRMBench} provides a better distinction between models, with a higher standard deviation (6.40 vs. 0.81), indicating its greater sensitivity to fine-grained differences in reasoning steps. 
Furthermore, based on the discoveries from PRMBench, we provide further discussions including several promising directions for future exploration, which are included in the appendix \ref{appendix:further_discussion}.


\section{Conclusion}

In this paper, we investigate a crucial question: \textbf{Can existing PRMs detect various types of erroneous reasoning steps and provide reasonable rewards?} To address this, we introduce \textsc{PRMBench}, a benchmark characterized by its fine-grained evaluation categories and challenging requirements. We carefully curate 6,216 data samples with 83,456 step-level labels through LLMs and human filtering. \textsc{PRMBench} can be used to evaluate different process-labeling models, ensuring its general applicability.
Through a comprehensive evaluation of existing PRMs and generative LLMs prompted as critic models, we can observe that PRMs exhibit partial capability in multi-step process evaluation, showcasing significant room for improvement.
Furthermore, we highlight the critical need for detecting detailed error types and conducting comprehensive evaluations of PRMs. Despite these advances, enhancing the reward accuracy of PRMs and improving models' reasoning abilities remain open research challenges. We encourage future work to leverage and expand upon \textsc{PRMBench} to address these issues.

\section{Acknowledgement}

We gratefully acknowledge the support and resources provided by the Shanghai Artificial Intelligence Laboratory, which were essential to the successful completion of this research. We are also grateful to Zhiyi Li at The Hong Kong University of Science and Technology for offering valuable examples and inspiration for this study.

\section{Limitations}
There are still some limitations in our work, which are summarized below:
\begin{itemize}[leftmargin=*]
\setlength{\itemsep}{0pt}
    \item While the PRMBench is large and comprehensive, comprising 6,216 samples and 83,456 step-level labels, a larger dataset could provide more robust evaluation and training opportunities. As the data construction method is flexible and data-agnostic, it can be adapted to more different data sources. We will continuously expand our dataset and explore PRM training in future versions.

    \item We evaluate the error-detection capabilities in terms of accuracy. However, a more detailed analysis, including the activation of the model’s neurons and hidden states~\cite{zhang2023multimodal}, would offer a deeper insight into improvements for PRMs. This limitation is not unique to our study and is common in most evaluations of Large Language Models.
    \item \textsc{PRMBench} currently focuses on textual reasoning. In future work, we will further explore its generalization to multimodal reasoning processes, such as those involving the integration of text and images~\cite{su2025openthinkimg, xia2024mmedrag}, to assess error detection capabilities in cross-modal scenarios.



\end{itemize}

\bibliography{main}

\clearpage
\appendix

\renewcommand{\arraystretch}{1}

\section{Detailed Information for \textsc{PRMBench}}
\subsection{Evaluation Categories}
\label{appendix:evaluation_subjects}
\begin{figure*}[h]
    \centering
    \includegraphics[width=1\linewidth]{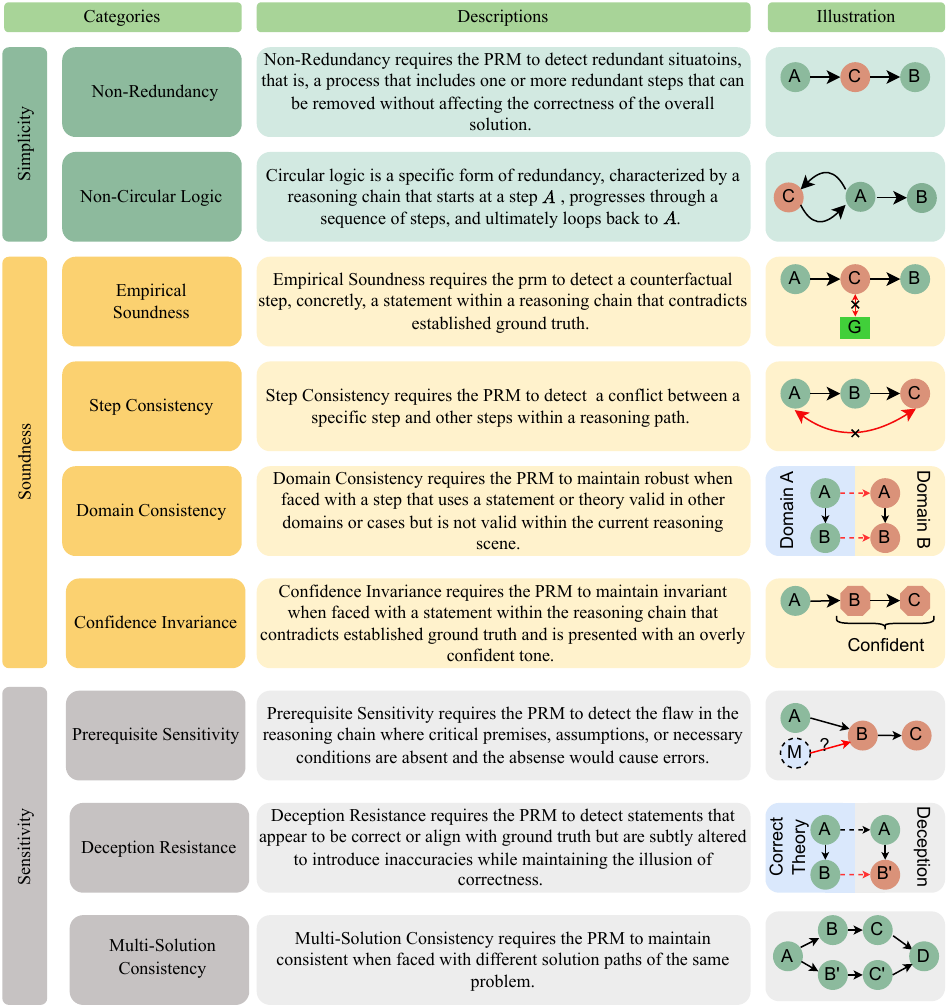}
    \caption{Examples of an in-depth evaluation of PRMBench.}
    \label{apdxfig:detailed_classifications}
\end{figure*}

In this section, We provide detailed information on our evaluation categories. The hierarchical categories, corresponding descriptions, and illustrations are shown in Figure \ref{apdxfig:detailed_classifications}. We carefully curated 6,216 data samples and 83,456 step-level labels. The benchmark spreads across three main evaluation categories: \textit{simplicity}, \textit{soundness}, and \textit{sensitivity}. Among them, Simplicity comprises two sub-categories: non-redundancy and non-circular Logic. Soundness includes four main sub-categories: empirical soundness, step consistency, domain consistency, and confidence invariance. Finally, Sensitivity mainly evaluates models in three main parts: prerequisite sensitivity, deception resistance, and multi-solution consistency. The detailed descriptions and illustrations of each sub-category are shown in Figure \ref{apdxfig:detailed_classifications}.

\label{sec:appendix}

\subsubsection{Simplicity}

Specifically, the Simplicity evaluation category is divided into two sub-categories: Non-Redundancy and Non-Circular Logic, with detailed descriptions provided below:

\paragraph{Non-Redundancy} requires PRM to detect the redundancy within the reasoning procedure. The redundancy situation refers to a process that is not the most concise or efficient, as it includes one or more redundant steps that can be removed without affecting the correctness of the overall solution path. 
For example, as shown in Figure \ref{apdxfig:detailed_classifications}, if $ A \to B $ represents a correct inference chain, the redundant reasoning procedure can be displayed as $ A \to C \to B $. where $C$ represents one or more redundant steps $ C = \{c | c \text{ is redundant} \} $. 

\paragraph{Non-Circular Logic} In this sub-category, PRMs are required to detect the potential circular logic within the reasoning process. Circular logic is a specific form of redundancy, distinct from general redundancy, in that it finally loops back to a previous reasoning step. For example, as shown in Figure \ref{fig:main_fig}, if $A \to B$ represents a correct inference chain, circular logic can be formulated as $A \to C \to A \to B$, where the reasoning starts at step $A$, progresses through a sequence of steps, and ultimately loops back to $A$. We list Non-Circular Logic separately due to its common occurrence in reasoning processes.


\begin{table*}[t]
  \centering
  \vspace{1mm}
  
  \resizebox{\textwidth}{!}{
    \begin{tabular}{l cccc cc}
    \toprule

& \textbf{Step Descriptions} &\textbf{GT}& \textbf{ReasonE} & \textbf{MathS} & \textbf{GPT-4o} & \textbf{Gemini} \\
\midrule

\textbf{Question} & \tabincell{c}{Compute $\left( 1 + \cos \frac {\pi}{8} \right) \left( 1 + \cos  \frac {3 \pi}{8} \right) \left( 1 + \cos \frac {5 \pi}{8} \right)  \left( 1 + \cos \frac {7 \pi}{8} \right).$}& -& -& -& -& -  \\
\textbf{Step 1} & \tabincell{c}{Let's call the expression we're trying to compute $x$.} & \textcolor{mygreen}{\ding{51}}& \textcolor{mygreen}{89.1}& \textcolor{mygreen}{87.5}& \textcolor{mygreen}{100.0}& \textcolor{mygreen}{100.0}\\
\textbf{Step 2} & \tabincell{c}{Use the fact that $\cos(\pi-a)=-\cos a$.} & \textcolor{mygreen}{\ding{51}}& \textcolor{mygreen}{77.5}& \textcolor{mygreen}{82.8}& \textcolor{mygreen}{100.0}& \textcolor{mygreen}{100.0}\\
\textbf{Step 3} & \tabincell{c}{Then we have $x=\left( 1 + \cos \frac {\pi}{8} \right) \left( 1  + \cos \frac {3 \pi}{8} \right) \left( 1 + \cos \frac {5 \pi}{8} \right) \left( 1 + \cos \frac {7 \pi}{8} \right)$.} & \textcolor{mygreen}{\ding{51}}& \textcolor{mygreen}{71.4}& \textcolor{mygreen}{76.7}& \textcolor{mygreen}{100.0}& \textcolor{mygreen}{100.0}\\
\textbf{Step 4} & \tabincell{c}{This equals $\left( 1 + \cos \frac {\pi}{8} \right) \left( 1 +  \cos \frac {3 \pi}{8} \right) \left( 1 - \cos \frac {\pi}{8} \right)  \left( 1 - \cos \frac {3 \pi}{8} \right)$.} & \textcolor{mygreen}{\ding{51}}& \textcolor{myred}{20.5}& \textcolor{mygreen}{76.9}& \textcolor{mygreen}{100.0}& \textcolor{mygreen}{100.0}\\
\textbf{Step 5} & \tabincell{c}{By the difference of squares, this equals $\left(1-\cos^2 \frac{\pi}{8}\right)\left(1-\cos^2 \frac{3\pi}{8}\right)$.} & \textcolor{mygreen}{\ding{51}}& \textcolor{mygreen}{56.0}& \textcolor{mygreen}{83.7}& \textcolor{mygreen}{100.0}& \textcolor{mygreen}{100.0}\\
\textbf{Step 6} & \tabincell{c}{And by the Pythagorean identity, this equals $\left(\sin^2 \frac{\pi}{8}\right)\left(\sin^2 \frac{3\pi}{8}\right)$.} & \textcolor{mygreen}{\ding{51}}& \textcolor{mygreen}{63.5}& \textcolor{mygreen}{80.4}& \textcolor{mygreen}{100.0}& \textcolor{mygreen}{100.0}\\
\textbf{Step 7} & \tabincell{c}{Ok, now we just need to compute the sines.} & \textcolor{mygreen}{\ding{51}}& \textcolor{mygreen}{84.1}& \textcolor{mygreen}{79.1}& \textcolor{mygreen}{50.0}& \textcolor{mygreen}{50.0}\\
\textbf{Step 8} & \tabincell{c}{We can use the formula for the sum of sines.} & \textcolor{myred}{\ding{55}}& \textcolor{mygreen}{73.6}& \textcolor{mygreen}{74.1}& \textcolor{myred}{0.0}& \textcolor{myred}{0.0}\\
\textbf{Step 9} & \tabincell{c}{That formula is $\sin(a)+\sin(b) = 2\sin\left(\frac{a+b}{2}\right)\cos\left(\frac{a-b}{2}\right)$.} & \textcolor{myred}{\ding{55}}& \textcolor{mygreen}{79.1}& \textcolor{mygreen}{78.5}& \textcolor{myred}{0.0}& \textcolor{mygreen}{100.0}\\
\textbf{Step 10} & \tabincell{c}{That means that $\sin\frac{\pi}{8} + \sin\frac{3\pi}{8} = 2\sin\left(\frac{\pi/8+3\pi/8}{2}\right)\cos\left(\frac{\pi/8-3\pi/8}{2}\right)$.} & \textcolor{myred}{\ding{55}}& \textcolor{mygreen}{67.4}& \textcolor{mygreen}{78.0}& \textcolor{myred}{0.0}& \textcolor{myred}{0.0}\\
\textbf{Step 11} & \tabincell{c}{This equals $2\sin\frac{\pi/2}{2}\cos(-\frac{\pi/4}{2})$.} & \textcolor{myred}{\ding{55}}& \textcolor{mygreen}{68.1}& \textcolor{mygreen}{72.1}& \textcolor{myred}{0.0}& \textcolor{mygreen}{100.0}\\
\textbf{Step 12} & \tabincell{c}{This equals $2\cdot 1 \cdot \frac{\sqrt{2}}{2}$.} & \textcolor{myred}{\ding{55}}& \textcolor{myred}{49.2}& \textcolor{mygreen}{73.8}& \textcolor{myred}{0.0}& \textcolor{mygreen}{100.0}\\
\textbf{Step 13} & \tabincell{c}{This equals $\sqrt{2}$.} & \textcolor{myred}{\ding{55}}& \textcolor{myred}{35.1}& \textcolor{mygreen}{73.8}& \textcolor{myred}{0.0}& \textcolor{mygreen}{100.0}\\
\textbf{Step 14} & \tabincell{c}{But this step doesn't help us compute the original product of sines.} & \textcolor{myred}{\ding{55}}& \textcolor{mygreen}{72.5}& \textcolor{mygreen}{64.3}& \textcolor{myred}{-50.0}& \textcolor{mygreen}{100.0}\\
\textbf{Step 15} & \tabincell{c}{So, $\left(\sin^2 \frac{\pi}{8}\right)\left(\sin^2 \frac{3\pi}{8}\right)$ remains the same and $x=\frac{1}{8}$.} & \textcolor{myred}{\ding{55}}& \textcolor{myred}{6.3}& \textcolor{myred}{35.8}& \textcolor{myred}{0.0}& \textcolor{myred}{-100.0}\\
\textbf{Step 16} & \tabincell{c}{Therefore, the mistake didn't change the value of $x$.} & \textcolor{mygreen}{\ding{51}}& \textcolor{myred}{22.6}& \textcolor{myred}{43.5}& \textcolor{myred}{-100.0}& \textcolor{myred}{-100.0}\\

\hline
\cellcolor{gray!10}\textbf{Final Acc.} &\cellcolor{gray!10}-& \cellcolor{gray!10}100  &\cellcolor{gray!10}56.2& \cellcolor{gray!10}50.0& \cellcolor{gray!10}93.8& \cellcolor{gray!10}62.5 \\
\hline
\cellcolor{gray!10}\textbf{Reason} & \multicolumn{6}{c}{\cellcolor{gray!10}\tabincell{c}{A counterfactual step was introduced in steps 8 through 13 by mistakenly using the formula \\for the sum of sines instead of the product of sines. This leads to incorrect intermediate calculations. \\However, due to fortunate errors, the end result ironically matches the correct answer in step 15.}} \\

    \bottomrule
    \end{tabular}
  }
  \caption{An example of a data instance and error cases from \textsc{PRMBench}. The numbers reported are step-level validity scores generated by models. Scores and labels in \textcolor{myred}{red} indicate negative samples, while those in \textcolor{mygreen}{green} indicate positive samples. ``GT'' represents ground truth, while ``ReasonE,'' ``MathS,'' and ``Gemini'' correspond to ReasonEval-7B, Math-Shepherd-7B, and Gemini-2.0-flash-thinking-exp, respectively.}
  \label{tab:error_cases}
\end{table*}
\begin{table}[t]
\centering
\resizebox{\linewidth}{!}{
    \begin{tabular}{lcc}
    \toprule
    \textbf{Abbr.} & \textbf{Full Name} & \textbf{Evaluation Category} \\
    
    \midrule
    NR. & Non-Redundancy & Simplicity \\
    NCL. & Non-Circular Logic & Simplicity \\
    ES. & Empirical Soundness & Soundness \\
    SC. & Step Consistency & Soundness \\
    DC. & Domain Consistency & Soundness \\
    CI. & Confidence Invariance & Soundness \\
    PS. & Prerequisite Sensitivity & Sensitivity \\
    DR. & Deception Resistance & Sensitivity \\
    MS. & Multi-Solution Consistency & Sensitivity \\
    \bottomrule
    \end{tabular}
}
\caption{The impact of ICL few-shot numbers on models' final performance. The number reported here is PRMScore. }
\label{apdxtab:abbrs}
\end{table}
\subsubsection{Soundness}
We divide the Soundness category into four sub-categories due to its complexity: Empirically Soundness, Step Consistency, Domain Consistency, and Confidence Invariance. The definition of each sub-category is discussed below.

\paragraph{Empirically Soundness} demands PRM to detect the implicit counterfactual mistakes within the reasoning process. A counterfactual step refers to a statement within a reasoning chain that contradicts established ground truth $G$. Such contradictions can arise from relying on outdated theories, omitting critical constraints in theory, or incorporating erroneous assumptions. 

\paragraph{Step Consistency} expects PRM to detect the implicit step-wise contradiction, which means a conflict between a specific step and other steps within a reasoning path. Given a reasoning path $ P = \{S_1, S_2, \dots, S_n\} $, a step contradiction exists if $ S_i \perp S_j $, where $ i, j \in [1, n] $ and $ i \neq j $. 

\paragraph{Domain Consistency} Under this circumstance, PRMs are required to detect potential domain inconsistency mistakes, where domain inconsistency is a special type of counterfactual. It refers to a step within the reasoning chain that uses a statement or theory valid in other domains or cases but is not valid within the current reasoning chain. 

\paragraph{Confidence Invariance} demands the PRM to detect over-confident hallucinations, a type of counterfactual where an incorrect statement is made with unwarranted certainty, contradicting established ground truth.

\subsubsection{Sensitivity}


This category includes three sub-categories: Prerequisite Sensitivity, Deception Resistance, and Multi-Solution Consistency, with detailed descriptions provided below.

\paragraph{Prerequisite Sensitivity} requires the PRM to maintain sensitivity to missing conditions or prerequisite mistakes, which means a flaw in the reasoning chain where critical premises, assumptions, or necessary conditions are absent. This omission results in logical gaps, incomplete reasoning, or biased conclusions. For example, when a missing condition occurs, the model is required to solve the problem through case analysis or further investigation. However, the answer becomes incorrect if the model overlooks the missing condition and proceeds with standard reasoning methods.

\paragraph{Deception Resistancy} demands the PRM to detect the implicit deception or trap within a reasoning process, that is, statements that appear to be correct or align with ground truth but are subtly altered to introduce inaccuracies while maintaining the illusion of correctness. 

\paragraph{Multi-Solution Consistency} expects the PRM to maintain consistency when faced with different solution paths of the same problem. Concretely, to evaluate the sensitivity and the generalizability of PRMs, we utilize multiple correct reasoning processes of the same question to test whether the PRM can perform correctly.

\subsection{Examples For Different Evaluation Categories}
\label{appendix_category_examples}
\begin{figure*}[h]
    \centering
    \includegraphics[width=1\linewidth]{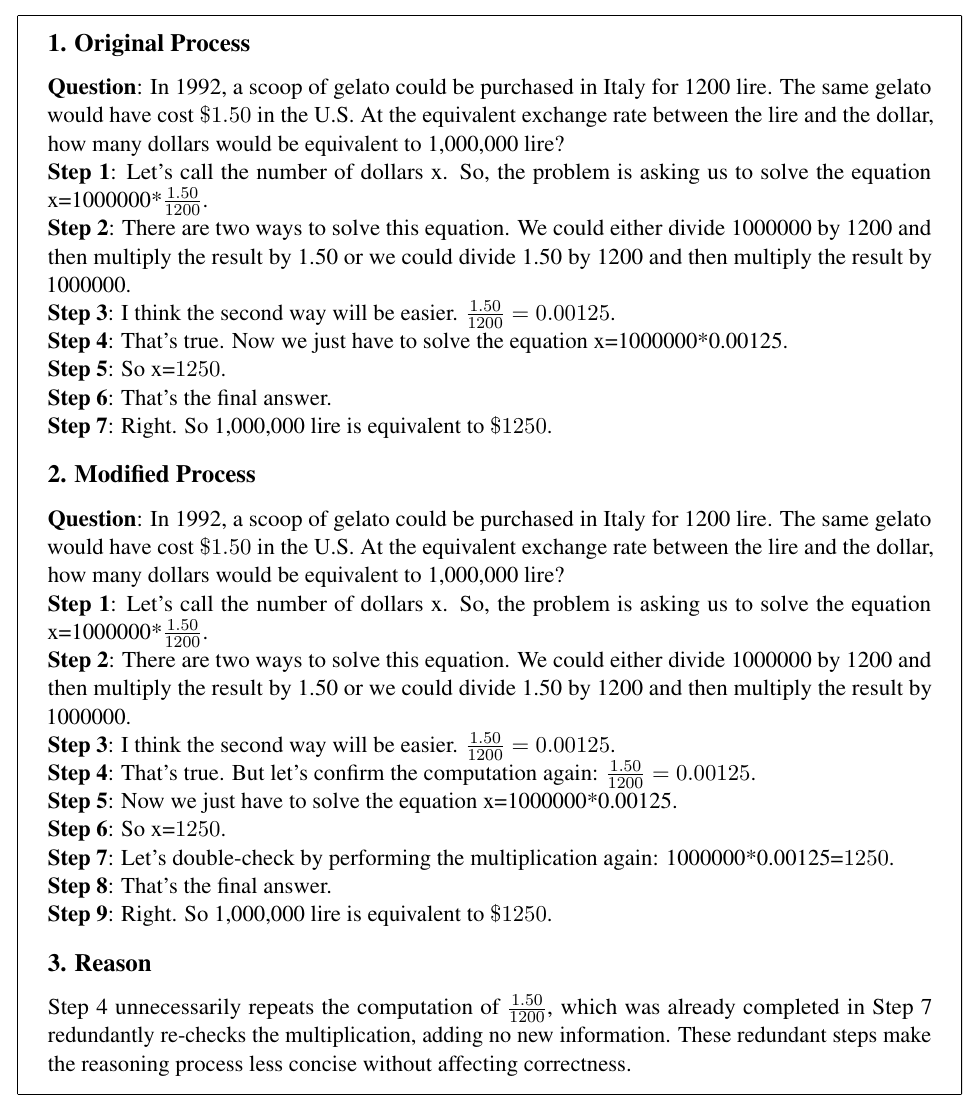}
    \caption{An example data instance for \textbf{NR.} category.}
    \label{apdxfig:example:nr}
\end{figure*}

\begin{figure*}[h]
    \centering
    \includegraphics[width=1\linewidth]{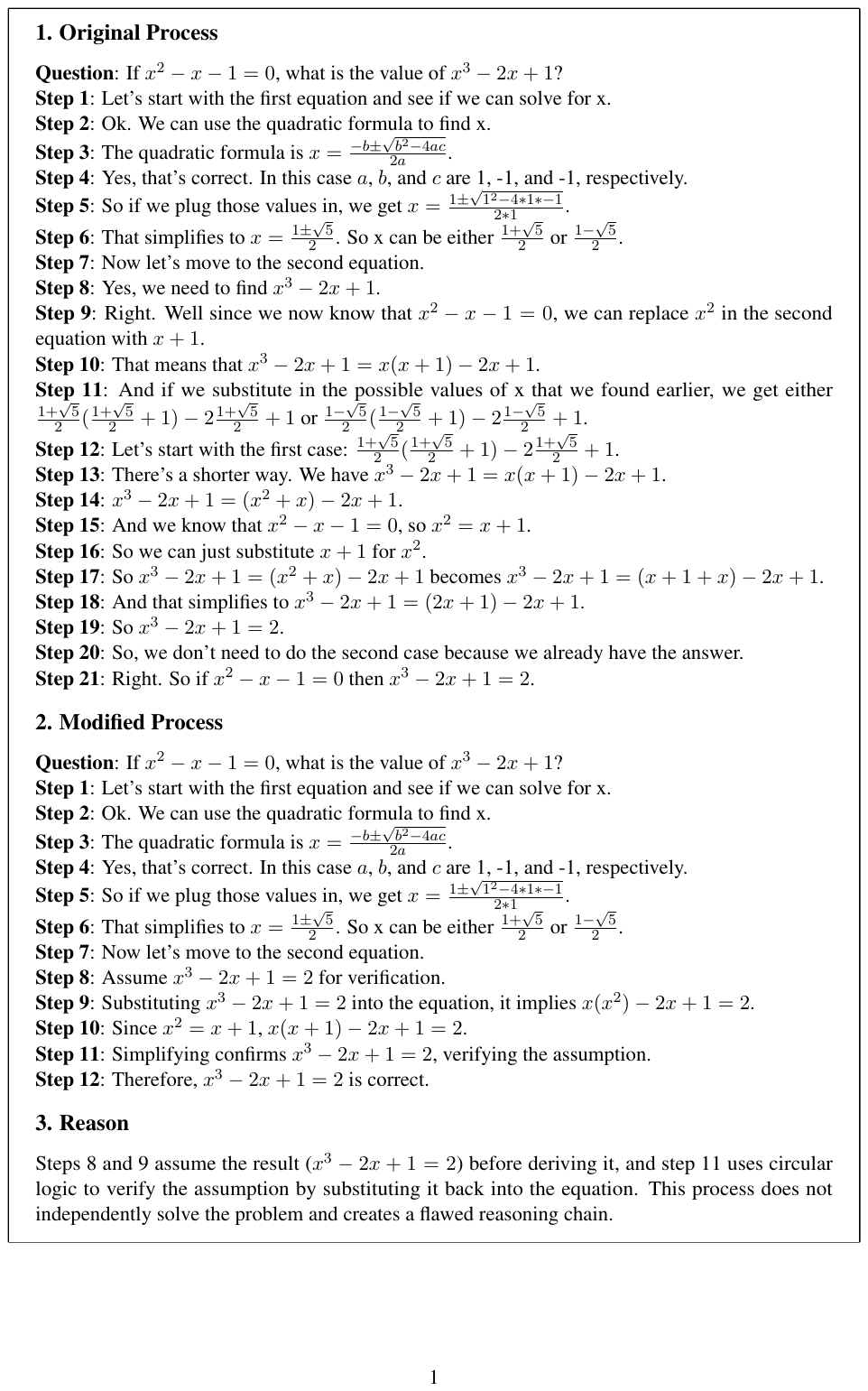}
    \caption{An example data instance for \textbf{NCL.} category.}
    \label{apdxfig:example:ncl}
\end{figure*}

\begin{figure*}[h]
    \centering
    \includegraphics[width=1\linewidth]{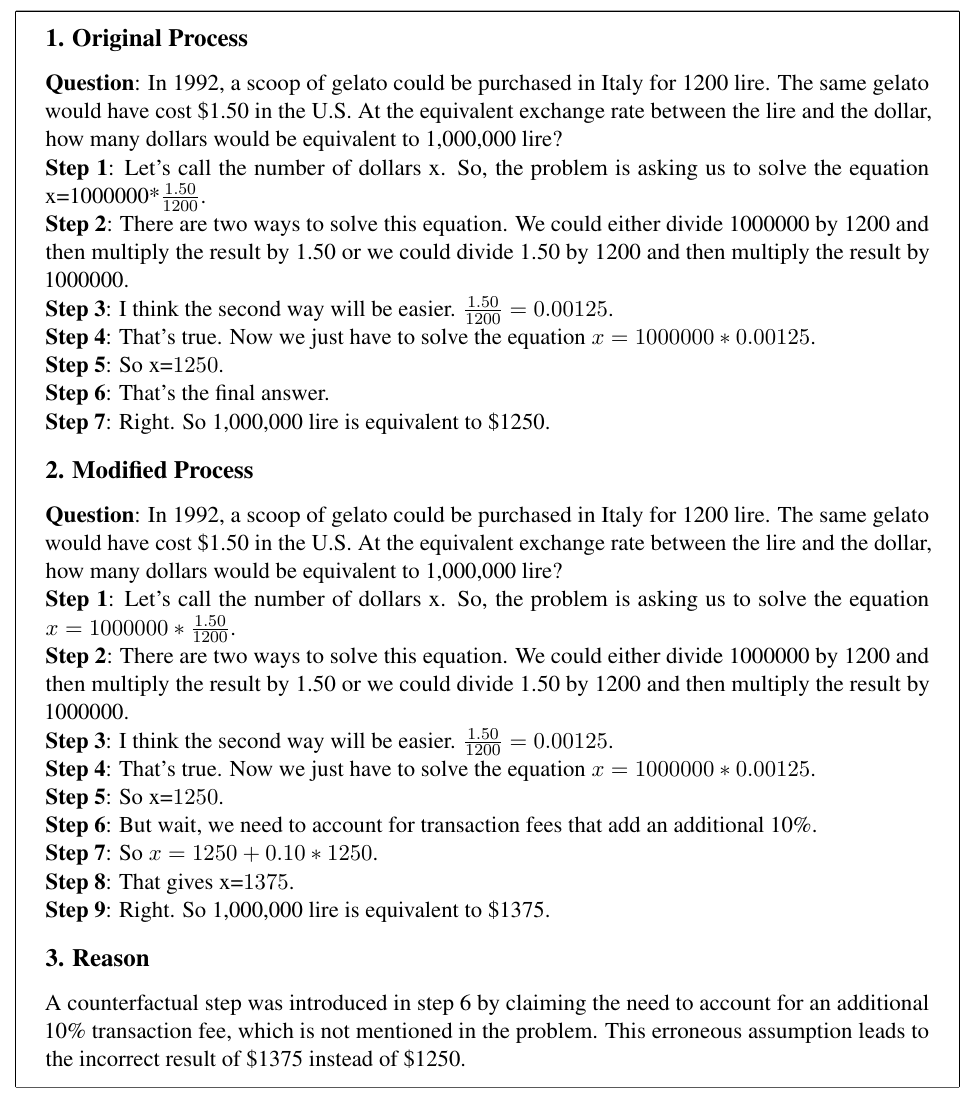}
    \caption{An example data instance for \textbf{ES.} category}
    \label{apdxfig:example:es}
\end{figure*}

\begin{figure*}[h]
    \centering
    \includegraphics[width=1\linewidth]{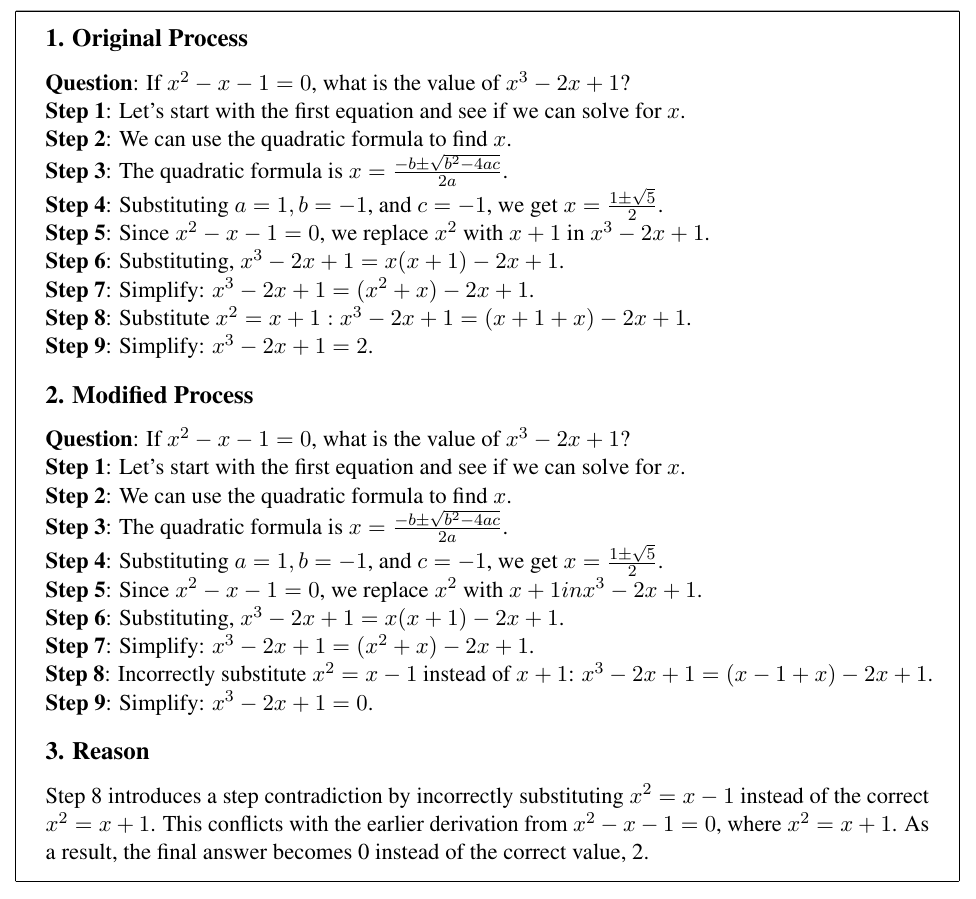}
    \caption{An example data instance for \textbf{SC.} category.}
    \label{apdxfig:example:sc}
\end{figure*}

\begin{figure*}[h]
    \centering
    \includegraphics[width=1\linewidth]{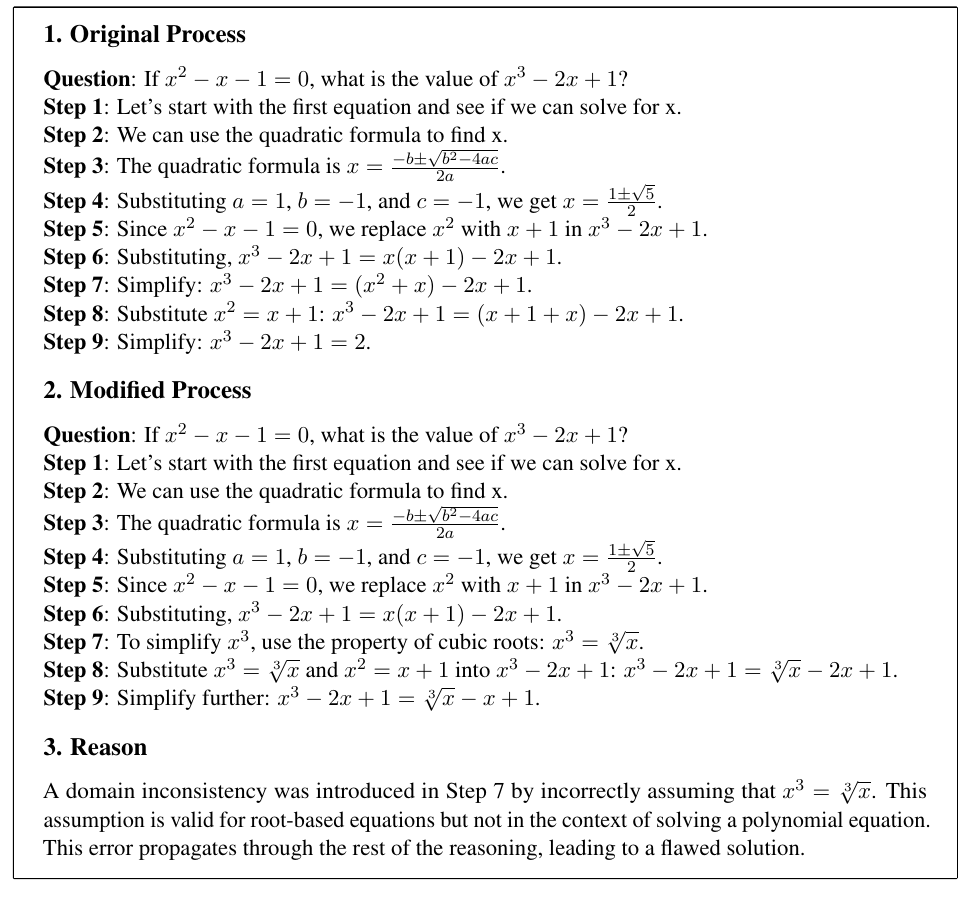}
    \caption{An example data instance for \textbf{DC.} category.}
    \label{apdxfig:example:dc}
\end{figure*}

\begin{figure*}[h]
    \centering
    \includegraphics[width=1\linewidth]{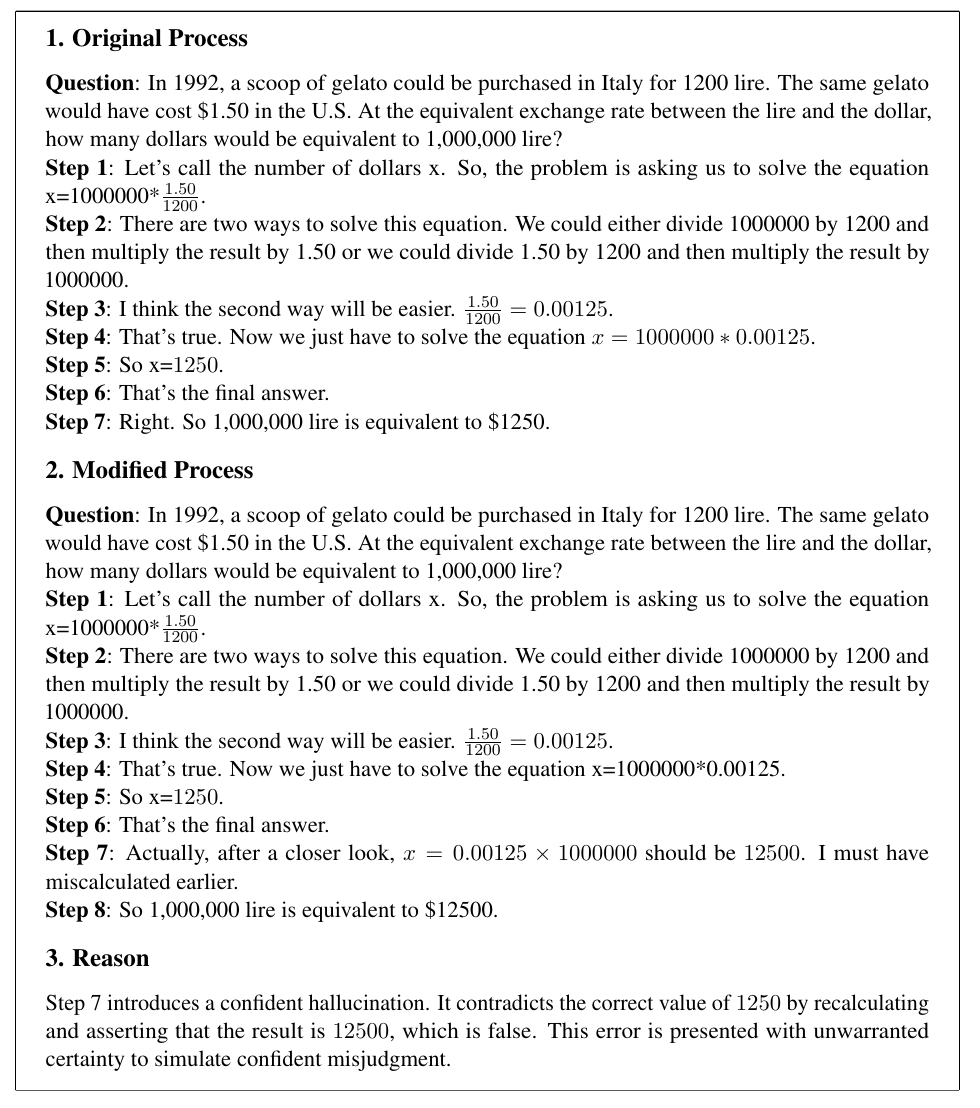}
    \caption{An example data instance for \textbf{CI.} category.}
    \label{apdxfig:example:ci}
\end{figure*}

\begin{figure*}[h]
    \centering
    \includegraphics[width=1\linewidth]{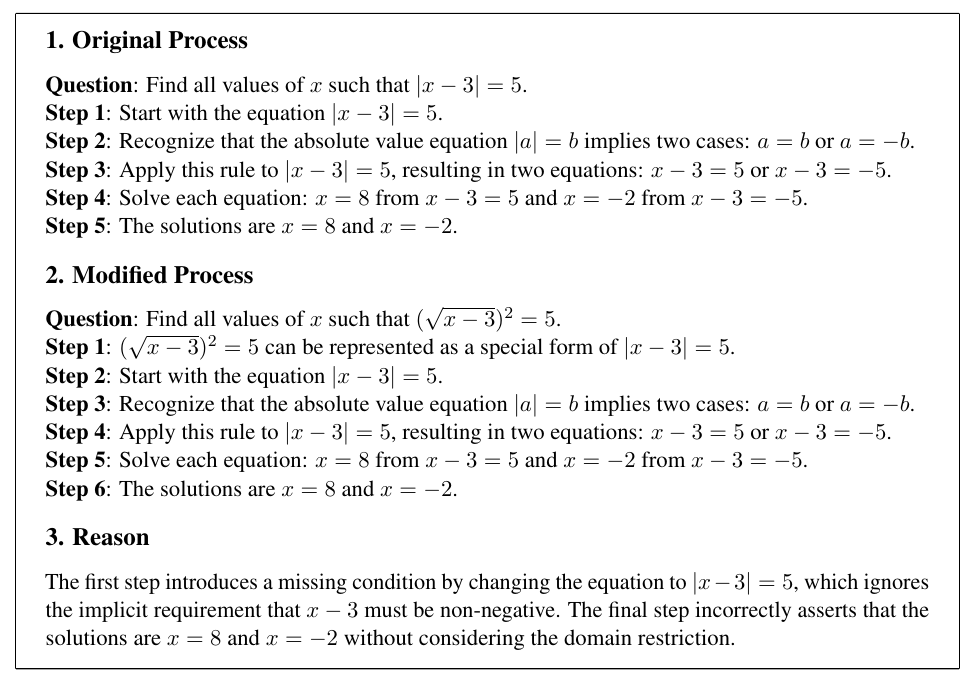}
    \caption{An example data instance for \textbf{PS.} category.}
    \label{apdxfig:example:ps}
\end{figure*}

\begin{figure*}[h]
    \centering
    \includegraphics[width=1\linewidth]{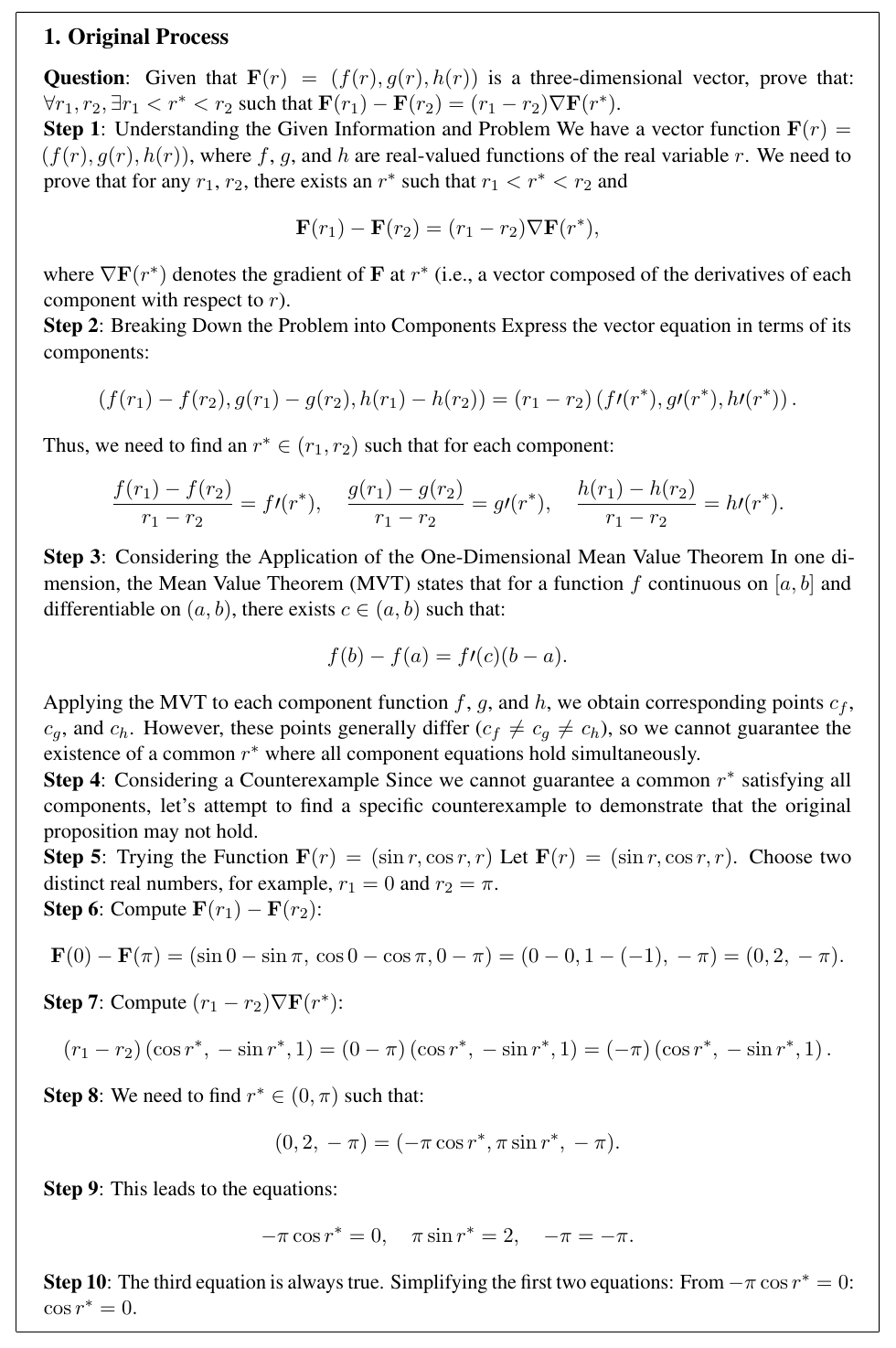}
    \caption{An example data instance for \textbf{DR.} category (1).}
    \label{apdxfig:example:dr1}
\end{figure*}

\begin{figure*}[h]
    \centering
    \includegraphics[width=1\linewidth]{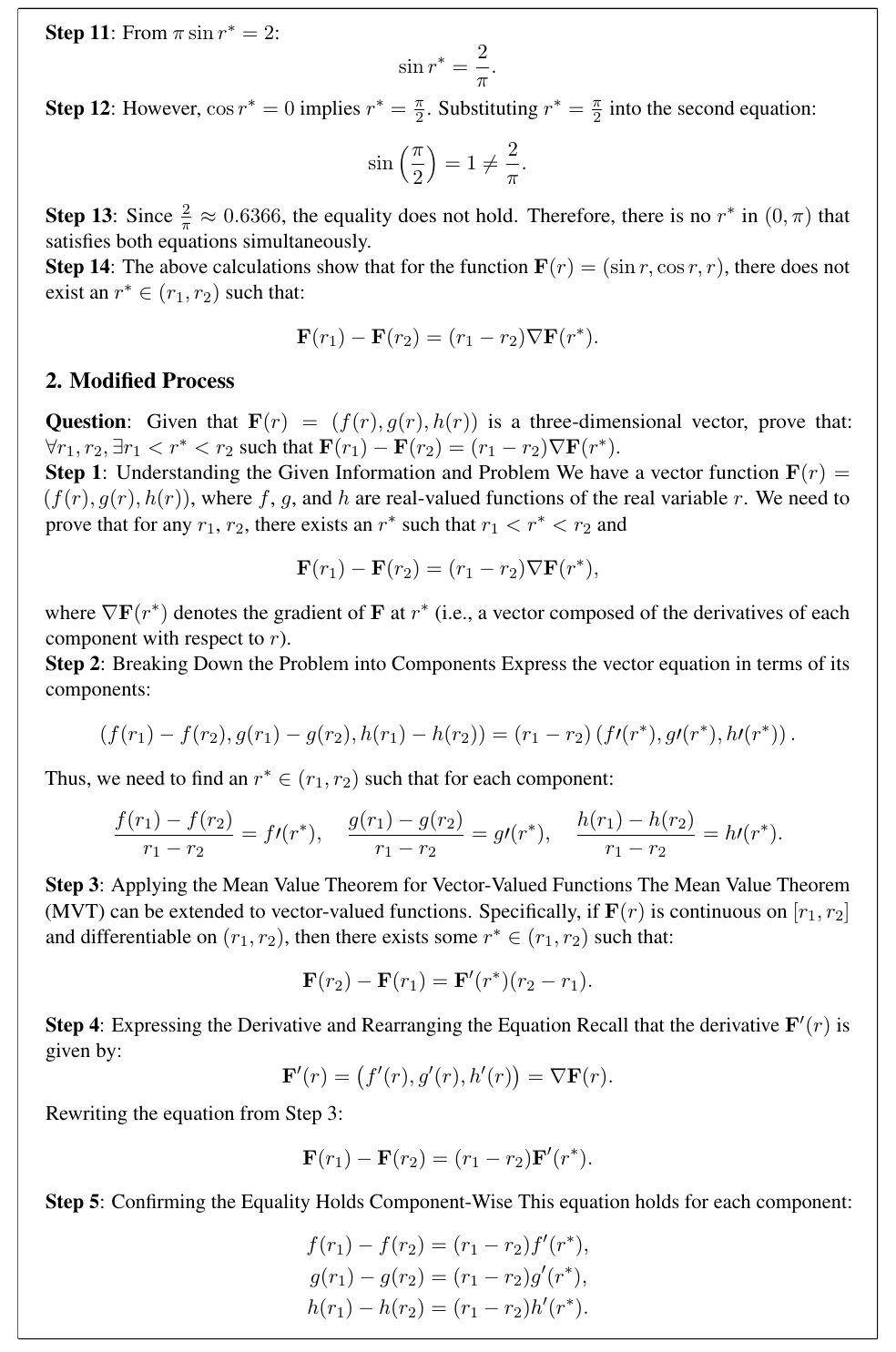}
    \caption{An example data instance for \textbf{DR.} category (2).}
    \label{apdxfig:example:dr2}
\end{figure*}

\begin{figure*}[h]
    \centering
    \includegraphics[width=1\linewidth]{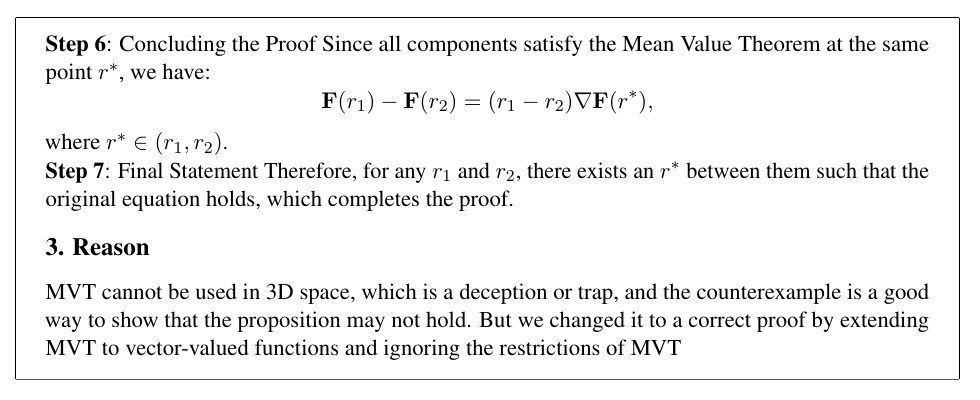}
    \caption{An example data instance for \textbf{DR.} category (3).}
    \label{apdxfig:example:dr3}
\end{figure*}

\begin{figure*}[h]
    \centering
    \includegraphics[width=0.9\linewidth]{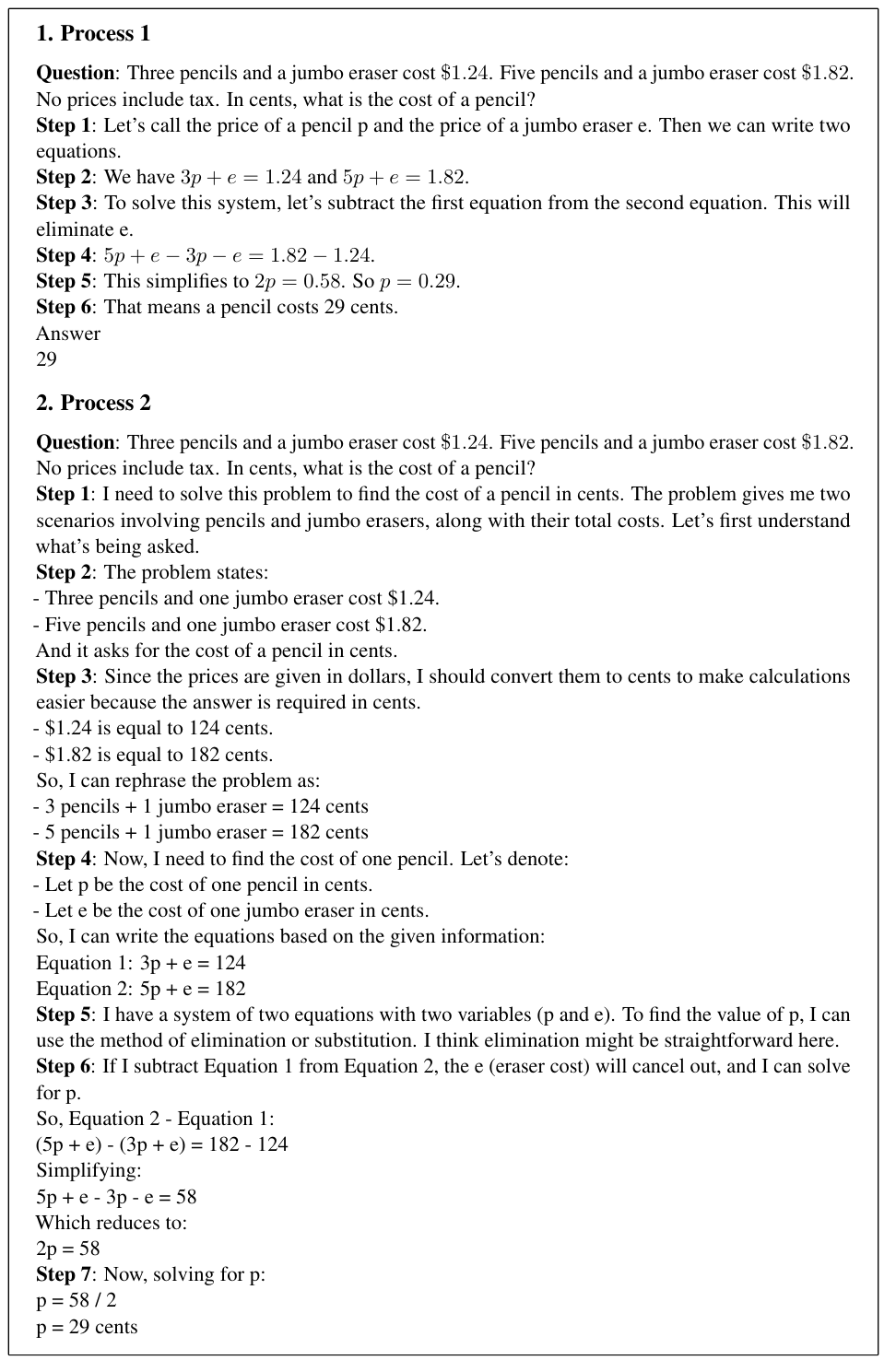}
    \caption{An example data instance for \textbf{MS.} category (1).}
    \label{apdxfig:example:ms1}
\end{figure*}

\begin{figure*}[h]
    \centering
    \includegraphics[width=0.9\linewidth]{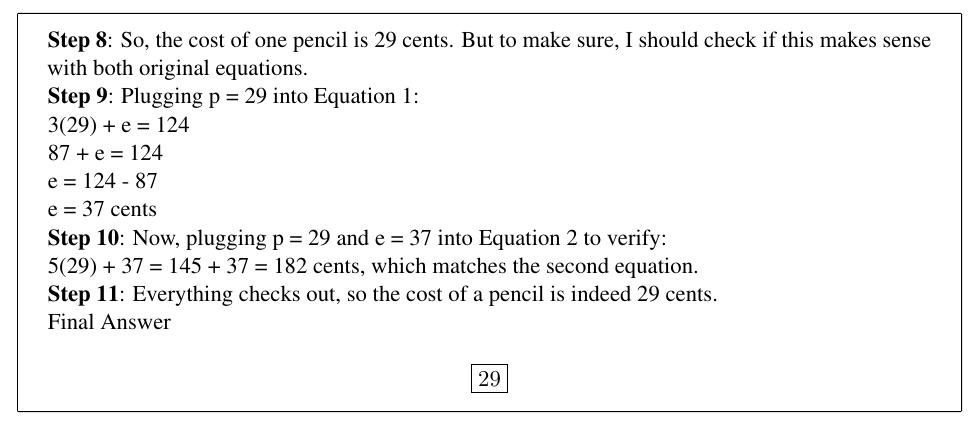}
    \caption{An example data instance for \textbf{MS.} category (2).}
    \label{apdxfig:example:ms2}
\end{figure*}

In this section, we provide detailed examples of the various evaluation categories and their corresponding sub-categories. The data instance examples are displayed in Figure \ref{apdxfig:example:nr}-\ref{apdxfig:example:ms2}.
All datasets used in this work are publicly available and have been released by their original creators, who are responsible for ensuring privacy protection. These datasets are utilized in accordance with their respective licenses and intended purposes, without introducing any harmful or sensitive information.

\subsection{Human Annotation Settings}
\begin{figure*}[h]
    \centering
    \includegraphics[width=1\linewidth]{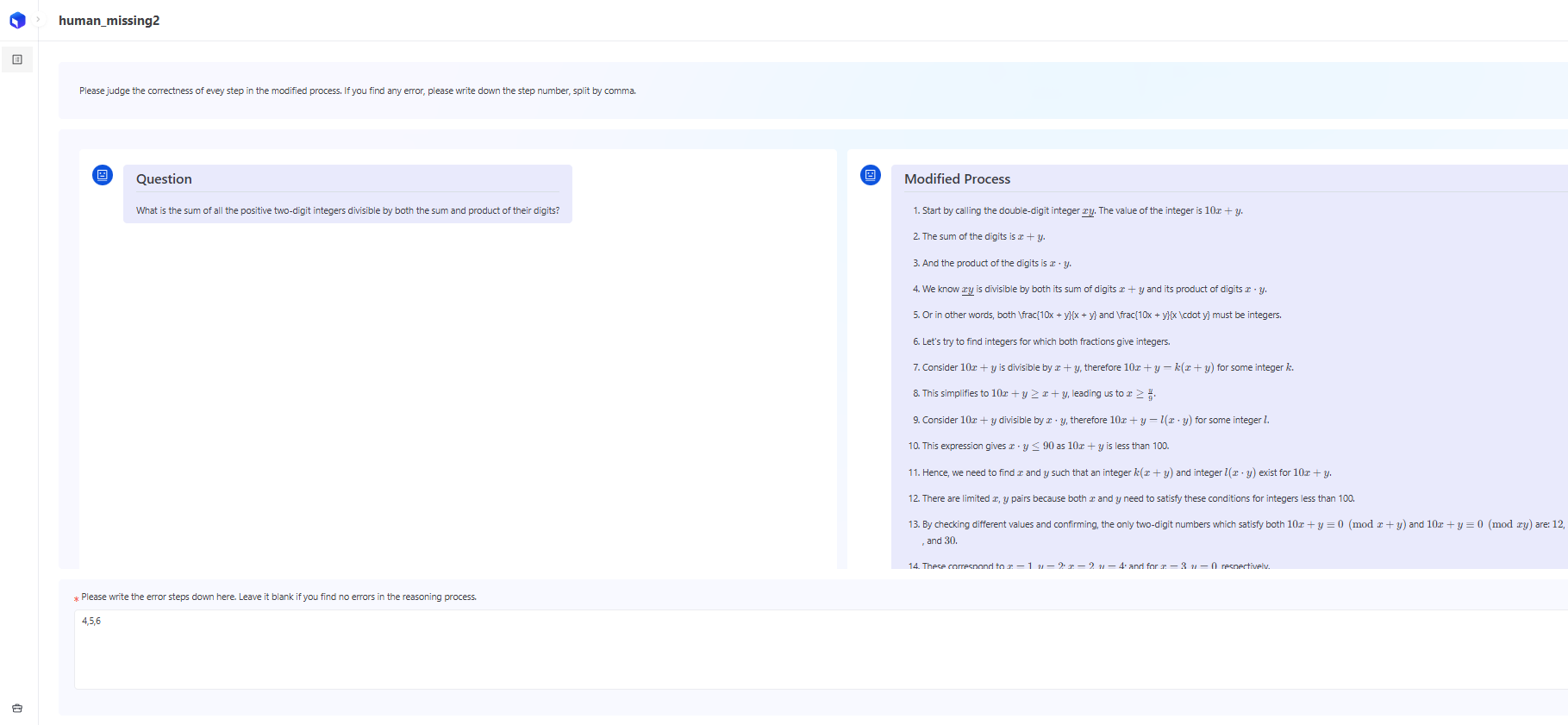}
    \caption{The data annotation platform.}
    \label{apdxfig:annotation_platform}
\end{figure*}
\label{appendix:human_annotation_setting}
\subsubsection{Quality Control}
\label{appendix:quality_control_setting}
In the second stage of the quality control process, we recruited five volunteers, each holding a bachelor’s degree or an equivalent qualification, to assess the correctness and validity of \textsc{PRMBench}. To facilitate high-quality labeling, we utilize LabelLLM~\cite{he2024labelllm} to help the data annotation procedure, as shown in Figure \ref{apdxfig:annotation_platform}. The annotators' instructions are shown in Appendix \ref{appendix:annotator_instruction}.

\subsubsection{Human Performance Evaluation}
\label{appendix:human_performance_setting}
For human performance evaluation, we recruited three volunteers, each holding a bachelor’s degree or an equivalent qualification, to assist with data annotation. Following \citet{hao2025emma}, we randomly selected 50 instances from each sub-category, resulting in a mini-test set of 450 samples. Each annotator was responsible for three subsets, and the results are presented in Table \ref{tab:main_results}.

\section{Detailed Experiment Results}

\subsection{Abbreviation Of Sub-Categories}
\label{appendix:abbr}
The full names of abbreviations used in our tables are shown in Table \ref{apdxtab:abbrs}.

\subsection{Models}
\label{appendix:models}
To provide a comprehensive evaluation of various models on \textsc{PRMBench}, we select a diverse set of models, including both open-source PRMs and different types of LLMs configured as critic models. Specifically, the open-source PRMs include Skywork-PRM-1.5B/7B~\cite{skyworkprms}, LlemmaPRMs~\cite{llemmaprm}, MathMinosPRM~\cite{mathminos}, MathShepherd-Mistral-7B~\cite{wang2023math_shepherd}, ReasonEval-7B/34B~\cite{reasoneval}, Pure-PRM-7B \cite{cheng2025pure} and Qwen-PRM-7B/72B~\cite{zhang2025qwen_lessons}. Additionally, we evaluate state-of-the-art general-purpose LLMs, including the open-source MetaMath-7B/34B~\cite{yu2023metamath}, Qwen2.5-Math-72B~\cite{yang2024qwen25math}, DeepSeek-R1 and its distill series~\cite{guo2025deepseekr1}, as well as closed-source LLMs such as GPT-4o \cite{openai2024gpt4o}, Gemini-2.0-flash~\cite{deepmind_gemini_2_flash}, and multi-step reasoning-enhanced LLMs like the o1 series models \cite{openai_o1_2024} and Gemini-2-Thinking \cite{deepmind_gemini_2_flash}.

All models used in this work are publicly available and have been released by their original authors, who are responsible for ensuring privacy protection. These models are utilized in accordance with their respective licenses and intended purposes, without any modifications.

\subsection{Evaluation Procedure}
\label{appendix:evaluation_procedure}
For each annotated question-solution pair, the reward models are tasked with evaluating the correctness and redundancy of each step, assigning a step-level validity score and a step-level redundancy score to each step. We subsequently utilize the specified threshold of each model to obtain the prediction indicating whether the step is correct or redundant. This task is therefore framed as a binary classification problem. Thus we can utilize the evaluation metric defined in Section \ref{sec:evaluation_metrics} to evaluate the performance of models on \textsc{PRMBench}.

\subsection{Detailed Results of \textsc{PRMBench}}
\label{appendix:detailed_results}
In addition to PRMScore displayed in Table \ref{tab:main_results}, we also list the full results with different metrics across different sub-categories here. The detailed evaluation results are shown in Table \ref{apdxtab:fullres_all}-\ref{apdxtab:fullres_dr}.

\subsection{Detailed Results of \textsc{PRMBench-STEM}}
\label{appendix:prmbench_stem}
We extend our benchmark by collecting additional data from various scientific domains, including physics, chemistry, and biology, and construct \textsc{PRMBench-STEM} using a similar data curation methodology as described in the main paper. The statistics of \textsc{PRMBench-STEM} are presented in Table \ref{apdxtab:statisticsprmbenchstem}. We then evaluate several representative PRMs on \textsc{PRMBench-STEM}, and the results are summarized in Tables \ref{apdxtab:stem_bio}, \ref{apdxtab:stem_chem}, and \ref{apdxtab:stem_physics} for the Biology, Chemistry, and Physics subsets, respectively. As shown, PRMs exhibit weaker performance on \textsc{PRMBench-STEM} compared to their performance on the original PRMBench. Notably, the Simplicity category remains the most challenging, consistent with our earlier observations.

\begin{table}
\centering
\resizebox{\linewidth}{!}{
    \begin{tabular}{lccc}
    \toprule
    \textbf{Domain} & \textbf{\# of Instances} & \textbf{Avg. Step Num} & \textbf{Avg. Error Num} \\
    
    \midrule
    \textbf{Physics} & 1619 & 7.32 & 2.74 \\
    \textbf{Chemistry} & 1543 & 7.74 & 2.78 \\
    \textbf{Biology} & 2342 & 6.76 & 2.49 \\
    \bottomrule
    \end{tabular}
}
\caption{Statistics of \textsc{PRMBench-STEM}. }
\label{apdxtab:statisticsprmbenchstem}
\vspace{-10pt}
\end{table}

\begin{table*}[h]
\belowrulesep=0pt
\aboverulesep=0pt
\fontsize{14}{21}\selectfont
\centering

\resizebox{\textwidth}{!}{
\begin{tabular}{l c|ccc| ccccc| cc c}
\toprule[1.5pt]
\multirow{2}{*}{\textbf{Model}} & \multirow{2}{*}{\textbf{Overall}} & \multicolumn{3}{c|}{\textbf{Simplicity}}  & \multicolumn{5}{c|}{\textbf{Soundness}}& \multicolumn{3}{c}{\textbf{Sensitivity}}\\
\cmidrule(lr){3-5} \cmidrule(lr){6-10} \cmidrule(lr){11-13} 
&& \textbf{NR.} & \textbf{NCL.} & \textbf{Avg.} &\textbf{ES} &\textbf{SC.}&\textbf{DC.} &\textbf{CI} & \textbf{Avg.} &\textbf{PS} & \textbf{DR.} & \textbf{Avg.}   \\
 \midrule

\hline \multicolumn{13}{c}{\textit{\textbf{Open-source Process Level Reward Models}}} \\   \hline 
\href{https://huggingface.co/Skywork/Skywork-o1-Open-PRM-Qwen-2.5-1.5B}{Skywork-PRM-1.5B} & 47.3 & 46.7 & 41.6 & 44.2 & 45.0 & 54.1 & 55.3 & 49.6 & 51.0 & 43.2 & 41.7 & 42.4\\
\href{https://huggingface.co/Skywork/Skywork-o1-Open-PRM-Qwen-2.5-7B}{Skywork-PRM-7B} & 67.4 & 54.6 & 53.6 & 54.1 & \underline{71.6} & \underline{76.2} & \textbf{77.9} & 74.4 & \underline{75.0} & \underline{64.0} & 65.3 & 64.6\\
\href{https://huggingface.co/ScalableMath/llemma-7b-prm-prm800k-level-1to3-hf}{Llemma-PRM800k-7B} & 52.7 & 49.1 & 52.4 & 50.7 & 52.3 & 49.4 & 53.9 & 55.1 & 52.7 & 51.1 & 52.6 & 51.9\\
\href{https://huggingface.co/ScalableMath/llemma-7b-prm-metamath-level-1to3-hf}{Llemma-MetaMath-7B} & 41.5 & 43.7 & 39.1 & 41.4 & 39.3 & 44.8 & 44.3 & 42.0 & 42.6 & 38.5 & 38.2 & 38.4\\
\href{https://huggingface.co/ScalableMath/llemma-7b-oprm-prm800k-level-1to3-hf}{Llemma-oprm-7B} & 50.7 & 44.4 & 37.3 & 40.9 & 54.5 & 50.3 & 53.0 & 56.8 & 53.6 & 51.0 & 53.6 & 52.3\\
\href{https://github.com/KbsdJames/MATH-Minos}{MATHMinos-Mistral-7B} & 47.1 & 43.0 & 37.9 & 40.4 & 48.0 & 51.6 & 47.1 & 53.6 & 50.1 & 45.1 & 45.3 & 45.2\\
\href{https://huggingface.co/peiyi9979/math-shepherd-mistral-7b-prm}{MathShepherd-Mistral-7B} & 53.1 & 47.2 & 43.1 & 45.1 & 52.3 & 62.1 & 63.5 & 57.5 & 58.9 & 50.0 & 46.4 & 48.2\\
\href{https://huggingface.co/GAIR/ReasonEval-7B}{ReasonEval-7B} & \underline{67.5} & \textbf{74.5} & \textbf{62.9} & \textbf{68.7} & 63.5 & 74.5 & 71.8 & 71.8 & 70.4 & 59.9 & 59.3 & 59.6\\
\href{https://huggingface.co/GAIR/ReasonEval-34B}{ReasonEval-34B} & 60.7 & \underline{68.2} & \underline{61.9} & \underline{65.0} & 61.0 & 53.8 & 52.0 & 59.9 & 56.7 & 57.8 & 62.6 & 60.2\\
\href{https://huggingface.co/RLHFlow/Llama3.1-8B-PRM-Mistral-Data}{RLHFlow-PRM-Mistral-8B} & 45.5 & 43.0 & 37.4 & 40.2 & 44.5 & 53.1 & 49.0 & 50.3 & 49.2 & 42.6 & 42.2 & 42.4\\
\href{https://huggingface.co/RLHFlow/Llama3.1-8B-PRM-Deepseek-Data}{RLHFlow-PRM-Deepseek-8B} & 47.0 & 42.9 & 37.4 & 40.1 & 44.7 & 59.0 & 57.7 & 52.6 & 53.5 & 41.6 & 41.6 & 41.6\\
\href{https://huggingface.co/Qwen/Qwen2.5-Math-PRM-7B}{Qwen2.5-Math-PRM-7B} & 65.8 & 42.9 & 49.0 & 46.0 & 69.6 & 75.1 & 72.5 & \underline{75.7} & 73.2 & 63.9 & \underline{65.4} & \underline{64.7}\\
\href{https://huggingface.co/Qwen/Qwen2.5-Math-PRM-72B}{Qwen2.5-Math-PRM-72B} & \textbf{73.7} & 42.8 & 50.6 & 46.7 & \textbf{82.3} & \textbf{79.6} & \underline{77.6} & \textbf{82.9} & \textbf{80.6} & \textbf{75.5} & \textbf{76.6} & \textbf{76.1}\\
\cellcolor{gray!10} \textbf{Avg.} & \cellcolor{gray!10} 55.4 & \cellcolor{gray!10} 49.5 & \cellcolor{gray!10} 46.5 & \cellcolor{gray!10} 48.0 & \cellcolor{gray!10} 56.1 & \cellcolor{gray!10} 60.3 & \cellcolor{gray!10} 59.7 & \cellcolor{gray!10} 60.2 & \cellcolor{gray!10} 59.0 & \cellcolor{gray!10} 52.6 & \cellcolor{gray!10} 53.1 & \cellcolor{gray!10} 52.9 \\
\hline \multicolumn{13}{c}{\textit{\textbf{Open LLMs, Prompted as Critic Models}}} \\   \hline 
\href{https://huggingface.co/meta-math/MetaMath-7B-V1.0}{MetaMath-7B} & 45.1 & 45.3 & 35.8 & 40.6 & 45.8 & 48.8 & 48.8 & 47.1 & 47.6 & 43.6 & 43.7 & 43.6\\
\href{https://huggingface.co/meta-math/MetaMath-13B-V1.0}{MetaMath-13B} & 43.6 & 44.7 & 35.8 & 40.2 & 42.2 & 46.3 & 47.1 & 45.0 & 45.2 & 42.4 & 41.1 & 41.7\\
\href{https://huggingface.co/Qwen/Qwen2.5-Math-PRM-72B}{Qwen2.5-Math-72B} & 42.3 & 49.5 & 35.6 & 42.5 & 38.0 & 56.1 & 45.7 & 42.8 & 45.6 & 37.0 & 35.6 & 36.3\\
\href{https://huggingface.co/Qwen/QwQ-32B-Preview}{QwQ-Preview-32B} & \textbf{54.5} & \textbf{54.8} & \textbf{43.3} & \textbf{49.0} & \textbf{53.6} & \textbf{63.9} & \underline{64.4} & \textbf{58.5} & \underline{60.1} & \textbf{51.8} & \textbf{48.3} & \textbf{50.1}\\
\href{https://huggingface.co/deepseek-ai/DeepSeek-R1-Distill-Llama-70B}{R1-Distill-Llama3.1-70B} & \underline{53.8} & \underline{49.8} & 41.7 & \underline{45.7} & \underline{53.4} & \underline{63.6} & \textbf{71.7} & \underline{57.8} & \textbf{61.6} & \underline{46.7} & \underline{44.5} & \underline{45.6}\\
\href{https://huggingface.co/deepseek-ai/DeepSeek-R1-Distill-Qwen-7B}{R1-Distill-Qwen-7B} & 50.4 & 39.9 & \underline{41.8} & 40.9 & 46.4 & 48.7 & 50.9 & 47.4 & 48.3 & 45.6 & 43.3 & 44.5\\
\cellcolor{gray!10} \textbf{Avg.} & \cellcolor{gray!10} 48.3 & \cellcolor{gray!10} 47.3 & \cellcolor{gray!10} 39.0 & \cellcolor{gray!10} 43.2 & \cellcolor{gray!10} 46.6 & \cellcolor{gray!10} 54.6 & \cellcolor{gray!10} 54.8 & \cellcolor{gray!10} 49.8 & \cellcolor{gray!10} 51.4 & \cellcolor{gray!10} 44.5 & \cellcolor{gray!10} 42.7 & \cellcolor{gray!10} 43.6 \\

\bottomrule[1.5pt]

\end{tabular}}
\caption{
Performance comparison of popular models on the \textbf{Biology} subset of \textsc{PRMBench-STEM}. The best performance for each category and task is in \textbf{bold}, while the second-best performance is \underline{underlined}.}
\vspace{-10pt}
\label{apdxtab:stem_bio}
\end{table*}


\begin{table*}[h]
\belowrulesep=0pt
\aboverulesep=0pt
\fontsize{14}{21}\selectfont
\centering

\resizebox{\textwidth}{!}{
\begin{tabular}{l c|ccc| ccccc| cc c}
\toprule[1.5pt]
\multirow{2}{*}{\textbf{Model}} & \multirow{2}{*}{\textbf{Overall}} & \multicolumn{3}{c|}{\textbf{Simplicity}}  & \multicolumn{5}{c|}{\textbf{Soundness}}& \multicolumn{3}{c}{\textbf{Sensitivity}}\\
\cmidrule(lr){3-5} \cmidrule(lr){6-10} \cmidrule(lr){11-13} 
&& \textbf{NR.} & \textbf{NCL.} & \textbf{Avg.} &\textbf{ES} &\textbf{SC.}&\textbf{DC.} &\textbf{CI} & \textbf{Avg.} &\textbf{PS} & \textbf{DR.} & \textbf{Avg.}   \\
 \midrule

\hline \multicolumn{13}{c}{\textit{\textbf{Open-source Process Level Reward Models}}} \\   \hline 
\href{https://huggingface.co/Skywork/Skywork-o1-Open-PRM-Qwen-2.5-1.5B}{Skywork-PRM-1.5B} & 48.1 & 45.9 & 43.9 & 44.9 & 46.3 & 52.5 & 54.0 & 51.3 & 51.0 & 45.3 & 43.3 & 44.3\\
\href{https://huggingface.co/Skywork/Skywork-o1-Open-PRM-Qwen-2.5-7B}{Skywork-PRM-7B} & 64.3 & 51.2 & 51.1 & 51.1 & 65.2 & \underline{70.4} & \underline{71.6} & 73.3 & 70.1 & 64.2 & 64.6 & 64.4\\
\href{https://huggingface.co/ScalableMath/llemma-7b-prm-prm800k-level-1to3-hf}{Llemma-PRM800k-7B} & 52.6 & 47.1 & 53.2 & 50.2 & 56.5 & 48.8 & 47.9 & 55.7 & 52.2 & 52.6 & 55.2 & 53.9\\
\href{https://huggingface.co/ScalableMath/llemma-7b-prm-metamath-level-1to3-hf}{Llemma-MetaMath-7B} & 45.2 & 45.3 & 41.5 & 43.4 & 44.9 & 48.2 & 46.3 & 47.6 & 46.7 & 42.3 & 42.8 & 42.5\\
\href{https://huggingface.co/ScalableMath/llemma-7b-oprm-prm800k-level-1to3-hf}{Llemma-oprm-7B} & 48.3 & 44.7 & 38.1 & 41.4 & 53.5 & 46.6 & 41.6 & 56.2 & 49.5 & 49.0 & 52.4 & 50.7\\
\href{https://github.com/KbsdJames/MATH-Minos}{MATHMinos-Mistral-7B} & 49.6 & 44.0 & 38.8 & 41.4 & 50.8 & 51.9 & 46.3 & 58.2 & 51.8 & 49.7 & 50.4 & 50.0\\
\href{https://huggingface.co/peiyi9979/math-shepherd-mistral-7b-prm}{MathShepherd-Mistral-7B} & 52.5 & 47.8 & 43.4 & 45.6 & 55.7 & 56.3 & 54.3 & 58.4 & 56.2 & 50.2 & 49.5 & 49.8\\
\href{https://huggingface.co/GAIR/ReasonEval-7B}{ReasonEval-7B} & 62.9 & \textbf{69.4} & \textbf{60.1} & \textbf{64.7} & 59.2 & 66.3 & 64.1 & 70.3 & 65.0 & 57.4 & 55.5 & 56.4\\
\href{https://huggingface.co/GAIR/ReasonEval-34B}{ReasonEval-34B} & 63.2 & \underline{63.5} & \underline{55.8} & \underline{59.6} & 65.6 & 58.6 & 57.8 & 66.9 & 62.2 & 61.9 & 65.6 & 63.7\\
\href{https://huggingface.co/RLHFlow/Llama3.1-8B-PRM-Mistral-Data}{RLHFlow-PRM-Mistral-8B} & 45.6 & 43.3 & 38.5 & 40.9 & 44.2 & 50.9 & 47.7 & 51.2 & 48.5 & 44.1 & 42.3 & 43.2\\
\href{https://huggingface.co/RLHFlow/Llama3.1-8B-PRM-Deepseek-Data}{RLHFlow-PRM-Deepseek-8B} & 47.3 & 43.5 & 38.3 & 40.9 & 46.9 & 51.8 & 50.5 & 56.3 & 51.4 & 45.4 & 43.1 & 44.2\\
\href{https://huggingface.co/Qwen/Qwen2.5-Math-PRM-7B}{Qwen2.5-Math-PRM-7B} & \underline{66.3} & 43.5 & 49.9 & 46.7 & \underline{70.9} & 66.8 & 68.8 & \underline{80.0} & \underline{71.6} & \underline{67.0} & \underline{68.1} & \underline{67.5}\\
\href{https://huggingface.co/Qwen/Qwen2.5-Math-PRM-72B}{Qwen2.5-Math-PRM-72B} & \textbf{71.4} & 44.8 & 53.2 & 49.0 & \textbf{77.3} & \textbf{71.6} & \textbf{74.6} & \textbf{81.0} & \textbf{76.1} & \textbf{73.1} & \textbf{77.5} & \textbf{75.3}\\
\cellcolor{gray!10} \textbf{Avg.} & \cellcolor{gray!10} 55.2 & \cellcolor{gray!10} 48.8 & \cellcolor{gray!10} 46.6 & \cellcolor{gray!10} 47.7 & \cellcolor{gray!10} 56.7 & \cellcolor{gray!10} 57.0 & \cellcolor{gray!10} 55.8 & \cellcolor{gray!10} 62.0 & \cellcolor{gray!10} 57.9 & \cellcolor{gray!10} 54.0 & \cellcolor{gray!10} 54.6 & \cellcolor{gray!10} 54.3 \\
\hline \multicolumn{13}{c}{\textit{\textbf{Open LLMs, Prompted as Critic Models}}} \\   \hline 
\href{https://huggingface.co/meta-math/MetaMath-7B-V1.0}{MetaMath-7B} & 45.6 & 44.2 & 37.5 & 40.9 & 46.0 & 48.5 & 48.0 & 47.3 & 47.4 & 45.6 & \underline{45.5} & \underline{45.5}\\
\href{https://huggingface.co/meta-math/MetaMath-13B-V1.0}{MetaMath-13B} & 44.3 & 42.9 & 37.0 & 40.0 & 44.7 & 48.6 & 47.0 & 45.3 & 46.4 & 45.2 & 42.2 & 43.7\\
\href{https://huggingface.co/Qwen/Qwen2.5-Math-PRM-72B}{Qwen2.5-Math-72B} & 43.2 & \textbf{50.5} & 34.8 & \underline{42.6} & 37.6 & \underline{61.7} & 43.8 & 43.3 & 46.6 & 37.5 & 36.6 & 37.0\\
\href{https://huggingface.co/deepseek-ai/DeepSeek-R1-Distill-Llama-70B}{R1-Distill-Llama3.1-70B} & \underline{50.6} & \underline{45.9} & \textbf{41.0} & \textbf{43.5} & \textbf{48.2} & \textbf{65.7} & \textbf{58.1} & \textbf{57.8} & \textbf{57.5} & \underline{46.7} & 42.8 & 44.8\\
\href{https://huggingface.co/deepseek-ai/DeepSeek-R1-Distill-Qwen-7B}{R1-Distill-Qwen-7B} & \textbf{51.8} & 35.6 & \underline{41.0} & 38.3 & \underline{47.5} & 51.0 & \underline{50.1} & \underline{49.2} & \underline{49.5} & \textbf{47.1} & \textbf{46.2} & \textbf{46.7}\\
\cellcolor{gray!10} \textbf{Avg.} & \cellcolor{gray!10} 47.1 & \cellcolor{gray!10} 43.8 & \cellcolor{gray!10} 38.2 & \cellcolor{gray!10} 41.0 & \cellcolor{gray!10} 44.8 & \cellcolor{gray!10} 55.1 & \cellcolor{gray!10} 49.4 & \cellcolor{gray!10} 48.6 & \cellcolor{gray!10} 49.5 & \cellcolor{gray!10} 44.4 & \cellcolor{gray!10} 42.7 & \cellcolor{gray!10} 43.5 \\

\bottomrule[1.5pt]

\end{tabular}}
\caption{
Performance comparison of popular models on the \textbf{Chemistry} subset of \textsc{PRMBench-STEM}. The best performance for each category and task is in \textbf{bold}, while the second-best performance is \underline{underlined}. }
\vspace{-10pt}
\label{apdxtab:stem_chem}
\end{table*}


\begin{table*}[h]
\belowrulesep=0pt
\aboverulesep=0pt
\fontsize{14}{21}\selectfont
\centering

\resizebox{\textwidth}{!}{
\begin{tabular}{l c|ccc| ccccc| cc c}
\toprule[1.5pt]
\multirow{2}{*}{\textbf{Model}} & \multirow{2}{*}{\textbf{Overall}} & \multicolumn{3}{c|}{\textbf{Simplicity}}  & \multicolumn{5}{c|}{\textbf{Soundness}}& \multicolumn{3}{c}{\textbf{Sensitivity}}\\
\cmidrule(lr){3-5} \cmidrule(lr){6-10} \cmidrule(lr){11-13} 
&& \textbf{NR.} & \textbf{NCL.} & \textbf{Avg.} &\textbf{ES} &\textbf{SC.}&\textbf{DC.} &\textbf{CI} & \textbf{Avg.} &\textbf{PS} & \textbf{DR.} & \textbf{Avg.}   \\
 \midrule

\hline \multicolumn{13}{c}{\textit{\textbf{Open-source Process Level Reward Models}}} \\   \hline 
\href{https://huggingface.co/Skywork/Skywork-o1-Open-PRM-Qwen-2.5-1.5B}{Skywork-PRM-1.5B} & 50.3 & 44.7 & 41.8 & 43.3 & 51.8 & 57.7 & 55.2 & 56.1 & 55.2 & 46.8 & 46.8 & 46.8\\
\href{https://huggingface.co/Skywork/Skywork-o1-Open-PRM-Qwen-2.5-7B}{Skywork-PRM-7B} & 65.2 & 51.1 & 51.9 & 51.5 & 68.3 & \underline{70.3} & \textbf{72.7} & 76.4 & 71.9 & 64.8 & 65.1 & 65.0\\
\href{https://huggingface.co/ScalableMath/llemma-7b-prm-prm800k-level-1to3-hf}{Llemma-PRM800k-7B} & 53.8 & 49.1 & 52.0 & 50.6 & 57.3 & 49.6 & 49.4 & 57.3 & 53.4 & 53.8 & 58.6 & 56.2\\
\href{https://huggingface.co/ScalableMath/llemma-7b-prm-metamath-level-1to3-hf}{Llemma-MetaMath-7B} & 46.9 & 46.4 & 42.0 & 44.2 & 46.5 & 47.2 & 44.2 & 51.7 & 47.4 & 47.8 & 45.4 & 46.6\\
\href{https://huggingface.co/ScalableMath/llemma-7b-oprm-prm800k-level-1to3-hf}{Llemma-oprm-7B} & 47.1 & 42.8 & 39.7 & 41.2 & 52.2 & 42.7 & 40.9 & 54.3 & 47.5 & 48.1 & 51.6 & 49.8\\
\href{https://github.com/KbsdJames/MATH-Minos}{MATHMinos-Mistral-7B} & 46.5 & 43.6 & 38.9 & 41.2 & 46.9 & 47.8 & 44.2 & 57.8 & 49.2 & 45.3 & 43.3 & 44.3\\
\href{https://huggingface.co/peiyi9979/math-shepherd-mistral-7b-prm}{MathShepherd-Mistral-7B} & 53.5 & 48.0 & 44.8 & 46.4 & 55.9 & 56.3 & 54.8 & 62.8 & 57.4 & 48.6 & 53.5 & 51.1\\
\href{https://huggingface.co/GAIR/ReasonEval-7B}{ReasonEval-7B} & 62.5 & \textbf{71.5} & \textbf{57.2} & \textbf{64.4} & 60.1 & 66.9 & 62.2 & 70.6 & 65.0 & 56.9 & 53.9 & 55.4\\
\href{https://huggingface.co/GAIR/ReasonEval-34B}{ReasonEval-34B} & 64.2 & \underline{68.5} & 52.0 & \underline{60.2} & 68.5 & 59.3 & 56.4 & 69.8 & 63.5 & 63.7 & 66.6 & 65.2\\
\href{https://huggingface.co/RLHFlow/Llama3.1-8B-PRM-Mistral-Data}{RLHFlow-PRM-Mistral-8B} & 46.5 & 42.7 & 37.5 & 40.1 & 46.0 & 51.0 & 50.5 & 55.7 & 50.8 & 44.2 & 42.2 & 43.2\\
\href{https://huggingface.co/Qwen/Qwen2.5-Math-PRM-7B}{Qwen2.5-Math-PRM-7B} & \underline{66.9} & 43.7 & 49.3 & 46.5 & \underline{73.6} & 66.4 & 70.3 & \underline{80.0} & \underline{72.6} & \underline{69.6} & \underline{68.8} & \underline{69.2}\\
\href{https://huggingface.co/Qwen/Qwen2.5-Math-PRM-72B}{Qwen2.5-Math-PRM-72B} & \textbf{71.1} & 44.2 & \underline{56.2} & 50.2 & \textbf{78.9} & \textbf{71.1} & \underline{71.1} & \textbf{80.3} & \textbf{75.4} & \textbf{72.9} & \textbf{77.6} & \textbf{75.2}\\
\cellcolor{gray!10} \textbf{Avg.} & \cellcolor{gray!10} 55.6 & \cellcolor{gray!10} 45.9 & \cellcolor{gray!10} 46.3 & \cellcolor{gray!10} 46.1 & \cellcolor{gray!10} 58.1 & \cellcolor{gray!10} 56.8 & \cellcolor{gray!10} 55.7 & \cellcolor{gray!10} 64.2 & \cellcolor{gray!10} 58.7 & \cellcolor{gray!10} 54.5 & \cellcolor{gray!10} 55.2 & \cellcolor{gray!10} 54.8 \\
\hline \multicolumn{13}{c}{\textit{\textbf{Open LLMs, Prompted as Critic Models}}} \\   \hline 
\href{https://huggingface.co/meta-math/MetaMath-7B-V1.0}{MetaMath-7B} & 45.6 & 43.5 & 36.9 & 40.2 & \underline{47.9} & 48.9 & \underline{49.1} & 48.1 & 48.5 & \underline{47.3} & \underline{43.6} & \underline{45.4}\\
\href{https://huggingface.co/meta-math/MetaMath-13B-V1.0}{MetaMath-13B} & 43.8 & 42.7 & 36.6 & 39.7 & 44.5 & 46.7 & 48.0 & 46.3 & 46.4 & 41.5 & 41.9 & 41.7\\
\href{https://huggingface.co/Qwen/Qwen2.5-Math-PRM-72B}{Qwen2.5-Math-72B} & 44.8 & \textbf{50.6} & \underline{37.8} & \textbf{44.2} & 39.8 & \textbf{64.5} & 45.5 & 47.5 & \underline{49.3} & 37.9 & 36.2 & 37.1\\
\href{https://huggingface.co/deepseek-ai/DeepSeek-R1-Distill-Llama-70B}{R1-Distill-Llama3.1-70B} & \textbf{53.3} & \underline{44.5} & \textbf{40.4} & \underline{42.5} & \textbf{53.0} & \underline{63.8} & \textbf{59.4} & \textbf{64.6} & \textbf{60.2} & \textbf{49.6} & \textbf{48.3} & \textbf{48.9}\\
\href{https://huggingface.co/deepseek-ai/DeepSeek-R1-Distill-Qwen-7B}{R1-Distill-Qwen-7B} & \underline{51.0} & 29.7 & 34.8 & 32.3 & 44.1 & 48.1 & 45.7 & \underline{48.9} & 46.7 & 44.0 & 40.7 & 42.3\\
\cellcolor{gray!10} \textbf{Avg.} & \cellcolor{gray!10} 47.7 & \cellcolor{gray!10} 42.2 & \cellcolor{gray!10} 37.3 & \cellcolor{gray!10} 39.8 & \cellcolor{gray!10} 45.9 & \cellcolor{gray!10} 54.4 & \cellcolor{gray!10} 49.5 & \cellcolor{gray!10} 51.1 & \cellcolor{gray!10} 50.2 & \cellcolor{gray!10} 44.0 & \cellcolor{gray!10} 42.2 & \cellcolor{gray!10} 43.1 \\

\bottomrule[1.5pt]

\end{tabular}}
\caption{
Performance comparison of popular models on the \textbf{Physics} subset of \textsc{PRMBench-STEM}. The best performance for each category and task is in \textbf{bold}, while the second-best performance is \underline{underlined}.}
\vspace{-10pt}
\label{apdxtab:stem_physics}
\end{table*}

\section{Details for Further Analysis}
\subsection{Error Analysis}
\label{appendix:error_analysis}
A representative test case and the corresponding model performances are presented in Table \ref{tab:error_cases}. This example involves a counterfactual reasoning process, where steps eight through thirteen contain information that contradicts the correct computational principles and should be classified as ``negative''. However, most models fail to identify these erroneous reasoning steps and assign relatively positive rewards, except for GPT-4o. While GPT-4o provides a relatively accurate reward, its judgments for key steps are only marginally negative, reflecting low confidence. This highlights a significant room for improvement in PRMs' detailed error-detection capabilities.

\begin{table}[tp]
\belowrulesep=0pt
\aboverulesep=0pt
\fontsize{14}{21}\selectfont
\centering

\resizebox{\columnwidth}{!}{
\begin{tabular}{l c c cccc }
\toprule[1.5pt]
\textbf{Method} &  \textbf{GSM8K}  &  \textbf{MATH} & \textbf{OlymBen}  &  \textbf{MMLU} & \textbf{Avg.} & \textbf{PRMScore} \\
 \midrule

\textbf{Pass@8} & 98.9 & 96.2 & 79.8 & 96.7 & 92.9 & - \\
\textbf{Maj@8} & 95.4 & 71.8 & 40.3 & 85.8 & 73.3 & - \\
 \midrule 
\href{https://huggingface.co/Skywork/Skywork-o1-Open-PRM-Qwen-2.5-1.5B}{Skywork-PRM-1.5B} & 96.7 & 89.2 & 58.8 & 90.2 & 83.7 & 61.1\\
\href{https://huggingface.co/Skywork/Skywork-o1-Open-PRM-Qwen-2.5-7B}{Skywork-PRM-7B} & 97.1 & 90.0 & 60.1 & 90.3 & 84.4 & 65.1\\
\href{https://huggingface.co/ScalableMath/llemma-7b-prm-prm800k-level-1to3-hf}{Llemma-PRM800k-7B} & 96.0 & 87.4 & 58.3 & 90.0 & 82.9 & 52.0\\
\href{https://huggingface.co/ScalableMath/llemma-7b-prm-metamath-level-1to3-hf}{Llemma-MetaMath-7B} & 96.0 & 88.2 & 58.6 & 90.0 & 83.2 & 50.5\\
\href{https://huggingface.co/ScalableMath/llemma-7b-oprm-prm800k-level-1to3-hf}{Llemma-oprm-7B} & 96.4 & 86.6 & 58.0 & 89.9 & 82.7 & 50.3\\
\href{https://github.com/KbsdJames/MATH-Minos}{MATHMinos-Mistral-7B} & 95.8 & 88.3 & 59.1 & 89.1 & 83.1 & 54.2\\
\href{https://huggingface.co/peiyi9979/math-shepherd-mistral-7b-prm}{MathShepherd-Mistral-7B} & 96.6 & 88.6 & 60.0 & 90.0 & 83.8 & 47.0\\
\href{https://huggingface.co/GAIR/ReasonEval-7B}{ReasonEval-7B} & 96.4 & 87.0 & 58.4 & 90.0 & 82.9 & 60.1\\
\href{https://huggingface.co/RLHFlow/Llama3.1-8B-PRM-Mistral-Data}{RLHFlow-PRM-Mistral-8B} & 96.5 & 88.4 & 59.1 & 90.4 & 83.6 & 54.4\\
\href{https://huggingface.co/RLHFlow/Llama3.1-8B-PRM-Deepseek-Data}{RLHFlow-PRM-Deepseek-8B} & 96.4 & 87.6 & 58.5 & 90.2 & 83.2 & 54.2\\
\href{https://huggingface.co/GAIR/ReasonEval-34B}{ReasonEval-34B} & 96.4 & 86.9 & 56.9 & 90.0 & 82.6 & 60.5\\
\href{https://huggingface.co/Qwen/Qwen2.5-Math-PRM-7B}{Qwen2.5-Math-PRM-7B} & 96.7 & 88.0 & 58.7 & 89.8 & 83.3 & 65.5\\
\href{https://huggingface.co/Qwen/Qwen2.5-Math-PRM-72B}{Qwen2.5-Math-PRM-72B} & 96.7 & 89.3 & 60.5 & 90.0 & 84.2 & 68.2\\
 \midrule 
\textbf{Standard Deviation ($\sigma$)} & 0.35 & 0.97 & 0.93 & 0.29 & 0.53 & 6.41 \\
\textbf{Somers' D correlation} & 0.50 & 0.26 & 0.22 & 0.19 & 0.28 & 1.00 \\

\bottomrule[1.5pt]

\end{tabular}}
\caption{
Performance comparison on Best-of-8 using different \textbf{PRMs}. $\sigma$ represents the standard deviation of model performances across all benchmarks. Somers’ D refers to the Somers’ D correlation between PRMScore and specific benchmarks.
}
\label{apdxtab:bon}
\end{table}

\begin{table}[tp]
\belowrulesep=0pt
\aboverulesep=0pt
\fontsize{14}{21}\selectfont
\centering

\resizebox{\columnwidth}{!}{
\begin{tabular}{l c c cccc }
\toprule[1.5pt]
\textbf{Method} &  \textbf{GSM8K}  &  \textbf{MATH} & \textbf{OlymBen}  &  \textbf{MMLU} & \textbf{Avg.} & \textbf{PRMScore} \\
 \midrule

\textbf{Pass@8} & 99.0 & 96.0 & 77.0 & 94.0 & 91.5 & - \\
\textbf{Maj@8} & 96.5 & 68.0 & 41.0 & 86.0 & 72.9 & - \\
 \midrule 
\href{https://huggingface.co/Qwen/QwQ-32B-Preview}{QwQ-Preview-32B} & 96.5 & 83.5 & 56.5 & 87.5 & 81.0 & 63.6\\
\href{https://huggingface.co/meta-math/MetaMath-7B-V1.0}{MetaMath-7B} & 95.5 & 82.0 & 58.0 & 88.0 & 80.9 & 49.7\\
\href{https://huggingface.co/meta-math/MetaMath-13B-V1.0}{MetaMath-13B} & 95.0 & 84.5 & 57.5 & 86.5 & 80.9 & 49.4\\
\href{https://huggingface.co/meta-math/MetaMath-70B-V1.0}{MetaMath-70B} & 96.5 & 85.0 & 58.5 & 86.0 & 81.5 & 45.9\\
\href{https://huggingface.co/Qwen/Qwen2.5-Math-7B-Instruct}{Qwen2.5-Math-7B} & 96.0 & 85.5 & 59.0 & 86.5 & 81.8 & 49.2\\
\href{https://huggingface.co/Qwen/Qwen2.5-Math-PRM-72B}{Qwen2.5-Math-72B} & 97.5 & 82.0 & 53.5 & 88.5 & 80.4 & 57.4\\
\href{https://huggingface.co/deepseek-ai/DeepSeek-R1-Distill-Llama-8B}{R1-Distill-Llama3.1-8B} & 96.5 & 84.5 & 53.5 & 87.5 & 80.5 & 52.7\\
\href{https://huggingface.co/deepseek-ai/DeepSeek-R1-Distill-Llama-70B}{R1-Distill-Llama3.1-70B} & 95.0 & 83.0 & 59.5 & 86.5 & 81.0 & 57.5\\
\href{https://huggingface.co/deepseek-ai/DeepSeek-R1-Distill-Qwen-7B}{R1-Distill-Qwen-7B} & 97.0 & 86.5 & 59.0 & 86.5 & 82.2 & 52.6\\
\href{https://huggingface.co/deepseek-ai/DeepSeek-R1-Distill-Qwen-32B}{R1-Distill-Qwen-32B} & 97.0 & 82.0 & 60.5 & 84.5 & 81.0 & 60.2\\
\href{https://huggingface.co/WizardLMTeam/WizardMath-7B-V1.0}{WizardMath-7B} & 96.5 & 82.5 & 60.0 & 87.0 & 81.5 & 49.2\\
\href{https://deepmind.google/technologies/gemini/flash/}{Gemini-2.0-flash-exp} & 97.0 & 81.5 & 56.5 & 86.5 & 80.4 & 66.0\\
\href{https://ai.google.dev/gemini-api/docs/thinking-mode}{Gemini-2.0-thinking-exp-1219} & 98.0 & 87.5 & 60.5 & 89.5 & 83.9 & 68.8\\
\midrule
\textbf{Standard Deviation ($\sigma$)} & 0.87 & 1.83 & 2.25 & 1.19 & 0.91 & 7.02 \\
\textbf{Somers' D correlation} & 0.36 & -0.21 & 0.03 & 0.24 & -0.15 & 1.00 \\

\bottomrule[1.5pt]

\end{tabular}}
\caption{
Performance comparison on Best-of-8 using different \textbf{LLMs as a Judge}. $\sigma$ represents the standard deviation of model performances across all benchmarks. Somers’ D refers to the Somers’ D correlation between PRMScore and specific benchmarks.
}
\label{apdxtab:bon2}
\end{table}

\subsection{Details for BoN Evaluation}
\label{appendix:bon_details}
Following \citet{zhang2025qwen_lessons}, we sample eight responses (i.e., N=8) from Qwen-QwQ-Preview 32B~\cite{qwen-qwq-32b-preview}. During evaluation, the PRMs are tasked with assigning a validity score to each step within every candidate response. The overall score for each candidate response is calculated by multiplying the individual step scores, as outlined in \citet{lets_verify_stepbystep}. We then select the highest-ranked candidate response, compare it with the correct answer, and calculate the accuracy, which we refer to as prm@8. Additionally, we report the result of majority voting among the eight sampled responses (maj@8) as the baseline, and we define pass@8 as the proportion of test samples where any of the eight samplings lead to the correct final answer, which serves as the upper bound.

We conduct BoN evaluation across all models on GSM8K~\cite{cobbe2021gsm8k}, MATH~\cite{hendrycks2020math_benchmark}, Olympiad Bench~\cite{he2024olympiadbench} and MMLU \cite{hendrycks2020MMLU}. Due to the limit of the space, We omit some results in Table \ref{tab:correlation}. Thus we provide the full results of all PRMs in Table \ref{apdxtab:bon}. 
Moreover, due to cost constraints, we select a subset of 200 samples for each benchmark and obtain a total of 800 samples to evaluate generative LLMs of their BoN performance. The results are shown in Table \ref{apdxtab:bon2}.

\section{Instructions for Human Annotators}
\label{appendix:annotator_instruction}
\subsection{Backgrounds}
With the emergence of multi-step reasoning enhanced language models such as OpenAI o1, and Deepmind Gemini-thinking, these models demonstrate the ability to decompose complex problems and solve them step by step. However, while their solutions often appear correct, they may contain errors in understanding, calculation, or reasoning logic, which is also known as false positive situations. A popular way to evaluate the results generated by these models is by utilizing process-level reward models (PRMs). Nevertheless, PRMs are fallible and not always correct. Existing benchmarks are not adequate for evaluating PRMs on different error types. Therefore, we are building a comprehensive evaluation benchmark for PRMs that can have a fine-grained detection of PRMs.

\subsection{Task Definition}
We begin by collecting completely correct multi-solution data and leveraging state-of-the-art LLMs to introduce various types of errors into these correct solutions, thereby generating our test cases. The detailed error types are described in Section 3. All synthesized data instances undergo an initial filtering process based on specific features.

Your task is to identify whether the modification taken is reasonable and whether the modified data instance is different from the original data instance.

\subsubsection*{Sub-task 1}
The first sub-task is a binary classification task whose options include yes and no. Your task is to decide whether the modified step-by-step solution generated by LLMs is reasonable. The word ``reasonable'' has two aspects for evaluation.
\begin{itemize}[leftmargin=*]
\setlength{\itemsep}{0pt}
\item The modified process generated by LLMs seems like a possible solution path that could happen. 
\item The modified process generated by LLMs is exactly wrong and the type of error is suitable for the current ``classification''. 
\end{itemize}

Please assign a ``yes'' for this sub-task if both of the answers to the above two questions are ``yes''. Otherwise, assign a ``no'' for this sub-task.

\subsubsection{Sub-task 2}
The second sub-task is a binary classification task whose options include yes and no. Your task is to decide whether the modified step-by-step solution generated by LLMs is different from the original solution process. The word ``different'' means the modified solution process is logically different from the original one, or there exist different statements compared to the original process.

Please assign a ``yes'' to this sub-task if your answer to the above question is ``yes''. Otherwise, assign a ``no'' for this sub-task.

\subsubsection{Error Types}

\paragraph{Redundancy} refers to a process that is not the most concise or efficient, as it includes one or more redundant steps that can be removed without affecting the correctness of the overall solution path. For example, if $ A \to B $ represents a correct inference chain, your task is to introduce one or more redundant steps $ C = \{c | c \text{ is redundent}\} $ and reformulate the solution chain as $ A \to C \to B $.

\paragraph{Circular logic} is a specific form of redundancy, characterized by a reasoning chain that starts at a step $ S $, progresses through a sequence of steps, and ultimately loops back to $ S $. Symbolically, this can be expressed as $ S \to A \to B \to S $, where $ S $, $ A $, and $ B $ represent individual reasoning steps. Your task is to modify the reasoning process to introduce such circular logic.

\paragraph{Counterfactual} A counterfactual step refers to a statement within a reasoning chain that contradicts ground truth or established theories. Such contradictions can arise from relying on outdated theories, omitting critical constraints in a theory, or incorporating erroneous assumptions. Your task is to modify the reasoning process to introduce such counterfactual steps.

\paragraph{Step contradiction} refers to a conflict between a specific step and other steps within a reasoning path. Given a reasoning path $ P = {S_1, S_2, \dots, S_n} $, a step contradiction exists if $ S_i \perp S_j $, where $ i, j \in [1, n] $ and $ i \neq j $. Your task is to modify the reasoning process to introduce such step contradiction steps.

\paragraph{Domain inconsistency} is a special type of counterfactual. It refers to a step or a few steps within the reasoning chain that uses a statement or theory valid in other domains or cases but is not valid within the current reasoning chain. Your task is to modify the reasoning process to introduce steps for such domain inconsistency.

\paragraph{Confident hallucination} is a special type of counterfactual. It refers to a statement within the reasoning chain that contradicts established ground truth and is presented with an overly confident tone. In other words, it involves stating an incorrect statement with unwarranted certainty. Your task is to modify the reasoning process to introduce such confident hallucination steps.

\paragraph{Missing condition or prerequisite} refers to a flaw in the reasoning chain where critical premises, assumptions, or necessary conditions are absent. This omission results in logical gaps, incomplete reasoning, or biased conclusions. For example, when a missing condition occurs, the model must solve the problem through case analysis or further investigation. However, the answer becomes incorrect if the model overlooks the missing condition and proceeds with standard reasoning methods. Your task is to modify the reasoning process to introduce such missing condition errors.

\paragraph{Deception or traps} refer to statements that appear to be correct or align with ground truth but are subtly altered to introduce inaccuracies while maintaining the illusion of correctness. Your task is to modify the reasoning process to introduce such deception or trap error steps.

\section{Prompts}
\subsection{Prompts For Generating Data}
\label{appendix_generation_prompts}
\begin{figure*}[h]
    \centering
    \includegraphics[width=0.9\linewidth]{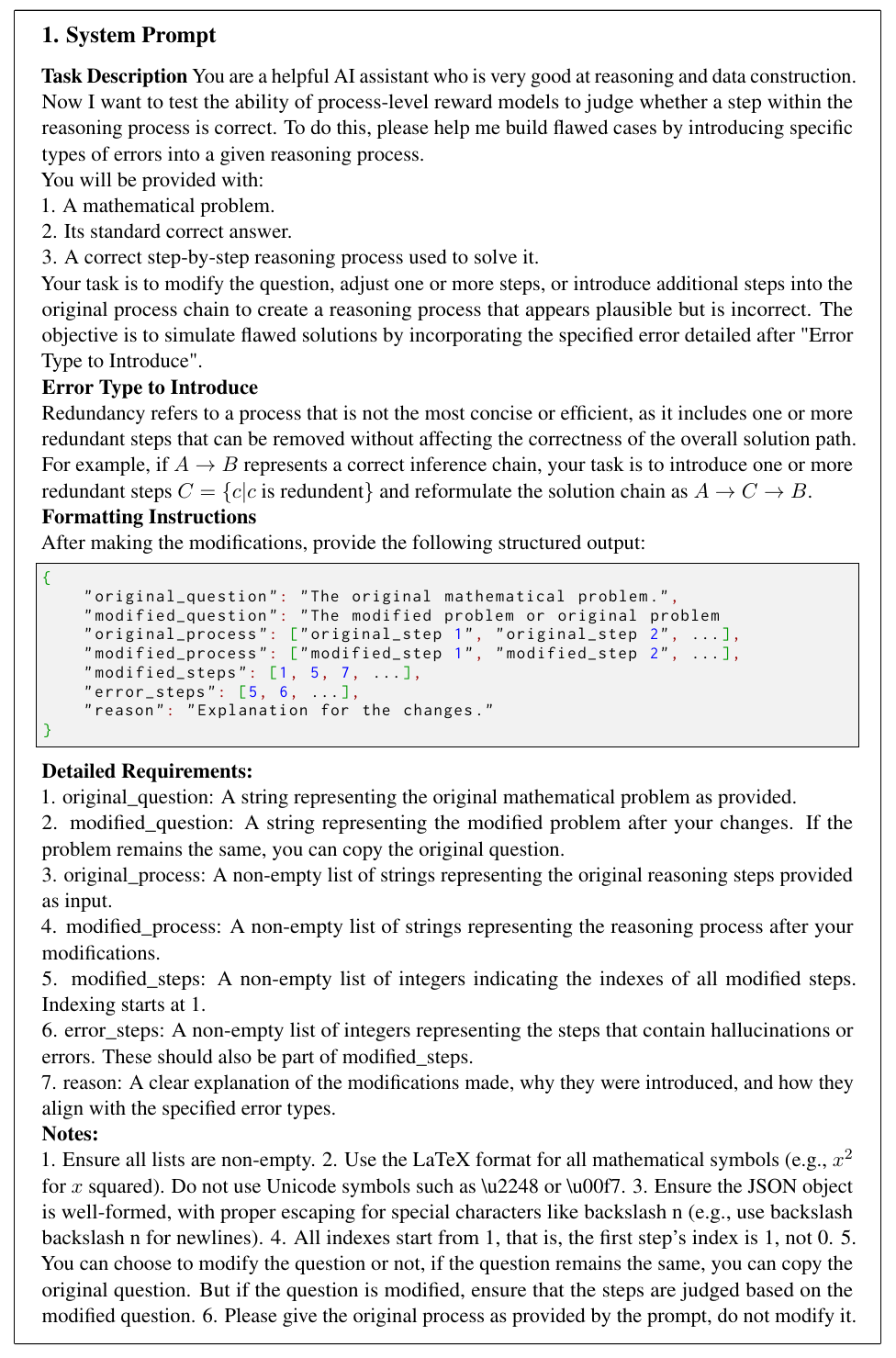}
    \caption{Prompts used for test case construction during data generation (1).}
    \label{apdxfig:example:dg1}
\end{figure*}

\begin{figure*}[h]
    \centering
    \includegraphics[width=0.9\linewidth]{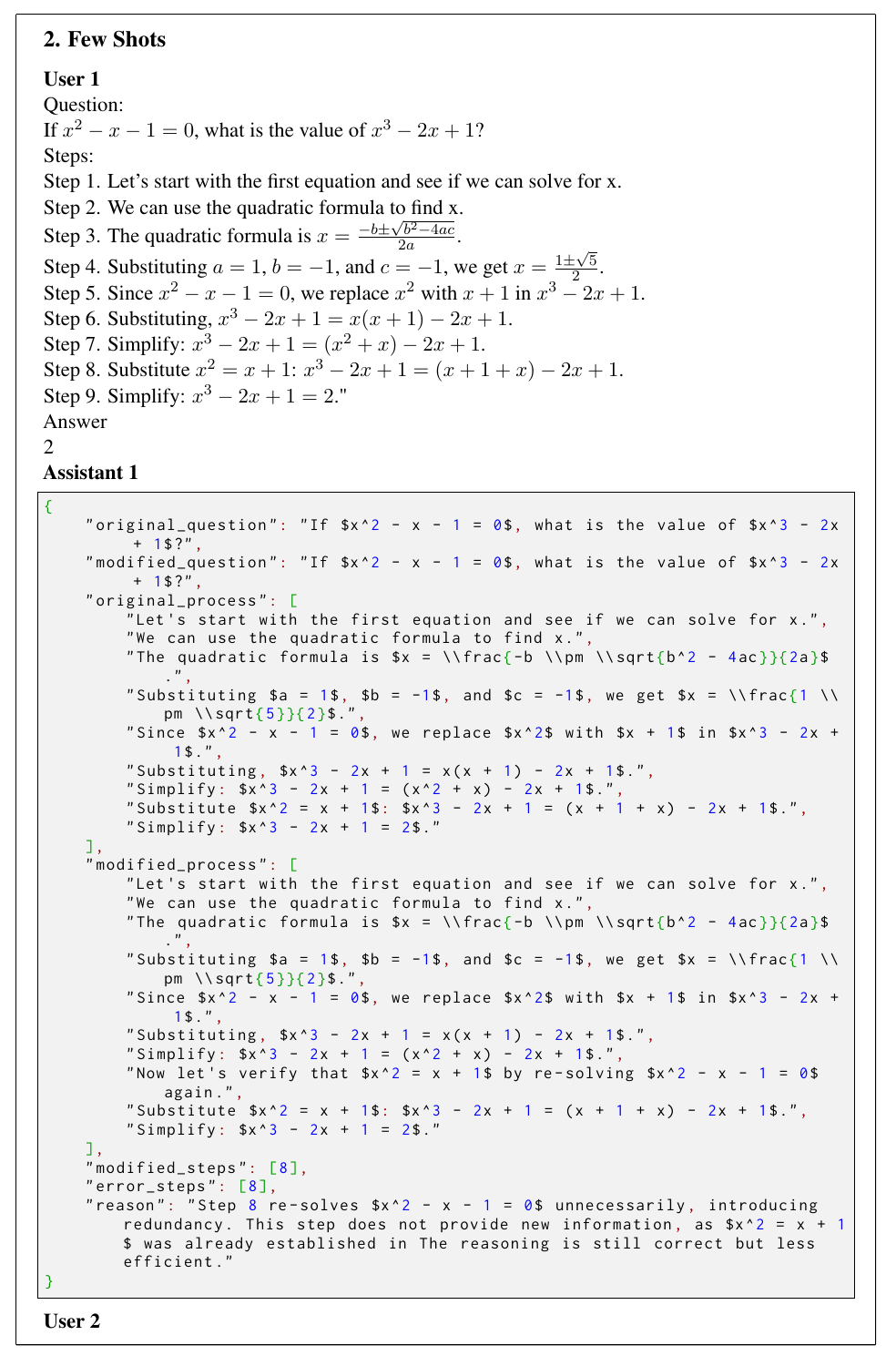}
    \caption{Prompts used for test case construction during data generation (2).}
    \label{apdxfig:example:dg2}
\end{figure*}

\begin{figure*}[h]
    \centering
    \includegraphics[width=0.9\linewidth]{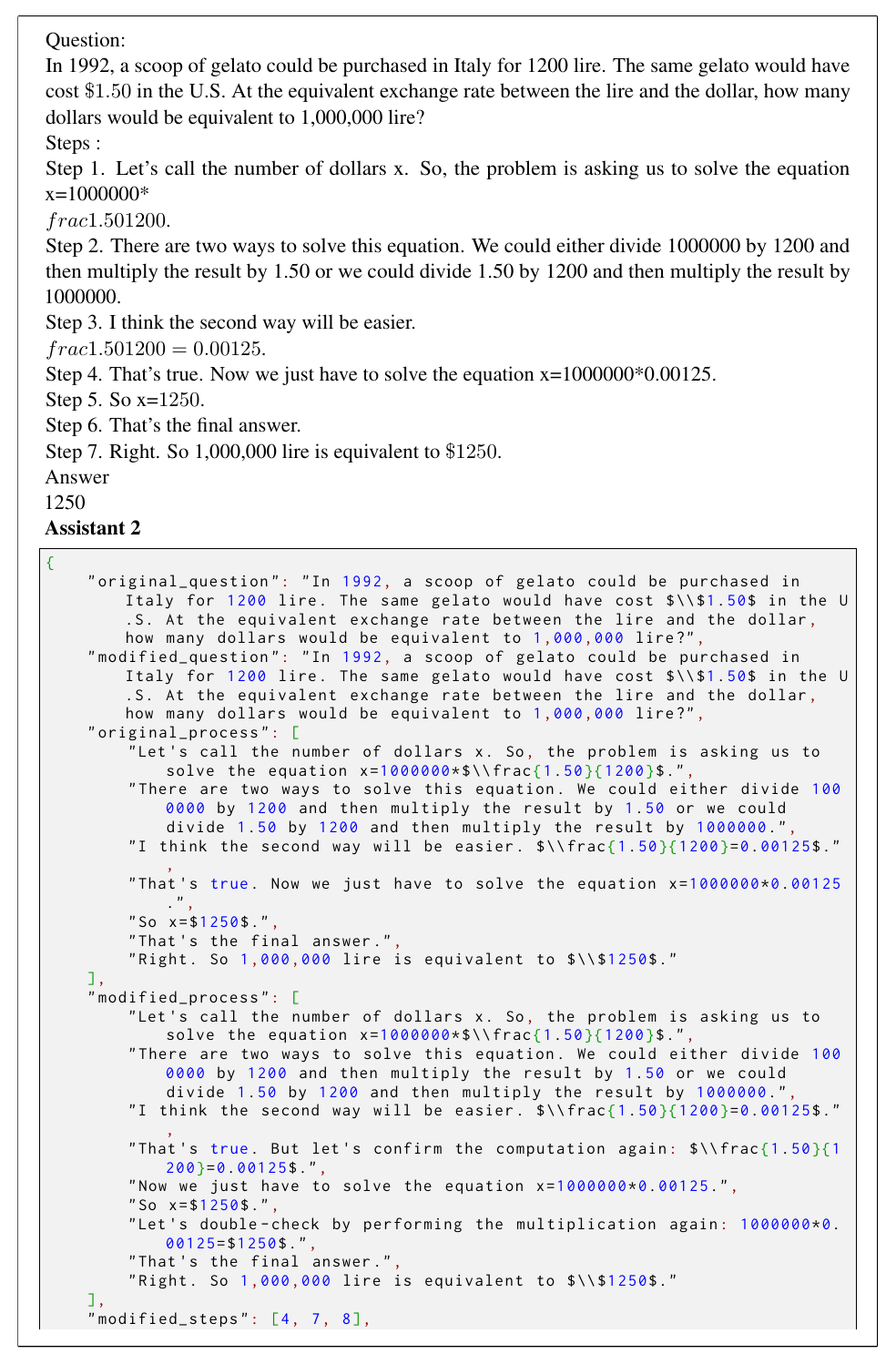}
    \caption{Prompts used for test case construction during data generation (3).}
    \label{apdxfig:example:dg3}
\end{figure*}

\begin{figure*}[h]
    \centering
    \includegraphics[width=0.9\linewidth]{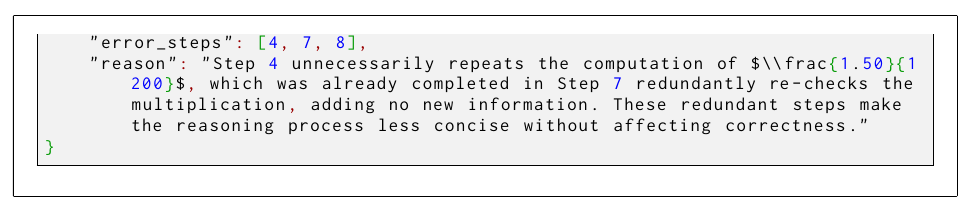}
    \caption{Prompts used for test case construction during data generation (4).}
    \label{apdxfig:example:dg4}
\end{figure*}

As introduced in Section \ref{sec:data_curation}, we query GPT-4o \cite{openai2024gpt4o} to synthesize the metadata at the very first step of our test case construction procedure. To better prompt LLMs to generate high-quality data instances, we carefully designed our prompts, which are displayed in Figure \ref{apdxfig:example:dg1}-\ref{apdxfig:example:dg4}. We display only one example here due to limitations in space. And the prompts can be found in our supplementary materials. 

\subsection{Prompts For Evaluating Generative LLMs}
\label{appendix_evaluation_prompts}
\begin{figure*}[h]
    \centering
    \includegraphics[width=0.9\linewidth]{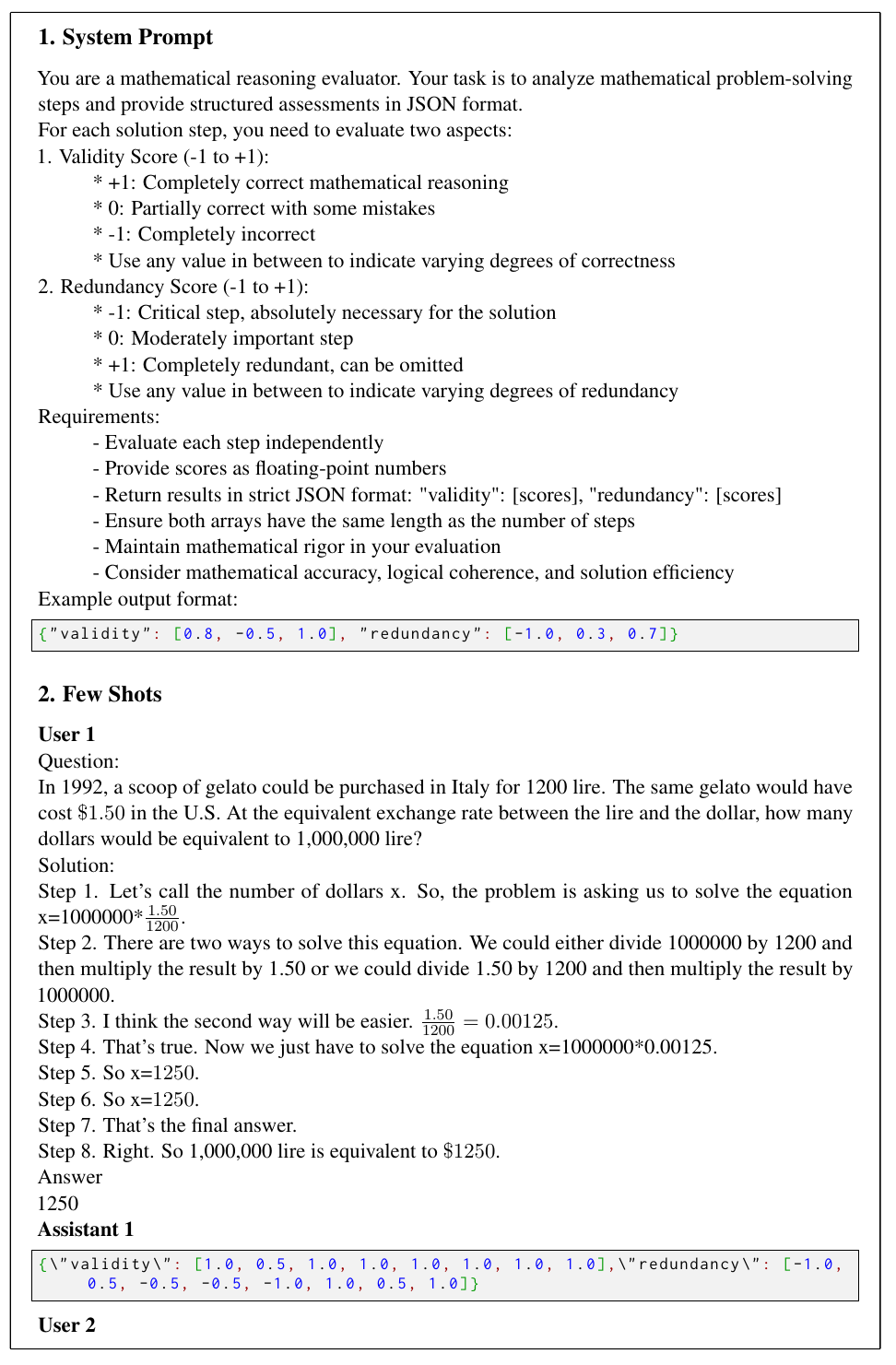}
    \caption{Prompts used for evaluating generative LLMs (1).}
    \label{apdxfig:example:elm1}
\end{figure*}

\begin{figure*}[h]
    \centering
    \includegraphics[width=0.9\linewidth]{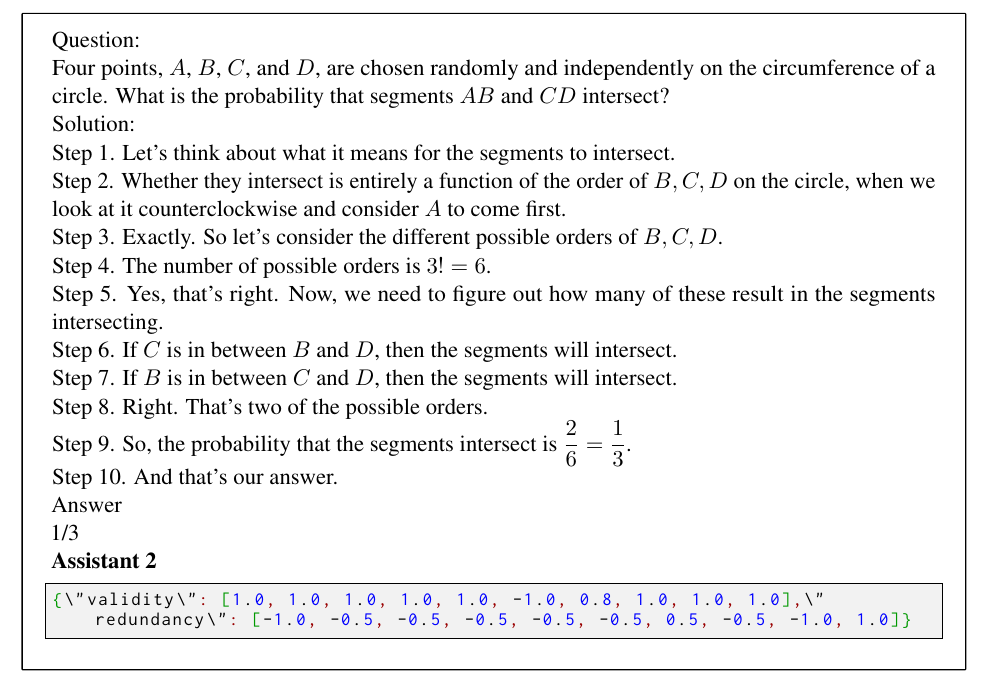}
    \caption{Prompts used for evaluating generative LLMs (2).}
    \label{apdxfig:example:elm2}
\end{figure*}

As introduced in Section \ref{sec:evaluate_models}, we prompt some state-of-the-art generative LLMs as critic models to evaluate their rewarding capabilities on \textsc{PRMBench}. To make a fair comparison between different models, we carefully design the prompts and utilize a unified prompt to query them. The prompt used is displayed in Figure \ref{apdxfig:example:elm1} and \ref{apdxfig:example:elm2}.

\section{Further Discussion}
\label{appendix:further_discussion}
Inspired by the results and discoveries on \textsc{PRMBench}, we further propose several promising directions for future research, which we hope can offer valuable insights and contribute to the advancement of the research community.

\paragraph{Anti-redundancy training:} As stated in Section 4.3, our work highlights a specific weakness of current PRMs in identifying redundant reasoning steps. To mitigate this, one possible approach is to modify the label distribution during training. PRM training data is typically labeled as Correct, Neutral, or Incorrect, where the Neutral label often corresponds to redundant steps. By reducing the proportion of Neutral samples in the training data, we can train PRMs with stronger anti-redundancy capabilities.
\paragraph{Contrastive training:} A high-quality data curation pipeline is introduced in Section 3.1, which can also be adapted to curate training samples labeled with fine-grained error types. By leveraging contrastive learning or preference alignment with the curated data, the error sensitivity and detection capabilities of PRMs can be further improved.
\paragraph{Step-level evaluation for LLMs:} As introduced in Section 5.4, the inconsistency between PRMBench and BoN evaluation reveals the false positive situation and the risk of reward hacking within LM post-training. Therefore, traditional outcome-level label-based evaluation is not enough, highlighting the need for a comprehensive step-level evaluation of LLM's reasoning procedure.

\begin{table*}[h]
\belowrulesep=0pt
\aboverulesep=0pt
\fontsize{14}{21}\selectfont
\centering

\resizebox{\textwidth}{!}{
\begin{tabular}{l c ccccccc}
\toprule[1.5pt]
\textbf{Model Name} & \textbf{PRMScore} & \textbf{F1} & \textbf{Negative F1} & \textbf{Acc} & \textbf{Positive Acc} & \textbf{Negative Acc} & \textbf{First} & \textbf{similarity} \\

\hline \multicolumn{9}{c}{\textit{\textbf{Open-source Process Level Reward Models}}} \\   \hline 
\href{https://huggingface.co/Skywork/Skywork-o1-Open-PRM-Qwen-2.5-1.5B}{Skywork-PRM-1.5B} & 61.1 & 88.6 & 33.6 & 80.6 & 88.8 & 33.2 & 44.7 & 91.7\\
\href{https://huggingface.co/Skywork/Skywork-o1-Open-PRM-Qwen-2.5-7B}{Skywork-PRM-7B} & 65.1 & 89.2 & \underline{40.9} & 81.7 & 88.5 & 42.7 & \textbf{56.6} & 90.1\\
\href{https://huggingface.co/ScalableMath/llemma-7b-prm-prm800k-level-1to3-hf}{Llemma-PRM800k-7B} & 52.0 & 75.7 & 28.3 & 63.7 & 66.4 & \underline{48.5} & 22.2 & 79.5\\
\href{https://huggingface.co/ScalableMath/llemma-7b-prm-metamath-level-1to3-hf}{Llemma-MetaMath-7B} & 50.5 & 80.4 & 20.7 & 68.5 & 75.6 & 27.7 & 15.1 & 84.2\\
\href{https://huggingface.co/ScalableMath/llemma-7b-oprm-prm800k-level-1to3-hf}{Llemma-oprm-7B} & 50.3 & 77.3 & 23.3 & 64.9 & 69.9 & 36.1 & 16.6 & 83.7\\
\href{https://github.com/KbsdJames/MATH-Minos}{MATHMinos-Mistral-7B} & 54.2 & 79.2 & 29.1 & 67.9 & 72.8 & 41.7 & 38.0 & 84.6\\
\end{tabular}}
\caption{
A performance comparison of popular models across detailed metrics in \textbf{ALL} categories of \textsc{PRMBench}. The best performance for each metric is highlighted in \textbf{bold}, while the second-best performance is \underline{underlined}. $^\dagger$: To reduce costs, we evaluated only a subset of 394 samples for o1-mini and Deepseek-R1.
}
\label{apdxtab:fullres_all}
\end{table*}
\begin{table*}[h]
\belowrulesep=0pt
\aboverulesep=0pt
\fontsize{14}{21}\selectfont
\centering

\resizebox{\textwidth}{!}{
\begin{tabular}{l c ccccccc}
\toprule[1.5pt]
\textbf{Model Name} & \textbf{PRMScore} & \textbf{F1} & \textbf{Negative F1} & \textbf{Acc} & \textbf{Positive Acc} & \textbf{Negative Acc} & \textbf{First} & \textbf{similarity} \\

\hline \multicolumn{9}{c}{\textit{\textbf{Open-source Process Level Reward Models}}} \\   \hline 
\href{https://huggingface.co/Skywork/Skywork-o1-Open-PRM-Qwen-2.5-1.5B}{Skywork-PRM-1.5B} & 52.0 & 88.4 & 15.6 & 79.6 & 88.8 & 15.1 & 18.7 & 95.0\\
\href{https://huggingface.co/Skywork/Skywork-o1-Open-PRM-Qwen-2.5-7B}{Skywork-PRM-7B} & \underline{56.4} & 89.0 & \underline{23.9} & 80.7 & 88.8 & 24.2 & 31.3 & 94.5\\
\href{https://huggingface.co/ScalableMath/llemma-7b-prm-prm800k-level-1to3-hf}{Llemma-PRM800k-7B} & 49.3 & 77.7 & 20.9 & 65.2 & 69.2 & \underline{36.9} & 20.5 & 86.4\\
\href{https://huggingface.co/ScalableMath/llemma-7b-prm-metamath-level-1to3-hf}{Llemma-MetaMath-7B} & 50.2 & 85.3 & 15.0 & 75.0 & 83.2 & 17.6 & 15.3 & 94.4\\
\href{https://huggingface.co/ScalableMath/llemma-7b-oprm-prm800k-level-1to3-hf}{Llemma-oprm-7B} & 48.7 & 80.3 & 17.2 & 68.1 & 74.1 & 26.5 & 16.0 & 92.0\\
\href{https://github.com/KbsdJames/MATH-Minos}{MATHMinos-Mistral-7B} & 48.8 & 82.7 & 14.9 & 71.2 & 79.2 & 19.0 & 20.8 & 95.7\\
\end{tabular}}
\caption{
A performance comparison of popular models across detailed metrics in \textbf{NR.} sub-category of \textsc{PRMBench}. The best performance for each metric is highlighted in \textbf{bold}, while the second-best performance is \underline{underlined}. $^\dagger$: To reduce costs, we evaluated only a subset of 394 samples for o1-mini and Deepseek-R1.
}
\label{apdxtab:fullres_nr}
\end{table*}

\begin{table*}[h]
\belowrulesep=0pt
\aboverulesep=0pt
\fontsize{14}{21}\selectfont
\centering

\resizebox{\textwidth}{!}{
\begin{tabular}{l c ccccccc}
\toprule[1.5pt]
\textbf{Model Name} & \textbf{PRMScore} & \textbf{F1} & \textbf{Negative F1} & \textbf{Acc} & \textbf{Positive Acc} & \textbf{Negative Acc} & \textbf{First} & \textbf{similarity} \\

\hline \multicolumn{9}{c}{\textit{\textbf{Open-source Process Level Reward Models}}} \\   \hline 
\href{https://huggingface.co/Skywork/Skywork-o1-Open-PRM-Qwen-2.5-1.5B}{Skywork-PRM-1.5B} & 35.8 & 34.7 & 36.9 & 35.8 & 22.7 & \textbf{75.2} & \underline{71.4} & 72.9\\
\href{https://huggingface.co/Skywork/Skywork-o1-Open-PRM-Qwen-2.5-7B}{Skywork-PRM-7B} & 41.2 & 44.2 & 38.1 & 41.3 & 30.9 & \underline{72.6} & \textbf{71.6} & 73.5\\
\href{https://huggingface.co/ScalableMath/llemma-7b-prm-prm800k-level-1to3-hf}{Llemma-PRM800k-7B} & 53.4 & 65.5 & \textbf{41.3} & 56.6 & 54.9 & 61.4 & 15.7 & 65.6\\
\href{https://huggingface.co/ScalableMath/llemma-7b-prm-metamath-level-1to3-hf}{Llemma-MetaMath-7B} & 50.5 & 70.8 & 30.3 & 58.8 & 66.4 & 36.0 & 15.7 & 74.8\\
\href{https://huggingface.co/ScalableMath/llemma-7b-oprm-prm800k-level-1to3-hf}{Llemma-oprm-7B} & 49.3 & 66.3 & 32.2 & 55.0 & 59.0 & 42.9 & 10.2 & 73.3\\
\href{https://github.com/KbsdJames/MATH-Minos}{MATHMinos-Mistral-7B} & 54.0 & 70.8 & 37.2 & 60.2 & 66.6 & 43.2 & 40.6 & 79.1\\
\end{tabular}}
\caption{
A performance comparison of popular models across detailed metrics in \textbf{NCL.} sub-category of \textsc{PRMBench}. The best performance for each metric is highlighted in \textbf{bold}, while the second-best performance is \underline{underlined}. $^\dagger$: To reduce costs, we evaluated only a subset of 394 samples for o1-mini and Deepseek-R1.
}
\label{apdxtab:fullres_ncl}
\end{table*}

\begin{table*}[h]
\belowrulesep=0pt
\aboverulesep=0pt
\fontsize{14}{21}\selectfont
\centering

\resizebox{\textwidth}{!}{
\begin{tabular}{l c ccccccc}
\toprule[1.5pt]
\textbf{Model Name} & \textbf{PRMScore} & \textbf{F1} & \textbf{Negative F1} & \textbf{Acc} & \textbf{Positive Acc} & \textbf{Negative Acc} & \textbf{First} & \textbf{similarity} \\

\hline \multicolumn{9}{c}{\textit{\textbf{Open-source Process Level Reward Models}}} \\   \hline 
\href{https://huggingface.co/Skywork/Skywork-o1-Open-PRM-Qwen-2.5-1.5B}{Skywork-PRM-1.5B} & 32.4 & 32.9 & 32.0 & 32.4 & 20.3 & \underline{85.4} & \underline{76.4} & 72.2\\
\href{https://huggingface.co/Skywork/Skywork-o1-Open-PRM-Qwen-2.5-7B}{Skywork-PRM-7B} & 36.7 & 39.1 & 34.4 & 36.8 & 24.9 & \textbf{88.9} & \textbf{82.5} & 67.6\\
\href{https://huggingface.co/ScalableMath/llemma-7b-prm-prm800k-level-1to3-hf}{Llemma-PRM800k-7B} & 56.4 & 76.8 & 36.1 & 65.9 & 69.2 & 51.7 & 20.0 & 81.2\\
\href{https://huggingface.co/ScalableMath/llemma-7b-prm-metamath-level-1to3-hf}{Llemma-MetaMath-7B} & 51.9 & 78.0 & 25.8 & 66.0 & 73.9 & 31.7 & 15.4 & 83.3\\
\href{https://huggingface.co/ScalableMath/llemma-7b-oprm-prm800k-level-1to3-hf}{Llemma-oprm-7B} & 54.2 & 77.3 & 31.1 & 65.9 & 71.4 & 41.6 & 15.6 & 83.5\\
\href{https://github.com/KbsdJames/MATH-Minos}{MATHMinos-Mistral-7B} & 57.0 & 77.3 & 36.7 & 66.6 & 71.1 & 48.6 & 39.7 & 82.0\\
\end{tabular}}
\caption{
A performance comparison of popular models across detailed metrics in \textbf{ES.} sub-category of \textsc{PRMBench}. The best performance for each metric is highlighted in \textbf{bold}, while the second-best performance is \underline{underlined}. $^\dagger$: To reduce costs, we evaluated only a subset of 394 samples for o1-mini and Deepseek-R1.
}
\label{apdxtab:fullres_es}
\end{table*}

\begin{table*}[h]
\belowrulesep=0pt
\aboverulesep=0pt
\fontsize{14}{21}\selectfont
\centering

\resizebox{\textwidth}{!}{
\begin{tabular}{l c ccccccc}
\toprule[1.5pt]
\textbf{Model Name} & \textbf{PRMScore} & \textbf{F1} & \textbf{Negative F1} & \textbf{Acc} & \textbf{Positive Acc} & \textbf{Negative Acc} & \textbf{First} & \textbf{similarity} \\

\hline \multicolumn{9}{c}{\textit{\textbf{Open-source Process Level Reward Models}}} \\   \hline 
\href{https://huggingface.co/Skywork/Skywork-o1-Open-PRM-Qwen-2.5-1.5B}{Skywork-PRM-1.5B} & 64.9 & 91.1 & 38.7 & 84.4 & 88.7 & 47.1 & 58.3 & 91.2\\
\href{https://huggingface.co/Skywork/Skywork-o1-Open-PRM-Qwen-2.5-7B}{Skywork-PRM-7B} & 67.1 & 91.0 & 43.2 & 84.4 & 87.7 & 56.6 & \underline{69.6} & 88.9\\
\href{https://huggingface.co/ScalableMath/llemma-7b-prm-prm800k-level-1to3-hf}{Llemma-PRM800k-7B} & 47.1 & 75.2 & 18.9 & 62.1 & 64.3 & 42.5 & 24.1 & 80.5\\
\href{https://huggingface.co/ScalableMath/llemma-7b-prm-metamath-level-1to3-hf}{Llemma-MetaMath-7B} & 47.6 & 81.3 & 13.9 & 69.2 & 74.5 & 23.9 & 15.5 & 85.0\\
\href{https://huggingface.co/ScalableMath/llemma-7b-oprm-prm800k-level-1to3-hf}{Llemma-oprm-7B} & 46.8 & 77.6 & 16.1 & 64.6 & 68.3 & 32.6 & 16.8 & 83.4\\
\href{https://github.com/KbsdJames/MATH-Minos}{MATHMinos-Mistral-7B} & 52.1 & 80.7 & 23.5 & 69.2 & 72.6 & 42.4 & 39.4 & 85.5\\
\end{tabular}}
\caption{
A performance comparison of popular models across detailed metrics in \textbf{SC.} sub-category of \textsc{PRMBench}. The best performance for each metric is highlighted in \textbf{bold}, while the second-best performance is \underline{underlined}. $^\dagger$: To reduce costs, we evaluated only a subset of 394 samples for o1-mini and Deepseek-R1.
}
\label{apdxtab:fullres_sc}
\end{table*}

\begin{table*}[h]
\belowrulesep=0pt
\aboverulesep=0pt
\fontsize{14}{21}\selectfont
\centering

\resizebox{\textwidth}{!}{
\begin{tabular}{l c ccccccc}
\toprule[1.5pt]
\textbf{Model Name} & \textbf{PRMScore} & \textbf{F1} & \textbf{Negative F1} & \textbf{Acc} & \textbf{Positive Acc} & \textbf{Negative Acc} & \textbf{First} & \textbf{similarity} \\

\hline \multicolumn{9}{c}{\textit{\textbf{Open-source Process Level Reward Models}}} \\   \hline 
\href{https://huggingface.co/Skywork/Skywork-o1-Open-PRM-Qwen-2.5-1.5B}{Skywork-PRM-1.5B} & 63.3 & 89.6 & 37.0 & 82.2 & 87.9 & 42.1 & 49.9 & 90.5\\
\href{https://huggingface.co/Skywork/Skywork-o1-Open-PRM-Qwen-2.5-7B}{Skywork-PRM-7B} & \underline{67.7} & 90.2 & \underline{45.2} & 83.3 & 87.3 & 55.4 & \underline{65.4} & 88.3\\
\href{https://huggingface.co/ScalableMath/llemma-7b-prm-prm800k-level-1to3-hf}{Llemma-PRM800k-7B} & 46.7 & 71.6 & 21.9 & 58.3 & 59.9 & 47.1 & 20.5 & 74.3\\
\href{https://huggingface.co/ScalableMath/llemma-7b-prm-metamath-level-1to3-hf}{Llemma-MetaMath-7B} & 44.4 & 76.0 & 12.9 & 62.3 & 68.0 & 22.5 & 12.9 & 79.1\\
\href{https://huggingface.co/ScalableMath/llemma-7b-oprm-prm800k-level-1to3-hf}{Llemma-oprm-7B} & 44.5 & 73.4 & 15.6 & 59.6 & 63.8 & 30.2 & 13.6 & 79.4\\
\href{https://github.com/KbsdJames/MATH-Minos}{MATHMinos-Mistral-7B} & 50.7 & 76.1 & 25.3 & 63.8 & 66.6 & 45.8 & 39.0 & 80.6\\
\end{tabular}}
\caption{
A performance comparison of popular models across detailed metrics in \textbf{DC.} sub-category of \textsc{PRMBench}. The best performance for each metric is highlighted in \textbf{bold}, while the second-best performance is \underline{underlined}. $^\dagger$: To reduce costs, we evaluated only a subset of 394 samples for the o1-mini and Deepseek-R1.
}
\label{apdxtab:fullres_dc}
\end{table*}

\begin{table*}[h]
\belowrulesep=0pt
\aboverulesep=0pt
\fontsize{14}{21}\selectfont
\centering

\resizebox{\textwidth}{!}{
\begin{tabular}{l c ccccccc}
\toprule[1.5pt]
\textbf{Model Name} & \textbf{PRMScore} & \textbf{F1} & \textbf{Negative F1} & \textbf{Acc} & \textbf{Positive Acc} & \textbf{Negative Acc} & \textbf{First} & \textbf{similarity} \\

\hline \multicolumn{9}{c}{\textit{\textbf{Open-source Process Level Reward Models}}} \\   \hline 
\href{https://huggingface.co/Skywork/Skywork-o1-Open-PRM-Qwen-2.5-1.5B}{Skywork-PRM-1.5B} & 66.5 & 91.3 & 41.7 & 84.8 & 89.7 & 47.3 & 62.2 & 92.5\\
\href{https://huggingface.co/Skywork/Skywork-o1-Open-PRM-Qwen-2.5-7B}{Skywork-PRM-7B} & 69.9 & 91.5 & 48.2 & 85.4 & 88.8 & 59.3 & 71.5 & 90.8\\
\href{https://huggingface.co/ScalableMath/llemma-7b-prm-prm800k-level-1to3-hf}{Llemma-PRM800k-7B} & 53.3 & 79.9 & 26.6 & 68.5 & 70.9 & 50.0 & 29.4 & 86.2\\
\href{https://huggingface.co/ScalableMath/llemma-7b-prm-metamath-level-1to3-hf}{Llemma-MetaMath-7B} & 52.1 & 83.7 & 20.4 & 73.0 & 78.5 & 30.3 & 15.6 & 88.9\\
\href{https://huggingface.co/ScalableMath/llemma-7b-oprm-prm800k-level-1to3-hf}{Llemma-oprm-7B} & 53.5 & 81.7 & 25.4 & 70.6 & 74.0 & 43.8 & 27.2 & 89.0\\
\href{https://github.com/KbsdJames/MATH-Minos}{MATHMinos-Mistral-7B} & 57.8 & 82.5 & 33.1 & 72.3 & 74.5 & 56.3 & 53.8 & 87.3\\
\end{tabular}}
\caption{
A performance comparison of popular models across detailed metrics in \textbf{CI.} sub-category of \textsc{PRMBench}. The best performance for each metric is highlighted in \textbf{bold}, while the second-best performance is \underline{underlined}. $^\dagger$: To reduce costs, we evaluated only a subset of 394 samples for o1-mini and Deepseek-R1.
}
\label{apdxtab:fullres_ci}
\end{table*}

\begin{table*}[h]
\belowrulesep=0pt
\aboverulesep=0pt
\fontsize{14}{21}\selectfont
\centering

\resizebox{\textwidth}{!}{
\begin{tabular}{l c ccccccc}
\toprule[1.5pt]
\textbf{Model Name} & \textbf{PRMScore} & \textbf{F1} & \textbf{Negative F1} & \textbf{Acc} & \textbf{Positive Acc} & \textbf{Negative Acc} & \textbf{First} & \textbf{similarity} \\

\hline \multicolumn{9}{c}{\textit{\textbf{Open-source Process Level Reward Models}}} \\   \hline 
\href{https://huggingface.co/Skywork/Skywork-o1-Open-PRM-Qwen-2.5-1.5B}{Skywork-PRM-1.5B} & 33.1 & 36.2 & 30.1 & 33.3 & 23.2 & \textbf{78.4} & \textbf{73.9} & 70.2\\
\href{https://huggingface.co/Skywork/Skywork-o1-Open-PRM-Qwen-2.5-7B}{Skywork-PRM-7B} & 36.8 & 42.9 & 30.8 & 37.4 & 28.8 & \underline{76.0} & \underline{72.8} & 70.2\\
\href{https://huggingface.co/ScalableMath/llemma-7b-prm-prm800k-level-1to3-hf}{Llemma-PRM800k-7B} & 51.0 & 71.7 & 30.3 & 59.7 & 62.3 & 47.9 & 24.4 & 75.0\\
\href{https://huggingface.co/ScalableMath/llemma-7b-prm-metamath-level-1to3-hf}{Llemma-MetaMath-7B} & 50.5 & 78.7 & 22.4 & 66.5 & 75.5 & 26.5 & 15.6 & 82.8\\
\href{https://huggingface.co/ScalableMath/llemma-7b-oprm-prm800k-level-1to3-hf}{Llemma-oprm-7B} & 49.2 & 74.4 & 24.1 & 61.7 & 68.0 & 33.4 & 16.1 & 81.2\\
\href{https://github.com/KbsdJames/MATH-Minos}{MATHMinos-Mistral-7B} & 52.8 & 77.5 & 28.0 & 65.7 & 73.3 & 34.2 & 29.5 & 83.1\\
\end{tabular}}
\caption{
A performance comparison of popular models across detailed metrics in \textbf{PS.} sub-category of \textsc{PRMBench}. The best performance for each metric is highlighted in \textbf{bold}, while the second-best performance is \underline{underlined}. $^\dagger$: To reduce costs, we evaluated only a subset of 394 samples for o1-mini and Deepseek-R1.
}
\label{apdxtab:fullres_ps}
\end{table*}

\begin{table*}[h]
\belowrulesep=0pt
\aboverulesep=0pt
\fontsize{14}{21}\selectfont
\centering

\resizebox{\textwidth}{!}{
\begin{tabular}{l c ccccccc}
\toprule[1.5pt]
\textbf{Model Name} & \textbf{PRMScore} & \textbf{F1} & \textbf{Negative F1} & \textbf{Acc} & \textbf{Positive Acc} & \textbf{Negative Acc} & \textbf{First} & \textbf{similarity} \\

\hline \multicolumn{9}{c}{\textit{\textbf{Open-source Process Level Reward Models}}} \\   \hline 
\href{https://huggingface.co/Skywork/Skywork-o1-Open-PRM-Qwen-2.5-1.5B}{Skywork-PRM-1.5B} & 32.3 & 36.3 & 28.2 & 32.5 & 22.8 & \textbf{84.0} & \textbf{79.1} & 74.5\\
\href{https://huggingface.co/Skywork/Skywork-o1-Open-PRM-Qwen-2.5-7B}{Skywork-PRM-7B} & 37.4 & 44.9 & 29.9 & 38.3 & 29.8 & \underline{83.3} & \underline{77.8} & 73.3\\
\href{https://huggingface.co/ScalableMath/llemma-7b-prm-prm800k-level-1to3-hf}{Llemma-PRM800k-7B} & 53.5 & 77.8 & 29.1 & 66.2 & 70.4 & 43.8 & 23.1 & 85.1\\
\href{https://huggingface.co/ScalableMath/llemma-7b-prm-metamath-level-1to3-hf}{Llemma-MetaMath-7B} & 51.3 & 80.6 & 22.1 & 68.9 & 76.6 & 27.9 & 15.1 & 85.6\\
\href{https://huggingface.co/ScalableMath/llemma-7b-oprm-prm800k-level-1to3-hf}{Llemma-oprm-7B} & 51.3 & 78.4 & 24.1 & 66.4 & 72.5 & 33.8 & 17.2 & 86.6\\
\href{https://github.com/KbsdJames/MATH-Minos}{MATHMinos-Mistral-7B} & 55.8 & 79.1 & 32.4 & 68.1 & 72.8 & 45.2 & 41.4 & 83.7\\
\end{tabular}}
\caption{
A performance comparison of popular models across detailed metrics in \textbf{DR.} sub-category of \textsc{PRMBench}. The best performance for each metric is highlighted in \textbf{bold}, while the second-best performance is \underline{underlined}. $^\dagger$: To reduce costs, we evaluated only a subset of 394 samples for o1-mini and Deepseek-R1.
}
\label{apdxtab:fullres_dr}
\end{table*}

\end{document}